\documentclass[conference]{IEEEtran}
\IEEEoverridecommandlockouts
\usepackage{cite}
\PassOptionsToPackage{numbers, sort&compress}{natbib}
\usepackage{paralist, tabularx}
\usepackage{amsmath}
\usepackage{float}
\usepackage{hyperref}
\hypersetup{colorlinks,urlcolor=blue,citecolor=blue,breaklinks=true}
\usepackage{url}

\usepackage{breakurl}
\usepackage{cellspace}
\usepackage{booktabs,adjustbox}
\usepackage{tcolorbox}
\newtcolorbox{mybox}{colback=black!5!white,colframe=black,bottomrule=.25mm,toprule=.25mm,leftrule=.25mm,rightrule=.25mm,left=.25mm,right=.25mm,top=-.25mm,bottom=-.25mm}
\usepackage{wrapfig}
\usepackage{amsfonts}
\usepackage{mathtools} 
\usepackage[english]{babel}
\usepackage{amsmath,amssymb,amsthm}
\usepackage{tabularx}
\usepackage{algorithm}
\usepackage{algpseudocode}

\usepackage{multirow}
\usepackage{array}
\usepackage{caption,subcaption}
\usepackage{listings}
\usepackage{diagbox}
\usepackage{tablefootnote}
\def\BibTeX{{\rm B\kern-.05em{\sc i\kern-.025em b}\kern-.08em
    T\kern-.1667em\lower.7ex\hbox{E}\kern-.125emX}}

\makeatletter
\def\addlegendimage{\pgfplots@addlegendimage}
\makeatother

\usepackage{pifont}% http://ctan.org/pkg/pifont
\usepackage{comment}

\usepackage[utf8]{inputenc}
\newcommand{\paragraphb}[1]{\noindent{\bf #1} }
\newcommand{\paragraphe}[1]{\vspace{0.03in} \noindent{\em #1} }
\newcommand{\paragraphbe}[1]{\vspace{0.03in} \noindent{\bf \em #1} }

\usepackage{colortbl}

\newcommand{\red}[1]{\textcolor{red}{#1}}

\definecolor{auburn}{rgb}{0.43, 0.21, 0.1}
\definecolor{burgundy}{rgb}{0.5, 0.0, 0.13}

\newcommand{\tabincell}[2]{\begin{tabular}{@{}#1@{}}#2\end{tabular}}

\makeatletter
\newcommand*\titleheader[1]{\gdef\@titleheader{#1}}
\AtBeginDocument{%
  \let\st@red@title\@title
  \def\@title{%
    \bgroup\small\large\centering\@titleheader\par\egroup
    \vskip.5em\st@red@title}
}
\makeatother
\usepackage{etoolbox}
\makeatletter
\patchcmd{\@maketitle}
  {\addvspace{0.5\baselineskip}\egroup}
  {\addvspace{-1em}\egroup}
  {}
  {}
\makeatother
\titleheader{To appear in the IEEE Symposium on Security \& Privacy, 2022}
\title{Back to the Drawing Board: \\A Critical Evaluation of Poisoning Attacks on Production Federated Learning}

\author{\normalsize Virat Shejwalkar$^*$,\ \ Amir Houmansadr$^*$,\ \  Peter Kairouz$^\dagger$, \ \  Daniel Ramage$^\dagger$\\
\normalsize {\normalsize $^*$University of Massachusetts Amherst\hspace{.5em}
$^\dagger$Google Research}\\
\normalsize $^*$\{vshejwalkar, amir\}@cs.umass.edu, $^\dagger$\{kairouz, dramage\}@google.com
}
\begin{document}
% \vspace*{-1em}
\maketitle
\pagestyle{plain}
{\allowdisplaybreaks
% !TEX root = main.tex
\begin{abstract}

While recent works have indicated that federated learning (FL) may be vulnerable to poisoning attacks by compromised clients,  their real impact on production FL systems  is not fully understood.
In this work, we aim to develop a comprehensive systemization for poisoning attacks on FL 
by enumerating all possible threat models, variations  of poisoning, and adversary capabilities. 
We specifically put our focus on untargeted poisoning attacks, as we argue that they are significantly relevant to production FL deployments. 

We present a critical analysis of untargeted poisoning attacks under practical, production FL environments by carefully characterizing the set of realistic threat models and adversarial capabilities.  Our findings are rather surprising: contrary to the established belief, we show that FL is highly robust in practice even when using simple, low-cost defenses. 
We go even further and propose novel, state-of-the-art data and model poisoning attacks, and show via an extensive set of experiments across three benchmark datasets how (in)effective poisoning attacks are in the presence of simple defense mechanisms. We aim to correct previous misconceptions and offer concrete guidelines to conduct more accurate (and more realistic) research on this topic. 

% that  will encourage the community to conduct more accurate research in this space and build stronger (and more realistic) attacks and defenses. 

%In Federated learning (FL), multiple clients (e.g., mobile devices) train a joint ML model (e.g., next word predictor) without sharing their private data with anyone. Recent literature has demonstrated that FL is vulnerable to poisoning attacks by compromised clients, which reduce the utility of the joint model, and therefore, has proposed multiple defenses to mitigate the attacks.

%We perform \emph{the first} critical analysis of the poisoning attacks under practical FL environments. We systematize the threat models of FL poisoning attacks in order to analyze their practical value. 
% 
%We propose state-of-the-art poisoning attacks under two threat models, which we argue are the only threat models of practical interest in poisoning attacks settings.
% 
%We show that, the unrealistic assumptions in previous works, e.g., using impractically high percentages of compromised clients or assuming unrealistic capabilities for the adversary, lead to highly misleading conclusions. For instance, contrary to the established norm, we show that FL, even without any defenses, is highly robust in practice. 

%With an extensive evaluation of existing and our improved attacks, across three benchmark datasets, we correct previous misconceptions and give concrete guidelines that will help the future FL robustness research to be more practical. 

% Finally, we highlight the interesting open problems in FL robustness research.

% \vspace*{-1.0em}
\end{abstract}
 
% !TEX root = main.tex
\section{Introduction}\label{intro}

Federated learning (FL) is an emerging learning paradigm in which data owners (called \emph{clients}) collaborate in training a common machine learning model without sharing their private training data. In this setting, a central \emph{server} (e.g., a service provider) repeatedly collects some updates that the clients compute using their local private data, aggregates the clients’ updates using an \emph{aggregation rule} (AGR), and finally uses the aggregated client updates to tune the jointly
trained model (called the \emph{global model}), which is broadcasted to a subset of the clients at the end of each FL training round. 
FL is increasingly adopted by various distributed platforms, in particular by Google's Gboard~\cite{gboard} for next word prediction, by Apple's Siri~\cite{paulik2021federated} for automatic speech recognition, and by WeBank~\cite{webank_credit} for credit risk predictions.
% Google's Gboard~\cite{gboard} and Apple's Siri~\cite{paulik2021federated} use FL to train next word prediction models, and WeBank~\cite{webank_credit} uses FL for credit risk predictions.
% \virat{few sentences about production FL here}

\paragraphb{The threat of poisoning FL:}
A key feature that makes FL highly attractive in practice is that it allows training models in collaboration between  \emph{mutually untrusted}   clients, e.g., Android users or competing banks. 
Unfortunately, this makes FL susceptible to a threat known as \emph{poisoning}: a small fraction of FL clients, called \emph{compromised clients}, who are either owned or controlled by an adversary, may act maliciously during the FL training process in order to corrupt the jointly trained global model. 
Specifically, the goal of the poisoning adversary is to attack FL by instructing its compromised clients to  contribute \emph{poisoned} model updates during FL training  in order to \emph{poison} the global model.

There are three major approaches to poisoning FL: \emph{targeted}, \emph{backdoor}, and \emph{untargeted} poisoning;  Figure~\ref{fig:attack_types} briefly illustrates them.
%  \emph{targeted} attacks~\cite{bhagoji2019analyzing,tolpegin2020data,sun2019can} aim to reduce the utility of the global FL model on specific test inputs of adversary's choice; 
%  \emph{untargeted} attacks~\cite{fang2020local,shejwalkar2021manipulating,baruch2019a}  aim to reduce the utility of global model on arbitrary test inputs;
%  and \emph{backdoor} attacks~\cite{bagdasaryan2018how,wang2020attack,xie2019dba}  aim to  reduce the utility on test inputs that contain a specific  signal called the trigger. 
The goal of this work is to understand the significance of poisoning attacks to \emph{production FL systems}~\cite{kairouz2019advances,bonawitz2019towards}, and to reevaluate the need for sophisticated techniques to defend against FL poisoning. \textbf{We choose to focus on untargeted FL poisoning} as we find it to be significantly relevant to production deployments:  it can be used to impact a large population of FL clients and it can remain undetected for long duration.
% and it is unexplored for production FL.
% but also because we believe they impose a greater (and more realistic) risk to production FL than other types of poisoning,  as we will discuss in Section~\ref{threat:attack_types}. 
% 
% As the focus of this work is on untargeted poisoning attacks, 
As we focus on untargeted poisoning, in the rest of this paper ``poisoning'' refers to untargeted poisoning, unless  specified otherwise.

% \paragraphb{Understanding untargeted poisoning in production FL:}
% In this work, we focus on \emph{untargeted} poisoning attacks for several reasons: \red{\textbf{(1)} \emph{Untargeted attacks affect a large portion of FL client population} (the head of the distribution) as they aim to misclassify most (or all) test inputs, while targeted and backdoor attacks affect only a very small FL client population (a tail of the distribution).

% \textbf{(1)} and \textbf{(2)} directly impact the success of adopting FL in practice. 
% \textbf{(3)} all the robust aggregation rules specifically aim to provide theoretical robustness guarantees against these attacks. 
% \textbf{(4)} Finally, unlike the targeted attacks~\cite{bagdasaryan2018how,wang2020attack,bhagoji2019analyzing}, untargeted poisoning attacks have \emph{not} been studied under production FL environments (Section~\ref{threat:practical_ranges}).}

\begin{figure*}
\centering
\includegraphics[scale=.75]{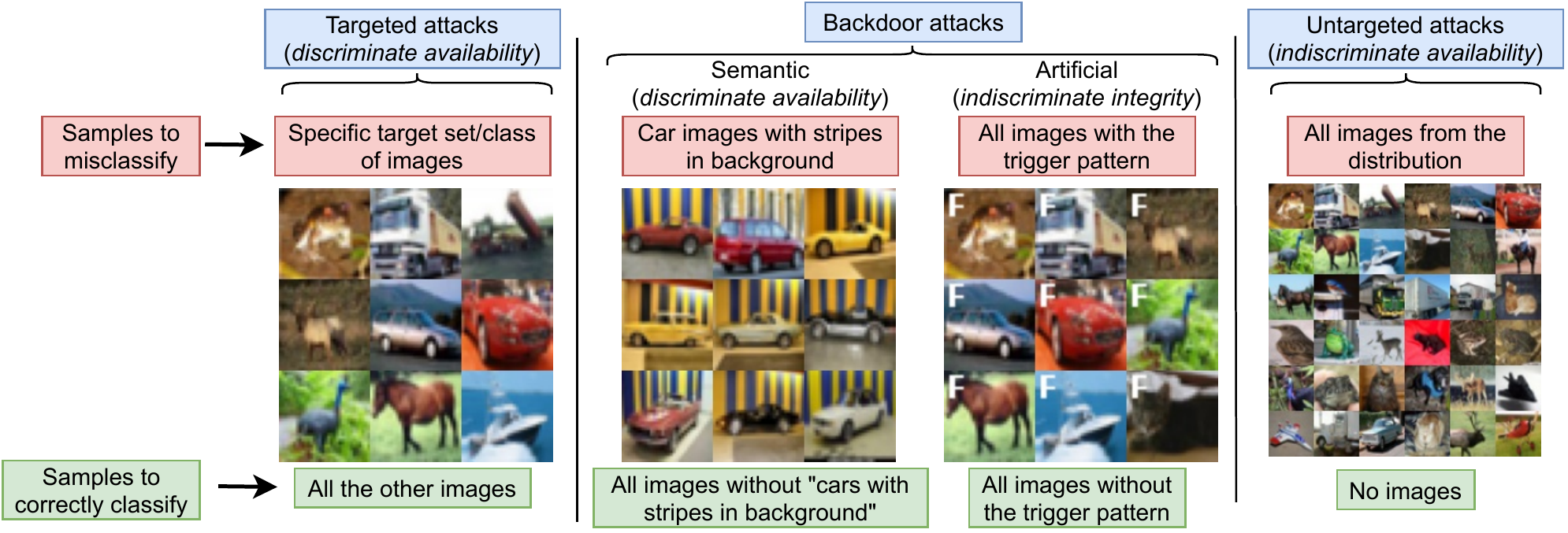}
\vspace*{-.25em}
\caption{Classes of FL poisoning attacks and their objectives defined using the taxonomy in Section~\ref{threat:dimensions_goal}: \emph{Targeted} attacks~\cite{sun2019can,bhagoji2019analyzing} aim to misclassify only a specific set/classes of inputs (e.g., certain 10 samples from CIFAR10), \emph{semantic backdoor} attacks~\cite{bagdasaryan2018how,wang2020attack} aim to misclassify inputs with specific properties (e.g., cars with stripes in background), \emph{artificial backdoor} attacks~\cite{xie2019dba} aim to misclassify  inputs with an artificial (visible or invisible) trigger pattern (e.g., shape of letter "F"), and
\emph{untargeted} attacks~\cite{shejwalkar2021manipulating,fang2020local} aim to reduce  model accuracy on \emph{arbitrary} inputs (e.g., the  entire CIFAR10 distribution). 
% Please check Section~\ref{threat:dimensions_goal} for a detailed discussion.
}
\label{fig:attack_types}
\vspace*{-1.75em}
\end{figure*}

\paragraphb{The literature on FL poisoning attacks and defenses:}
Recent works have presented various techniques (Section~\ref{existing}) to poison FL~\cite{fang2020local,baruch2019a,shejwalkar2021manipulating}. Their core idea is to generate poisoned updates (either by direct manipulation of model updates, called \emph{model poisoning}~\cite{bagdasaryan2018how,shejwalkar2021manipulating,fang2020local,baruch2019a,bhagoji2019analyzing}, or through fabricating poisoned data, called \emph{data poisoning}~\cite{tolpegin2020data,wang2020attack}) that deviate maximally from a benign direction, e.g., the average of benign clients' updates, and at the same time \emph{circumvent} the given robust AGR, i.e., by bypassing its detection criteria. 
% For instance, Baruch et al.~\cite{baruch2019a} generate poisoned updates by adding small noises, proportional to the standard deviation of some benign updates, to the average of benign updates. Alternatively, other recent works~\cite{fang2020local,shejwalkar2021manipulating} generate poisoned updates by solving an optimization problem aimed at maximizing the poisoning impact while circumventing the target AGR.
% 
% 
To protect FL against such poisoning attacks, the literature has designed various \emph{robust} aggregation rules~\cite{blanchard2017machine,yin2018byzantine,chang2019cronus,mhamdi2018the,fang2020local,shejwalkar2021manipulating} (Section~\ref{background:agr}) which aim to remove or attenuate the updates that are more likely to be malicious according to some criterion. 

\paragraphb{The gap between the literature and practice:}
The existing literature on poisoning attacks and defenses for FL makes \emph{unrealistic assumptions} that do not hold in real-world FL deployments, e.g., assumptions about the percentages of compromised clients, total number of FL clients, and the types of FL systems~\cite{kairouz2019advances}.
% Typically, the assumptions are about the convexity of optimization problem, percentages of compromised clients, total number of FL clients, and type of the FL setting.
% 
For instance, state-of-the-art attacks~\cite{fang2020local,baruch2019a,shejwalkar2021manipulating} (defenses~\cite{blanchard2017machine,yin2018byzantine,xie2018generalized,chen2018draco}) assume adversaries who can compromise up to 25\% (50\%) of FL clients.
For an app like Gboard with $\sim1B$ installations~\cite{kairouz2019advances}, 25\% compromised clients would mean an attacker controls \emph{250 million Android devices}! 
We argue that, although interesting from theoretical perspectives, the assumptions in recent FL robustness works do not represent common real-world adversarial scenarios that account for the difficulty and cost of at-scale compromises.
% (Section~\ref{threat:practical_ranges}).

\paragraphb{\em Our contributions:}
% 
% To address the aforementioned gap, 
In this work, we perform a critical analysis of the literature on FL robustness against (untargeted) poisoning under practical considerations. Our ultimate goal is to understand the significance of poisoning attacks and the need for sophisticated  robust FL algorithms in production FL. Specifically, we make the following key contributions:

\paragraphbe{I. Systemization of FL poisoning threat models.}
We start by establishing a comprehensive systemization of threat models of FL poisoning.
% We focus on untargeted poisoning attacks.
% Note that, we aim to provide a systematic framework to analyze existing and future FL poisoning threat models, as opposed to providing taxonomies~\cite{jere2020taxonomy,goldblum2020dataset} of existing attacks.\red{i dont understand what you mean by this}\virat{i want to say that we don't just want to categorize/review existing attacks, which is what the given taxonomies are doing}
Specifically, we discuss three key dimensions of the poisoning threat to FL: The adversary's objective, knowledge, and capability.
% The adversary's capability based on the FL stage they poison, the adversary's knowledge of the global model, and the attack's mode based on the frequency of poisoning. 
We discuss the practicality of all possible threat models obtained by combining 
these dimensions. 
% the dimensions corresponding to the objective of untargeted poisoning.
% \red{this has also become confusing. it was clear before}\virat{this change is mainly due to new systematization; but essentially it is same as from previous version: "We discuss the practicality of all possible threat models obtained by combining these dimensions."}
As we will discuss, out of all possible combinations, only two threat models, i.e., \emph{nobox offline data poisoning} and \emph{whitebox online model poisoning}, 
are of practical value to production FL.
We believe that prior works~\cite{baruch2019a,blanchard2017machine,fang2020local,shejwalkar2021manipulating} have neglected the crucial constraints of production FL systems on the parameters relevant to  FL robustness.
\emph{Our work is the first to consider  production FL environments}~\cite{kairouz2019advances,bonawitz2019towards} and provide practical ranges for various parameters of poisoning threat models.
As a result, \emph{our evaluations lead to  conclusions that contradict the common beliefs in the literature}, e.g., we show that production FL even with the non-robust Average AGR is significantly more robust than previously thought.
% we show that contrary to the previous understanding, it is not trivial to mount untargeted attacks on production FL.

% In our work, we focus on untargeted poisoning attack, because they are not well studied, in spite of being the sole focus of all of the existing works on robust aggregation rules.
% Here, the adversary's goal is to degrade the utility of the global model on most test inputs. In Section~\ref{threat:dimensions}, we discuss three key dimensions of the untargeted poisoning threat to FL: The adversary's capability based on the FL stage she poisons, the adversary's knowledge of global model, and the attacks' mode based on the frequency of poisoning.

\paragraphbe{II. Introducing improved poisoning attacks.}
First, we overview all of the existing untargeted poisoning attacks on FL that consider the two aforementioned threat models (Section~\ref{existing}).
Then, we design improved attacks for these threat models. 
\textbf{\em 1) Our improved data poisoning attacks:} 
We present  \emph{the first} attacks that systematically consider the data poisoning threat model for FL (Section~\ref{improved:dp}).
We build on the classic label flipping data poisoning attack~\cite{xiao2012adversarial,xiao2015support,newell2014practicality} designed for centralized ML. 
Our data poisoning attacks rely on  the observation that increasing the amount of label flipped data increases the loss and norm of the resulting updates, and therefore, can produce poisoned updates that can effectively reduce the global model's accuracy. However, using arbitrarily large amounts of label flipped data may result in updates that cannot circumvent the robustness criterion of the target AGR. Hence, to circumvent the target AGR, we propose to adjust the amount of label flipped data used.
% Hence, we propose to circumvent the robustness criterion of the target aggregation rule by adjusting the amount of label flipped data used.
% 
\textbf{\em 2) Our improved model poisoning attacks:}
We propose novel model poisoning attacks that outperform the state-of-the-art.
Our attacks 
(Section~\ref{improved:mp}) use \emph{gradient ascent}  to fine-tune the global model and increase its loss on benign data. Then, they adjust the $L_2$-norm of the corresponding poisoned update in order to circumvent robustness criterion of the target AGR.

\paragraphbe{III. Analysis of FL robustness in practice.}
% To understand the impact of poisoning attacks on real-world FL deployments, 
We extensively evaluate all existing poisoning attacks as well as our own improved attacks across three benchmark datasets, for various FL parameters, and for different types of FL deployments.
We make several significant deductions about the state of FL poisoning literature for production FL. {For production cross-device FL},
% \footnote{See Section~\ref{background:fl} for distinction between cross-silo and cross-device FL.}
which contains  thousands to billions of clients, following are our key lessons:

\paragraphb{(1)} \emph{For practical percentages of compromised clients ($M$), even the most basic, non-robust FL algorithm, i.e., Average AGR, converges with high accuracy, i.e., it is highly robust.} For instance,   data poisoning with $M=0.1\%$  (model poisoning with $M=0.01\%$)  reduces the global model accuracy of FEMNIST from 83.4\% to 81.4\% (73.4\%), CIFAR10 from 86.6\% to 85.1\% (82.9\%), and Purchase from 85.4\% to 85.3\% (76.4\%). These findings directly contradict the claims of previous works~\cite{blanchard2017machine,mhamdi2018the,xie2018generalized} that Average AGR cannot converge even with a \emph{single} compromised client. 

\paragraphb{(2)} \emph{Poisoning attacks have no impact on existing robust FL algorithms even with impractically high $M$'s}: At $M$=1\%, {data or model poisoning attacks} reduce the accuracy by only $<$1\% for most of the settings.

\paragraphb{(3)} {\emph{Enforcing a limit on the size of the dataset contributed by each client    can act as a highly effective (yet simple) defense against data poisoning attacks} with no need to any of the state-of-the-art, sophisticated robust FL aggregation algorithms.}

% \paragraphb{(3)} The impacts of DPAs increase with increasing the sizes of local poisoning data, $|D_p|$, used to compute the poisoned updates. However, the resource constraints of FL devices and short duration of FL rounds in practice~\cite{bonawitz2019towards,kairouz2019advances} implicitly limit $|D_p|$. \emph{We show that DPAs have no impact on FL with robust AGRs even when $|D_p|$ is $100\times$ the average size of the benign clients' local data.} \amir{not sure what exciting thing I am learning from this item. remember we talked about making these points exciting}

% \vspace{.25em}

\paragraphb{(4)} While  recent works have introduced sophisticated and theoretically robust AGRs that incur high computation and memory costs, \emph{the simple and low-cost defenses, e.g., norm-bounding~\cite{sun2019can}, provide an equivalent protection to FL against state-of-the-art poisoning attacks.} 

% \vspace{.25em}

% \paragraphb{(5)} \blue{Theoretical robustness guarantees of AGRs can be misleading in practice, because for any $M$, AGRs with theoretical guarantees do not outperform those without guarantees. Hence, thorough empirical assessments of AGRs must accompany theoretical guarantees. %, as only latter can be very misleading in practice.
% }\red{this is not a surprising point. the papers that have theoretical guarantees do not claim to be better than those without guarantees. so, ...}

% \vspace*{-.5em}

For production cross-silo FL,  which contains up to hundred clients~\cite{kairouz2019advances}, 
we show that \emph{data poisoning attacks are completely ineffective, even against non-robust Average AGR}.
We also argue that  model poisoning attacks are unlikely to play a major risk to production cross-silo FL, where the clients involved are bound by contract and their software stacks are professionally maintained (e.g., in banks, hospitals, etc.). 

\paragraphb{\em Implications of our study: } 
Numerous recent works have proposed sophisticated aggregation rules  for FL with strong theoretical robustness guarantees~\cite{blanchard2017machine,mhamdi2018the,yin2018byzantine,alistarh2018byzantine,xie2018generalized,pillutla2019robust,data2020byzantine,el2019sgd}.
% However, we show that,  although interesting from theoretical perspectives, in production FL settings, conclusions of these works are not relevant.
% 
% such scientific efforts are futile, as they do not represent production FL scenarios due to their misleading conclusions about FL robustness and  impractically high costs (computation, memory, and accuracy) of the aggregations.
However, our work shows that, when it comes to production FL deployments, even simple, low-cost defenses can effectively protect FL against poisoning.
We also believe that our systematization of practical poisoning threat models can steer the  community towards practically significant research problems in FL robustness.
% 
% Therefore, we hope that this work can steer the community towards FL robustness research of practical interest, e.g., establishing theoretical guarantees for simple yet effective aggregations under realistic threat models.

% Our work directly impacts both the theoretical and empirical research on FL robustness. 
% % 
% We motivate the research in establishing theoretical guarantees of cheaper yet effective defenses, such as Norm-bounding and bounding the local dataset sizes. 
% % 
% Recently, numerous works have proposed aggregation rules~\cite{blanchard2017machine,mhamdi2018the,yin2018byzantine,alistarh2018byzantine,chen2018draco,xie2019zeno,xie2018generalized,fung2018mitigating,pillutla2019robust,yang2019byrdie,damaskinos2019aggregathor,rajput2019detox,xie2018phocas,sohn2020election,data2020data,data2020byzantine,el2019sgd} with strong theoretical robustness guarantees. We highlight that, although interesting in theory, such scientific efforts are futile, as they do not apply to FL in practice. Because, most of these rules incur high computation and memory costs, and some even reduce the global model accuracy. 
% % 
% Finally, we believe that our systematization of practical poisoning threat models can steer  the  community towards practically significant research problems in FL robustness.

% \input{related_work}
% !TEX root = main.tex
\vspace*{-.25em}
\section{Background} \label{background}
\vspace*{-.15em}
\subsection{Federated Learning (FL)}\label{background:fl}

In FL~\cite{kairouz2019advances, mcmahan2017communication, konevcny2016federated}, a service provider, called \emph{server}, trains a \emph{global model}, $\theta^g$, on the private data of multiple collaborating clients without directly collecting their data.
In the $t^{th}$ FL round, the server selects $n$ out of total $N$ clients and shares the most recent global model, i.e., $\theta^t_g$, with them. Then, a client $k$ uses their local data $D_k$ to fine-tune $\theta^t_g$ using stochastic gradient descent (SGD) for a fixed number of local epochs $E$ and obtains updated model by $\theta^t_k$. Then, the $k^{th}$ client computes her FL \emph{update} as the difference $\nabla^t_k= \theta^t_k-\theta^t_g$ and shares $\nabla^t_k$ with the server.
The server then computes an aggregate of all  client updates using some aggregation rule, $f_\mathsf{agg}$, i.e., using $\nabla^t_\mathsf{agg}= f_\mathsf{agr}(\nabla^t_{\{k\in[n]\}})$.
Then, the server updates the global model of the $(t+1)^{th}$ round using SGD as $\theta^{t+1}_g\leftarrow \theta^{t}_g+\eta\nabla^t_\mathsf{agg}$; here $\eta$ is the server's learning rate.
Section~\ref{threat:production_env} discusses salient features of production FL.
% 

% FL can be either \textbf{cross-device} or \textbf{cross-silo}  \cite{kairouz2019advances}. Each of these types have certain salient features, e.g.,  in cross-device FL, $N$ is large (from few thousands to billions) and only a small fraction of them is chosen in each FL training round, i.e., $n\ll N$. While, in cross-silo FL, $N$ is moderate (up to 100) and all of them are chosen in each round, i.e., $n=N$.
% We list the most important FL parameters that affect its robustness and their ranges in production FL in Table~\ref{tab:practical_ranges}.
% Please refer to Table 1 of the comprehensive survey by Kairouz et al.~\cite{kairouz2019advances} for more details on production FL environments.
% 
% In this work, we thoroughly evaluate robustness of both cross-device and cross-silo FL under realistic production FL settings.
% In this work, we start by analysing cross-device FL because, in spite of being more prone to poisoning attacks, its robustness against untargeted poisoning is \emph{never} studied; all the previous works~\cite{fang2020local,shejwalkar2021manipulating,baruch2019a} mainly focus on cross-silo FL. Then,  we will study  untargeted poisoning on cross-silo FL under practical considerations.}\virat{do we need this in background?

\begin{table*}
\caption{Comparing state-of-the-art aggregation rules (AGRs) in terms of accuracy, computation/memory cost, and theoretical guarantees. We show results for CIFAR10 with 1,000 clients. Red cells show limitations of the corresponding AGR. 
} \label{tab:other_agrs}
\vspace*{-.5em}
\centering
% \fontsize{9}{9}\selectfont{}
% \setlength{\extrarowheight}{0.04cm}
\begin{tabular} {c|c|c|c|c|c}
\hline
\tabincell{c}{Type of aggregation\\ rule (AGR)}  & Example AGR & \tabincell{c}{Accuracy\\in non-iid FL}  & \tabincell{c}{Computation\\at server} & \tabincell{c}{Memory cost\\to client} & \tabincell{c}{Theoretical robustness\\based on} \\ \hline \hline

Non-robust & \textbf{Average}~\cite{mcmahan2017communication} & 86.6 & $\mathcal{O}(d)$ & $\mathcal{O}(d)$ & \cellcolor{red!30}None \\ \hline
 
\multirow{3}{*}{\tabincell{c}{Dimension-wise\\ filtering}  } & Median~\cite{yin2018byzantine} & 84.2 &  $\mathcal{O}(dn\text{log}n)$ & \multirow{3}{*}{$\mathcal{O}(d)$} & convergence \\
 & \textbf{Trimmed-mean}~\cite{yin2018byzantine} & 86.6 & $\mathcal{O}(dn\text{log}n)$ & & convergence \\
 & \tabincell{c}{Sign-SGD +\\ majority voting}~\cite{bernstein2018signsgd} & \cellcolor{red!30}35.1 & $\mathcal{O}(d)$ & & convergence \\ \hline
 
% \tabincell{c}
{Vector-wise scaling} & \textbf{Norm-bound}~\cite{sun2019can} & 86.6 & $\mathcal{O}(d)$ & $\mathcal{O}(d)$ & \cellcolor{red!30} {Not established} \\ \hline

\multirow{6}{*}{\tabincell{c}{Vector-wise\\ filtering} } & Krum~\cite{blanchard2017machine} & \cellcolor{red!30}46.9 & $\mathcal{O}(dn^2)$ & \multirow{6}{*}{$\mathcal{O}(d)$} & convergence \\ 
 & \textbf{Multi-krum}~\cite{blanchard2017machine} & 86.2 & $\mathcal{O}(dn^2)$ & & convergence \\ 
 & Bulyan~\cite{mhamdi2018the} & \cellcolor{red!30}81.1 & $\mathcal{O}(dn^2)$ & & convergence \\
 & RFA~\cite{pillutla2019robust} & 84.6 & $\mathcal{O}(dn^2)$ & & convergence \\ 
 & RSA~\cite{li2019rsa} & \cellcolor{red!30}35.6 & $\mathcal{O}(d)$ & & convergence \\ 
 & DnC~\cite{shejwalkar2021manipulating} & 86.1 & $\mathcal{O}(d)$ & & filtering \\ \hline

\multirow{2}{*}{Certification} & Emsemble~\cite{cao2021provably} & \cellcolor{red!30}74.2 & \multirow{2}{*}{$\mathcal{O}(d)$} & \cellcolor{red!30} & Certification \\ 
 & CRFL~\cite{xie2021crfl} & \cellcolor{red!30}64.1 & & \multirow{-2}{*}{\cellcolor{red!30}$\mathcal{O}(Md)$}  & Certification \\ \hline
 
{\tabincell{c}{Knowledge\\ transfer}} & Cronus~\cite{chang2019cronus} & \cellcolor{red!30} \tabincell{c}{Needs public\\ data} & {$\mathcal{O}(d)$} & {$\mathcal{O}(d)$} & filtering \\ \hline
%  & Ensemble~\cite{lin2020ensemble} & \cellcolor{red!30} data & & & \cellcolor{red!30} None \\ \hline
 
Personalization & \tabincell{c}{Ditto~\cite{li2021ditto}\\ EWC~\cite{yu2020salvaging}} & 86.6 & $\mathcal{O}(d)$ & $\mathcal{O}(d)$ &\cellcolor{red!30} \tabincell{c}{None (depends\\ on server's AGR)} \\ \hline

\hline
\end{tabular}
\vspace*{-1.5em}
\end{table*}

% \vspace*{-.25em}
\subsection{Existing Defenses Against Untargeted Poisoning}\label{background:agr}

As we focus on untargeted poisoning, below we discuss defenses against untargeted poisoning in detail and defer the discussion for targeted/backdoor poisoning to Appendix~\ref{related}.

The literature has presented various directions towards making FL robust against Byzantine or compromised clients. Note that, from detection perspective, there is no difference between untargeted attacks (adversary deliberately corrupts model updates) and Byzantine failures (arbitrary system failures corrupt updates).
The core approach  is to replace
FL's vanilla average aggregation rule (AGR)~\cite{mcmahan2017communication} with a  \textbf{robust AGR} (also called a  \emph{defense}). 
Below, we  introduce the types of robust AGRs designed to defend FL against untargeted poisoning attacks.
% , and then analyze their practicality in production FL based on performance, computation, and memory implications.

% \subsubsection{Types of Robust AGRs and Their Representatives}

% \begin{compactitem}
    \noindent\emph{\underline{Dimension-wise filtering}}  defenses  separately filter potentially malicious values for each dimension of clients' updates. Example AGRs are  Median~\cite{yin2018byzantine}, Trimmed-mean~\cite{yin2018byzantine}, and sign-SGD with majority voting~\cite{bernstein2018signsgd}.
    
    \noindent\emph{\underline{Vector-wise filtering}}  defenses aim at removing potentially poisoned client updates. They differ from  dimension-wise filtering, as they attempt to remove entire malicious  updates, as opposed to removing malicious values. Example AGRs include   RFA~\cite{pillutla2019robust}, RSA~\cite{li2019rsa}, Krum~\cite{blanchard2017machine}, Multi-krum~\cite{blanchard2017machine}, Bulyan~\cite{mhamdi2018the}, and Divide-and-conquer (DnC)~\cite{shejwalkar2021manipulating}.
    
    \noindent\emph{\underline{Vector-wise scaling}} defenses, e.g., Norm-bounding~\cite{sun2019can}, reduce the impact of poisoned updates by scaling their norms.
    
    \noindent\emph{\underline{Certified}} defenses~\cite{cao2021provably,xie2021crfl} provide certified accuracy for each test input when the number of compromised clients or perturbation to the test sample is below a certified threshold.
    
    \noindent\emph{\underline{Knowledge transfer}} based defenses~\cite{chang2019cronus,lin2020ensemble} aim to reduce the dimensionality of the client updates, because theoretical robustness guaranty of most of robust AGRs is directly proportional to updates' dimensionality. Hence, they use knowledge transfer and, instead of sharing parameters of client models, share predictions of client models on some public data.
    
    \noindent\emph{\underline{Personalization}} techniques, e.g. Ditto~\cite{li2021ditto} and EWC~\cite{yu2020salvaging},  fine-tune the potentially corrupt global model on each client's private data to improve its performance for the client.

\subsection{Defenses We Evaluate in Our Work}

For a robust AGR to be usable in production FL, it needs to provide \emph{high performing models at low compute and memory costs}. 
In Table~\ref{tab:other_agrs}, we compare the performance and overhead implications of state-of-the-art defenses overviewed above. Each red cell demonstrates a hindrance to  adoption in production FL.   
In particular, (1) SignSGD + majority voting, Krum, Bulyan, RSA, and certified defenses  incur significant performance losses, (2) certified defenses incur high memory cost to clients, (3) knowledge transfer based defenses require public data and incur high performance losses in cross-device, non-iid FL settings, and (4) personalization techniques cannot improve performance if the global model is significantly corrupt. Hence, to be effective, they should be coupled with a robust AGR and rely completely on the robustness of the AGR against poisoning. Therefore, the evaluation of personalization techniques is orthogonal to our evaluation of robust AGRs.

As the focus of our work is production FL, for brevity and space limitations, we only choose representative AGRs (bold in Table~\ref{tab:other_agrs}) from each class that offer practical performance and overheads. 
% \virat{repeated}
Below we introduce the selected defenses in detail. 
Note that more sophisticated (e.g., higher overhead) defenses may provide better robustness to poisoning, however this does not impact our main conclusions: we will show that even such simple defenses are enough to protect production  FL.  

\subsubsection{Average}\label{agr:average}
In non-adversarial FL settings, dimension-wise Average~\cite{mcmahan2017communication} is an effective AGR. 
Due to its efficiency, Average is the only AGR implemented by FL applications in practice~\cite{gboard,webank_credit,ludwig2020ibm,paulik2021federated}.
% \red{Multiple works have argued that, in theory, even a single compromised client can poison the global model of an Average-based FL. But, unlike our work, none of the previous work systematically studies the robustness of Average AGR under practical threat models.}\virat{repeat at the start of Section~\ref{exp:nonrobust_fl}}

% Therefore,  multiple \emph{Byzantine-robust} AGRs for FL~\cite{blanchard2017machine,mhamdi2018the,xie2018generalized,yin2018byzantine,alistarh2018byzantine,he2020byzantine,chang2019cronus} are proposed to defend against poisoning attacks by compromised clients.

\subsubsection{Norm-bounding}\label{agr:normb}
This AGR~\cite{sun2019can} bounds the $L_2$ norm of all submitted client updates to a fixed threshold, with the intuition that the effective poisoned updates should have high norms. For a threshold $\tau$ and an update $\nabla$, if the norm, $\Vert\nabla\Vert_2 > \tau$, $\nabla$ is scaled by $\frac{\tau}{\Vert\nabla\Vert_2}$, otherwise the update is not changed.
The final aggregate is an average of all the updates, scaled or otherwise.
% The intuition here is that, to effectively poison FL, poisoned updates should have high norms, and hence, clipping them will reduce their poisoning impact. 
% Norm-bounding is a practical and easy-to-implement AGR, and is shown to effectively thwart backdoor attacks on FL~\cite{sun2019can}. 
% In this paper, for the first time, we study efficacy of Norm-bounding against untargeted poisoning attacks~\cite{fang2020local,shejwalkar2021manipulating}.

\subsubsection{Multi-krum}\label{agr:mkrum}
Blanchard et al.~\cite{blanchard2017machine} proposed Multi-krum AGR as a modification to their own Krum AGR~\cite{blanchard2017machine}. Multi-krum selects an update using Krum and adds it to a \emph{selection set}, $S$. Multi-krum repeats this for the \emph{remaining updates} (which remain after removing the update that Krum selects) until $S$ has $c$ updates such that $n-c > 2m+2$, where $n$ is the number of selected clients and $m$ is the number of compromised clients in a given round.
% Multi-krum selects an update using Krum~\cite{blanchard2017machine} from a remaining-set (initialized to the set of all the received updates), adds it to a selection-set (initialized to an empty set), and removes it from the remaining-set.
% This way, Multi-krum selects $c$ updates such that $n-c > 2m+2$, \red{where n and m are ...}.
Finally, Multi-krum averages the updates in $S$.
% Multi-krum significantly outperforms Krum in terms of the global model accuracy.

\subsubsection{Trimmed-mean}\label{agr:trimmed-mean}
Trimmed-mean~\cite{yin2018byzantine,xie2018generalized} aggregates each dimension of input updates separately.
It sorts the values of the $j^{th}$-dimension of all updates.
% , i.e., sorts $\nabla^j_{\{i\in[n]\}}$.
Then it removes $m$ (i.e., the number of compromised clients) of the largest and smallest values of that dimension, and computes the average of the rest of the values as its aggregate of the dimension $j$.

% !TEX root = main.tex

\section{Systemization of FL Poisoning Threat Models}\label{threat}

We discuss the key dimensions of the threat models of poisoning attacks on FL, and argue that only two combinations of these dimensions are of practical interest for production FL. 
Note that, there exist taxonomies of FL poisoning attacks~\cite{jere2020taxonomy,goldblum2020dataset} which provide comprehensive overviews of the poisoning attacks in existing literature. In contrast, our work aims to provide a systematic framework to model the existing and future poisoning threats to federated learning.

\begin{table*}[h]
\vspace*{-.5em}
\centering
% \fontsize{9}{10}\selectfont{}
% \setlength{\extrarowheight}{0.01cm}
\caption{The key dimensions of the threat models of poisoning attacks on FL. Each combination of these dimensions constitutes a threat model (Table~\ref{tab:threat_combinations}). However, we argue in Section~\ref{threat:justification} that only two of these combinations are   practical threat models.}
\label{table:poisoning_dimensions}
\vspace*{-.25em}
% \begin{tabular}{ |p{2cm}|p{2cm}|p{12.45cm}| } 
\begin{tabular}{ |p{1.3cm}|p{2cm}|p{1.5cm}|p{11.65cm}| } 
\hline

\textbf{Dimension} & \textbf{Attribute} & \textbf{Values} & \textbf{Description} \\ \hline \hline

% \multicolumn{3}{c}{\bf \cellcolor{gray!20} Adversary's objective} \\ \hline

\multirow{6}{*}{\bf \tabincell{c}{Objective\\of the\\adversary}} & \multirow{2}{2cm}{\em \tabincell{c}{Security violation}} & Integrity & Misclassify a (adversarially crafted) test input in order to evade detection. \\ \cline{3-4}

& & Availability & Misclassify an unmodified test input to cause service disruption for benign users. \\ \cline{2-4}

& \multirow{2}{2cm}{\em  \tabincell{c}{Attack specificity}} & Discriminate & Misclassify a small and/or  specific set of inputs at the test time.\\ \cline{3-4}

& & {Indiscriminate} & Misclassify all or most of inputs at the test time. \\ \cline{2-4}

& \multirow{2}{2cm}{\em  \tabincell{c}{Error specificity}} & Specific & Misclassify a given modified/pristine test input to a specific class.  \\ \cline{3-4}

& & Generic & Misclassify a given modified/pristine test input to any class. \\ \hline \hline

% \multicolumn{3}{c}{\bf \cellcolor{gray!20} Adversary's knowledge} \\ \hline

\multirow{15}{*}{\bf \tabincell{c}{Knowledge\\of the\\ adversary}} & \multirow{10}{2cm}{\em \tabincell{c}{Knowledge of\\the global model}} 
& \multirow{2}{2cm}{Whitebox} & Adversary can access the global model parameters as well as its predictions, e.g., in the model poisoning case.\\ 
& & & \multicolumn{1}{c|}{\includegraphics[scale=.5]{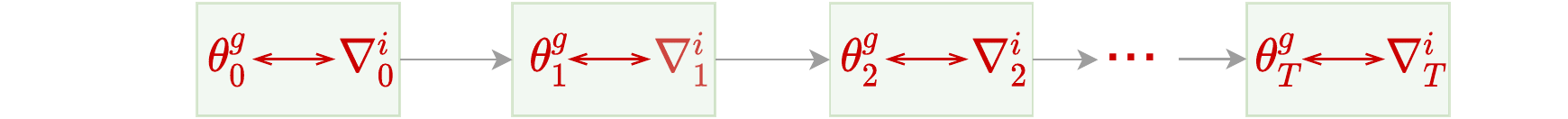}}\\ \cline{3-4}

& & \multirow{2}{2cm}{Nobox} & Adversary cannot access parameters or predictions of global model, e.g., in the data poisoning case. \\
& & & \multicolumn{1}{c|}{\includegraphics[scale=.5]{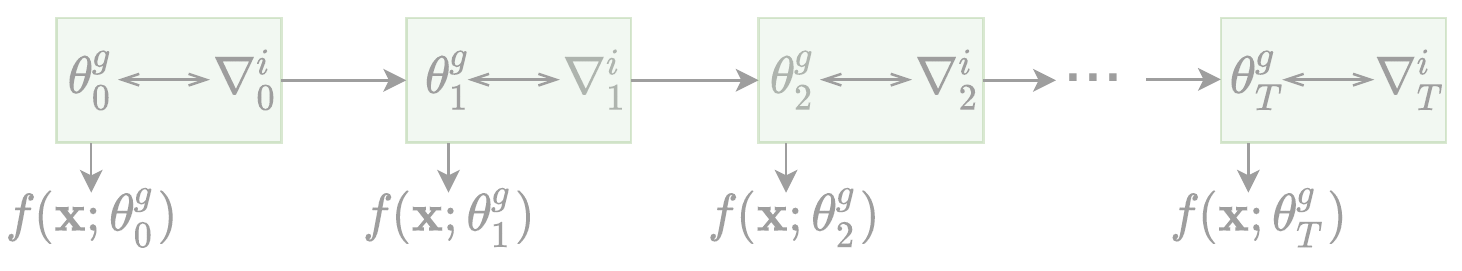}} \\  \cline{2-4}

% & \multirow{2}{*}{\em \tabincell{c}{Knowledge of\\the training data\\of clients}} 
% & \multirow{1}{*}{Full} & Adversary can access the data of all (benign and compromised) of the collaborating clients. \\  \cline{3-4}

% & & \multirow{1}{*}{Partial} & Adversary can access the local data of only the compromised clients.\\ \hline
% \hline

& \multirow{7}{*}{\em \tabincell{c}{Knowledge of\\the data from\\the distribution of\\benign clients'\\data}} 
& \multirow{4}{*}{Full} & \multirow{7}{*}{\hspace{-.48em}\includegraphics[scale=.57]{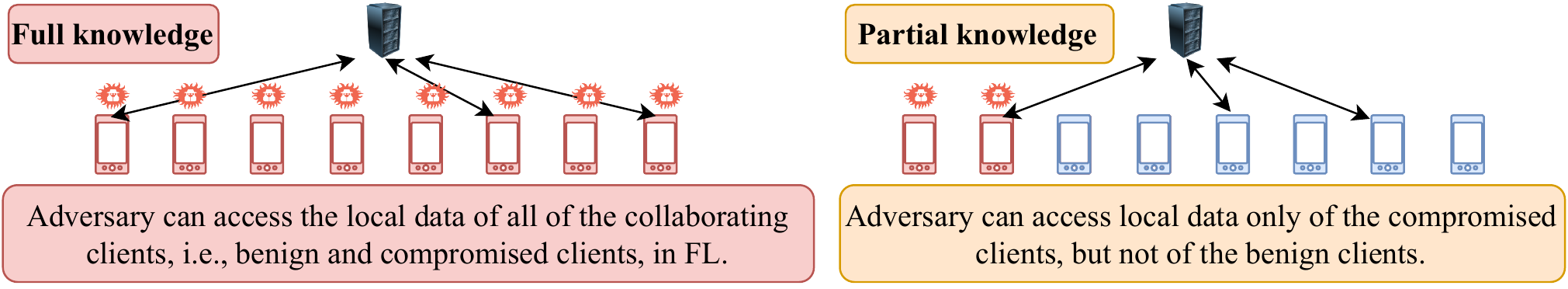}} \\  
&&&\\ 
&&&\\
&&&\\\cline{3-3}

& & \multirow{3}{*}{Partial} & \\ 
&&&\\
&&&\\
\hline
\hline

\multirow{20}{*}{\bf \tabincell{c}{Capabilities\\of the\\adversary}} & \multirow{14}{2cm}{\em \tabincell{c}{Capabilities in\\terms of access\\to client devices}} & \multirow{2}{2cm}{Model poison} & Adversary breaks into the compromised clients (e.g., by circumventing security protocols of operating systems such as Android) and directly manipulates their model updates.  \\
& & & \multicolumn{1}{c|}{\includegraphics[scale=.5]{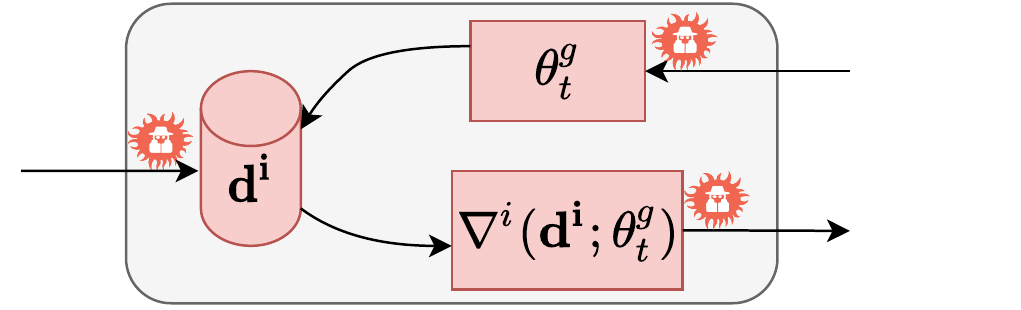}}\\ \cline{3-4}

& & \multirow{2}{2cm}{Data poison} & Adversary can only manipulate local data of the compromised clients; the clients use this data to compute their updates. Adversary does not break into the compromised clients.\\ 
& & & \multicolumn{1}{c|}{\includegraphics[scale=.5]{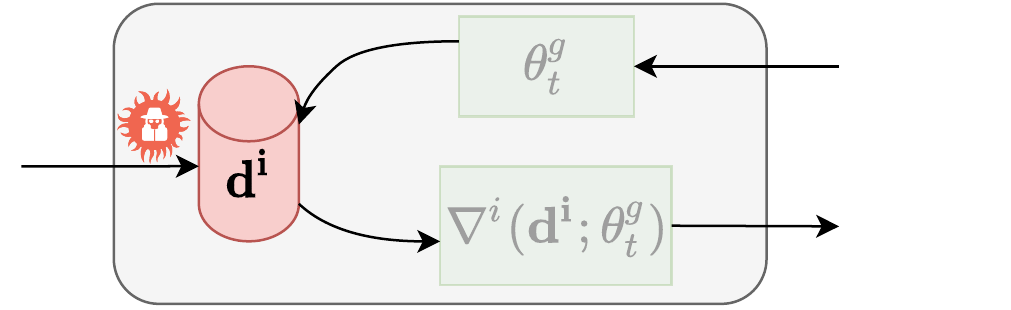}} \\ \cline{2-4}

% \multirow{8}{2cm}{\em \tabincell{c}{Capabilities in\\terms of frequency\\of attack\\(Attack mode)}} 
& \multirow{10}{2cm} {\em \tabincell{c}{Capabilities in\\terms of frequency\\of the attack\\(Attack mode)}} & \multirow{2}{2cm}{Online} & Adversary repeatedly and adaptively poisons the compromised clients during FL, e.g., model poisoning attacks~\cite{bhagoji2019analyzing,fang2020local,shejwalkar2021manipulating}. Impacts of these attacks can persist over the entire FL training.\\ 
&  & & \multicolumn{1}{c|}{\includegraphics[scale=.5]{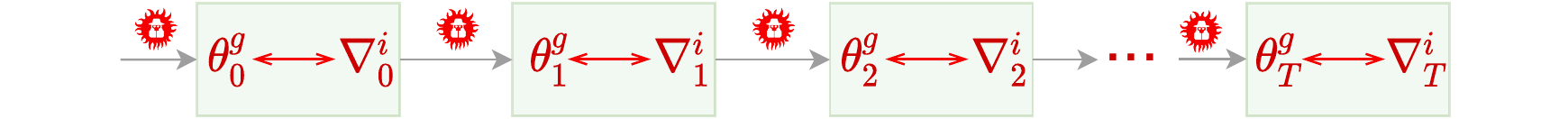}} \\ \cline{3-4}

& &  \multirow{2}{2cm}{Offline} & Adversary poisons the compromised clients only once at the beginning of FL, e.g., baseline label flipping attacks~\cite{fang2020local,wang2020attack}. Impact of such attacks may quickly fade away.
\\
&  & & \multicolumn{1}{c|}{\includegraphics[scale=.5]{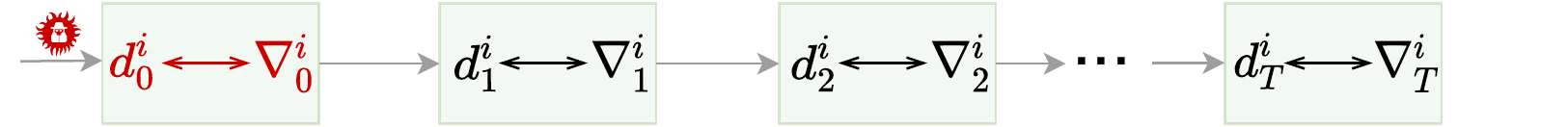}} \\ \hline

\end{tabular}
\vspace*{-1.75em}
\end{table*}

% \begin{figure}
% \hspace*{-1em}
% \centering
% \includegraphics[scale=.55]{figures/attacker_goals.pdf}
% \caption{Connection between previous taxonomies on attacks on the centralized learning and and the poisoning attacks on FL in existing literature.}
% \label{fig:attack_goal}
% \vspace*{-1.25em}
% \end{figure}

% \begin{table}
% % \vspace*{-0.5em}
% \caption{} \label{tab:attack_goal}
% \vspace*{-0.5em}
% \centering
% % \fontsize{9}{10}\selectfont{}
% \begin{tabular} {|c|c|c|} \hline
%  \multirow{2}{*}{Attack specificity} & \multicolumn{2}{c|}{Security violation} \\ \cline{2-3}
%   & Integrity & Availability \\ \hline
%  Targeted   &  & \tabincell{c}{Semantic backdoor\\Targeted} \\ \hline
%  Untargeted & Artificial backdoor & Untargeted \\ \hline
  
% \end{tabular}
% \vspace*{-1.25em}
% \end{table}

\subsection{Dimensions of Poisoning Threat to FL}\label{threat:dimensions}
In this section, we build on  previous systemization efforts for adversarial ML~\cite{barreno2010the,huang2011adversarial,munoz2017towards,biggio2018wild} and present three key dimensions for  the threat model of FL poisoning, as shown in  Table~\ref{table:poisoning_dimensions}. 

% \subsubsection{Adversary's Objective}\label{threat:dimensions_goal}

% The goal can be \emph{untargeted}, \emph{targeted}, or \emph{backdoor} poisoning. In \emph{untargeted} poisoning,  the adversary aims to make the global model misclassify every (or most) test inputs.

% In \emph{targeted} poisoning, the goal is to make the model misclassify a specific (and small) set of test inputs of adversary's choice. 
% % 
% In \emph{backdoor} poisoning, the goal is to make the model misclassify any test input with a specific trigger pattern of feature values. The trigger can be a pixel-pattern for image data and word-pattern for text data.  Backdoored inputs with a naturally present trigger are called \emph{semantic} backdoors~\cite{bagdasaryan2018how,wang2020attack}, while those with manually embedded trigger are called \emph{artificial} backdoors~\cite{xie2019dba}.
% % 
% In the artificial backdoor attacks, the adversary should modify the inputs at the inference time. 
% % 
% Finally, we note that, in targeted and backdoor attacks, the adversary can aim to misclassify the target inputs to a specific label~\cite{wang2020attack,fung2020limitations}, but this does not apply to untargeted poisoning.

\subsubsection{Adversary's Objective}\label{threat:dimensions_goal}
% \red{define three objective dims from previous works in your language; connect this systematization to the existing fl poisoning attacks; state that the taxonomy aims to define the intentions of the attacker and not the effect of attacker's actions, hence we are interested in 'availability untargeted' attacks that can last for long duration.}
Inspired by \cite{biggio2018wild}, we define three attributes of the adversary's objectives.
% , and then, elaborate on their connections with existing FL poisoning attacks from Figure~\ref{fig:attack_types}.
% \red{the attributes are exactly the same as in~\cite{biggio2018wild} so can be removed}

\paragraphbe{Security violation:} The adversary may aim to cause an \emph{integrity} violation, i.e., to evade detection without disrupting normal service operations, or an \emph{availability} violation, i.e., to compromise the service for legitimate users.

\paragraphbe{Attack specificity:} 
The attack is \emph{discriminate} 
if it aims to cause misclassification of a specific set/class of samples; it is \emph{indiscriminate} otherwise. 
% This attribute ranges from \emph{discriminate}\footnote{We use the term \emph{discriminate} instead of \emph{targeted} as in~\cite{biggio2018wild} to avoid confusion with existing \emph{targeted attacks}~\cite{bhagoji2019analyzing,tolpegin2020data} from FL poisoning literature.} to \emph{indiscriminate}, respectively, if the attack aims to cause misclassification of a specific set/class of samples (to target a given client), or of any sample (to target any client).

\paragraphbe{Error specificity:} This attribute is especially relevant in multi-class classification settings.
% to disambiguate the notion of misclassification. ?????
It is \emph{specific} if the attacker's goal is to have a sample misclassified as a specific class; the attack is \emph{generic} if the attacker does not care about the wrong label assigned to the misclassified samples.

% \paragraphb{\em Connections between the characteristics and the existing attacks.}
% Next, we extend the recent taxonomy~\cite{jere2020taxonomy} and illustrate the major approaches to FL poisoning from FL literature in Figure~\ref{fig:attack_types}. \red{this sentence here is weird! does not connect }
% 
% \red{the following does not fit in this section. maybe make its new section of "poisoning types"}

\paragraphe{\underline{Adversary objectives in different classes of poisoning:}}
Here, based on the above taxonomy, we discuss the adversary's objective for different types of poisoning attacks (Figure~\ref{fig:attack_types}).

\paragraphe{Targeted attacks~\cite{bhagoji2019analyzing,tolpegin2020data}}  aim to misclassify  specific sets/classes of input, hence they are ``discriminate.'' Such discriminate attacks can be either used for  ``integrity'' or ``availability'' violations, depending on how the poisoned data is used.
% If the targeted adversary uses target samples during training~\cite{bhagoji2019analyzing} this is an ``integrity'' violation, otherwise~\cite{sun2019can} it is an ``availability'' violation.\red{what??}\virat{i need to explain this to you and will modify it based on suggestions; this is very debatable i feel}

\paragraphe{Semantic backdoor attacks~\cite{bagdasaryan2018how,wang2020attack}} have the same goal as the targeted attacks, but the targeted inputs should have specific properties, e.g., a pixel pattern or a word sequence. 
% They cannot be completely arbitrary as in  targeted attacks. 
Hence, these are ``discriminate,''  ``availability'' or ``integrity'' attacks.

\paragraphe{Artificial backdoor attacks~\cite{xie2019dba}} aim to misclassify \emph{any} input containing a backdoor \emph{trigger}, hence these attacks are ``indiscriminate'' attacks.
Note that, such test inputs should be modified to have the backdoor trigger and only the adversary or a malicious client know the trigger. Hence, these attacks aim to evade the detection, i.e., cause an integrity violation. Hence, these are ``integrity indiscriminate'' attacks.

\paragraphe{Untargeted attacks~\cite{shejwalkar2021manipulating,fang2020local,baruch2019a}} aim to misclassify any test input, i.e., they are ``indiscriminate'' attacks. But, test inputs need not be modified in order to misclassify. Hence, these are ``availability'' attacks.

Finally, the error specificity of each of these attacks can be either ``specific'' or ``generic.''

% Note that, the objective represents the adversary's intentions, and not the attacks' effects. For instance, all untargeted attacks in poisoning ML/FL literature~\cite{shejwalkar2021manipulating,fang2020local,munoz2017towards,biggio2012poisoning} are \emph{indiscriminate availability} attacks. But, their effect does not always reflect the objective, i.e., they do not always succeed at misclassifying each test input.\red{i think remove this par}

\paragraphbe{\underline{Focus of our work:}}
In this work, {\em we focus on  untargeted attacks}, i.e., indiscriminate availability attacks with generic error specificity, for the following reasons.
% We focus on untargeted attacks because: (1) they aim to impact all FL clients and pose a great threat; (2) their resulting effect can be reduction of the accuracy by only a few percentages (which can be significant in production FL), and hence, they may remain undetected for long duration; (3) they are never studied in production FL environments unlike backdoor/targeted attacks.

% \red{we need to improve this part. add a ref to works that say backdoor/targeted impact fewer users (peter's work, prateeks video that you sent). we have to say WHY reducing accuracy by a few percentages is still important. maybe add the example of competitor service providers. i thnk we had some discussion in previous III.A. state that untargeted is likely more challenging. Note that this section is important to reader, otherwise the paper will look not motivated}

\paragraphe{{Untargeted attacks pose a great threat to production FL:}} 
% While targeted and backdoor attacks lead to  models that misclassify a limited set of target input samples/classes, untargeted attacks are designed to impact all clients and all test inputs.
% Also, the extensively studied backdoor attacks require to  poison input samples in addition to poisoning the  model (in the case of ``Artificial'' backdoors, Figure~\ref{fig:attack_types}) or will only be able to corrupt  input samples that contain specific patterns (in the case of ``Semantic'' backdoors, Figure~\ref{fig:attack_types}).
% 
Untargeted attacks are designed to impact all clients and all test inputs. For instance, FL on FEMNIST  achieves an 85\%~\cite{reddi2020adaptive} accuracy in a benign setting, and untargeted attacks reduce the accuracy to, e.g., $[78, 82]\%$ depending on the percentages of compromised clients. Such an accuracy drop is significant for production FL, as \emph{a malicious service provider can gain advantage over their competitors by causing such small, yet noticeable, accuracy reductions in the competing services} and \emph{such small accuracy reductions can impact most clients and data from all classes in arbitrary fashion}.

\paragraphe{{Untargeted attacks can go undetected for long duration:}} As discussed above, the untargeted attack aims at reducing the  overall accuracy of the global model, even by only a few percentage points. 
Such a small reduction in accuracy is hard to detect in practical settings due to the absence of reliable benchmarks for the target application.
% , e.g., it is hard to detect that a 94\% accuracy is reduced from.
For instance, the affected service provider will never know that they could have achieved an 85\% accuracy and will believe that $[78, 82]\%$ is the highest achievable accuracy.

% By contrast, a (successful) targeted or backdoor attack~\cite{bhagoji2019analyzing,bagdasaryan2018how,sun2019can,wang2020attack} aims at reducing the accuracy of global model on \emph{specific} samples by \emph{large margins}. These attacks misclassify only the samples with specific properties (e.g., a backdoor trigger~\cite{bagdasaryan2018how,xie2019dba}, an underrepresented class~\cite{wang2020attack}, or just a few samples~\cite{bhagoji2019analyzing,sun2019can}), hence they are more localized and can be easily detected. For instance, backdoor attacks substantially change the values of a small set of neurons of target model, and hence, can be detected by probing the neurons and reverse engineering the trigger, e.g., using Neural Cleanse~\cite{wang2019neural} or ABS~\cite{liu2019abs} defenses. By contrast,  untargeted attacks cannot be detected this way as they change all the neurons of the target model~\cite{baruch2019a,fang2020local,shejwalkar2021manipulating}. 

\paragraphe{{Constructing untargeted attacks is more challenging:}} Untargeted attacks aim to solve a more challenging problem, which is  affecting arbitrary test inputs. 
% This is in part why existing works on untargeted poisoning~\cite{fang2020local,shejwalkar2021manipulating,tolpegin2020data} use unrealistic parameters in the assessment of their attacks. 
% \paragraphe{\underline{They are  underexplored:}} 
However, while there exist several defenses to protect FL against untargeted poisoning~\cite{blanchard2017machine,mhamdi2018the,shejwalkar2021manipulating,yin2018byzantine},  \emph{these attacks are not studied under production FL environments} (as discussed later on). 
% On the other hand, targeted attacks have been studied in production FL environments~\cite{bhagoji2019analyzing,bagdasaryan2018how,sun2019can,wang2020attack,xie2019dba,tolpegin2020data}  (Table~\ref{tab:practical_ranges}).

% , as shown in Table~\ref{tab:practical_ranges}.

\subsubsection{Adversary's Knowledge}
Below we elaborate on two dimensions of adversary's knowledge: knowledge of the global model and knowledge of the data from the benign distribution.

\paragraphbe{Knowledge of the global model:} This can be \emph{nobox} or \emph{whitebox}. In the nobox case, the adversary does not know the model architecture, parameters, or outputs, and is the most practical setting in FL~\cite{kairouz2019advances}, e.g., the data poisoning adversary has nobox knowledge of the global model. In the whitebox case, the adversary knows the global model parameters and outputs, whenever the server selects at least one compromised client. The model poisoning adversary always has whitebox knowledge of the global model. As we will explain in Section~\ref{threat:practical_ranges}, this is a relatively less practical setting in FL, as it assumes complete control of the compromised devices.

\paragraphbe{Knowledge of the data from benign distribution:} This can be \emph{full} or \emph{partial}. In full knowledge case, the adversary can access the benign local data of compromised as well as benign clients. In partial knowledge case, the adversary can access the benign local data only of the compromised clients. 
% Previous works on untargeted poisoning in FL~\cite{shejwalkar2021manipulating,fang2020local,baruch2019a} commonly consider the full knowledge case, but it is less practical, as accessing all client devices in production FL is not possible. 
We only consider the partial knowledge case, because accessing the data of all the clients is impractical in production FL.

% We discuss the other relevant parameters of the adversary's knowledge, e.g., data from benign distribution, in Section~\ref{threat:practice}.

\subsubsection{Adversary's Capability}
Below, we elaborate on the the adversary's capability in terms of \emph{access to client devices} and \emph{frequency of attack}, i.e., the \emph{attack mode}.

\paragraphbe{Capability in terms of access to client devices:} Based on the FL stages (part of FL pipeline on client device) that the adversary poisons, there can be a \emph{model poisoning adversary} or a \emph{data poisoning adversary}. The model poisoning adversary can break into a compromised device (e.g., by circumventing the security protocols of operating systems such as Android) and can \emph{directly manipulate} the poisoned updates~\cite{fang2020local,baruch2019a,shejwalkar2021manipulating,blanchard2017machine,mhamdi2018the,pillutla2019robust}.
This adversary can craft highly effective poisoned updates, but due to unreasonable amount of access to client devices, it can compromise very small percentages of FL clients~\cite{kairouz2019advances,fed_learning_workshop}.

On the other hand, a data poisoning adversary cannot break into a compromised device and can only poison its local dataset. The compromised clients use their local poisoned datasets to compute their poisoned updates, hence this adversary \emph{indirectly manipulates} the poisoned updates.
Due to the indirect manipulation, these updates may have less poisoning impact than the model poisoning updates. But, due to the limited access required to the compromised clients, this adversary can compromise relatively large percentages of FL clients~\cite{kairouz2019advances,fed_learning_workshop}.

% In our work, we consider that the adversary does not tamper with the inputs at the inference time, because an important goal of poisoning in FL is to reduce the performance of the global model for benign clients

% \subsubsection{Attack Mode}

\paragraphbe{Capability in terms of attack frequency (Attack mode):}
The mode of poisoning attacks on FL can be either \emph{offline} or \emph{online}. In the offline mode, the adversary poisons the compromised clients only once before the start of FL training, e.g., the baseline label flip attack~\cite{fang2020local} flips the labels of data of compromised clients once before the FL training starts. In the online mode, the adversary \emph{repeatedly} and \emph{adaptively} poisons the compromised clients, e.g., existing model poisoning attacks~\cite{shejwalkar2021manipulating,baruch2019a} repeatedly poison the updates of compromised clients selected by the server.

Finally, we assume that the compromised clients can collude to exchange their local data and model updates in order to increase impacts of their attacks.

\begin{table*}
\caption{Practical ranges of FL parameters based on the literature and discussions on FL production systems~\cite{fed_learning_workshop,kairouz2019advances,bonawitz2019towards} and the ranges used in \emph{untargeted} FL poisoning and robust AGRs literature~\cite{fang2020local,baruch2019a,shejwalkar2021manipulating,blanchard2017machine,mhamdi2018the}. MPA means model poisoning attack and DPA means data poisoning attack. Red (green) cells denote impractical (practical) ranges.} \label{tab:practical_ranges}
\vspace*{-.5em}
\centering
% \fontsize{9}{9}\selectfont{}
\setlength{\extrarowheight}{0.03cm}
\begin{tabular} {|c|c|c|}
  \hline
\bf \tabincell{c}{Parameters/Settings} & \bf  What we argue to be practical  & \bf  \tabincell{c}{Used in previous\\ \emph{untargeted} works} \\ \hline
  
  \multirow{2}{*}{\tabincell{c}{FL type + Attack type}} & \multirow{2}{*}{\tabincell{c}{Cross-silo + DPAs\\ Cross-device + \{MPAs, DPAs\}}} & \cellcolor{red!25} \\ 
  & & \multirow{-2}{*}{\cellcolor{red!25}{Cross-silo + MPAs}} \\ \hline
  
  \multirow{2}{*}{\tabincell{c}{Total number of FL\\ clients, $N$}} & \multirow{2}{*}{\tabincell{c}{Order of  $[10^3,10^{10}]$ for cross-device\\ $[2, 100]$ for cross-silo}} & \cellcolor{red!25} \\ 
  &  & \multirow{-2}{*}{\cellcolor{red!25}[50, 100]} \\ \hline
  
  \multirow{2}{*}{\tabincell{c}{Number of clients\\ chosen per round, $n$}} & \multirow{2}{*}{\tabincell{c}{Small fraction of N for cross-device\\ All for cross-silo}} & \cellcolor{red!25} \\
  & & \multirow{-2}{*}{\cellcolor{red!25}All} \\ \hline
  
  \tabincell{c}{\% of compromised\\ clients, $M$} & \tabincell{c}{$M\leq$0.1\% for DPAs\\ $M\leq$0.01\% for MPAs} & \cellcolor{red!25}$[20, 50]\%$  \\ \hline
  
  \tabincell{c}{Average size of benign\\ clients' data, $|D|_\mathsf{avg}$} & \tabincell{c}{$[50, 1000]$ for cross-device\\ Not applicable to cross-silo} & \tabincell{c}{Not studied for cross-device\\$[50,1000]$ for cross-silo} \\ \hline
  
  \tabincell{c}{Maximum size of\\ local poisoning data} & \tabincell{c}{Up to $100\times |D|_\mathsf{avg}$ for DPAs\\ Not applicable to MPAs} & \cellcolor{green!25} $\sim |D|_\mathsf{avg}$ \\
  \hline
\end{tabular}
\vspace*{-1.75em}
\end{table*}

\subsection{Practical Considerations for Poisoning Threat Models}\label{threat:practical_consideration}

\begin{comment}
\red{significance of untargeted compared to backdoor? targeted attacks work; why care about untargeted in practice? 1) with backdoors one can only poison samples with trigger, so poisoning happens in training and test. But with targeted/untargeted poisoning happens at training only. There are scenarios in which untargeted/targeted can perform but backdoors cannot. 2) accuracy of model can reduce without server knowing.}
\end{comment}

% \red{i moved this here. double check}\virat{in the background (section II.A), we will have to repeat what is cross-silo, cross-device etc. it is not there anymore if we move this part here}

\subsubsection{Salient Features of Production Federated Learning}\label{threat:production_env}
Production FL can be either \textbf{cross-device} or \textbf{cross-silo}  \cite{kairouz2019advances}. 
In \emph{cross-device FL}, the number of clients ($N$) is large (from few thousands to billions) and only a small fraction of them is chosen in each FL training round, i.e., $n\ll N$. 
In cross-device FL, clients' devices are highly resource constrained, and therefore, they can process only a limited amounts of data in an FL round. 
% But, the clients have complete control on the data they inject into their devices, e.g., clients can insert completely garbage data and the FL application will still process it. 
Also, as the devices have highly unreliable network connections, it is expected that a small fraction of the selected devices may drop out in any given FL round. Note that, this equally impacts both benign and compromised clients and does not affect the robustness; this is similar to how the choice of $n$ has no impact on the robustness (Section~\ref{exp:round_nclients}). 
In \emph{cross-silo FL}, $N$ is moderate (up to 100) and all clients are selected in each round, i.e., $n=N$. Clients are large corporations, e.g., banks, and have devices with ample resources. Hence, they can process very large amounts of data and client drop-outs do not happen. 
% But, the compromised users who contribute poisoned data to clients do not have complete control over the data, because the poisoned data may be screened by the clients before using in FL training.

% 
In both FL types, the on-device model used for inference and the on-device model being trained are different. Hence, an adversary cannot gain any insight into the training-model by querying the inference-model, i.e., nobox access (Table~\ref{table:poisoning_dimensions}), and must break into the device, i.e., get whitebox access (Table~\ref{table:poisoning_dimensions}).

Finally, we assume that production systems are adequately protected against standard attack vectors and vulnerabilities such as Sybil attacks. 
For instance, if the adversary manages to operate millions of fake accounts~\cite{fake_acc_1}, we argue that the service provider should prioritize improving their security attestation protocols instead of deploying FL.
Section~\ref{threat:practical_ranges} also explains that the cost of operating a large scale, persistent botnet in modern operating systems, e.g., Android, is non-trivial.
Please refer to~\cite{kairouz2019advances} for more details on production FL.

\subsubsection{Understanding the practicality of threat models}\label{threat:justification}

For {our goal of untargeted poisoning with the partial knowledge of the benign data},  we can combine the rest of the dimensions in Table~\ref{table:poisoning_dimensions} and obtain eight possible threat models (Table~\ref{tab:threat_combinations}). We argue that only T4 (nobox offline data poison) and T5 (whitebox online model poison) are of practical value, and below, justify why other models are less relevant in practice: \noindent \textbf{(1)} With model poisoning capability,
the adversary has whitebox access by default, hence, T1 and T2 in Table~\ref{tab:threat_combinations} are not valid.
\noindent \textbf{(2)} In cross-device FL, only a few selected clients get the most recent global model in each round. Hence, to gain whitebox access to the model, the adversary needs to control (i.e., break into) a large number of  devices (so that in most FL rounds, the FL server picks at least one of them), which is impractical in practice as we explain in Section~\ref{threat:practical_ranges}. With whitebox access, the adversary can mount the stronger online model poisoning attacks (MPAs) instead of data poisoning attacks (DPAs). Therefore, T3, T7, and T8 are not reasonable threat models, as they combine whitebox access with either offline attacks or DPAs. 
% Note that, as argued before, whitebox access in cross-silo FL is not practical.
%
\noindent \textbf{(3)} Under T6 (nobox online data poison), the adversary mounts an online attack, i.e., they \emph{adaptively} poison the local data of compromised clients. But, as the adversary has no knowledge of the (current) global model due to nobox access, they cannot generate new poisoning data adaptively. Hence, the combination of nobox and online is not practical.

\begin{table}
% \vspace*{-0.5em}
\caption{The eight possible threat models for \emph{untargeted poisoning attacks} on FL. T3-T8 are valid, but only T4 and T5 represent practical FL deployments (Section~\ref{threat:practice}).
% Only six (T3-T8) of these are valid and only two (T4, T5 in dark green cells) represent practical FL deployments (Section~\ref{threat:practice}).
} \label{tab:threat_combinations}
\vspace*{-0.5em}
\centering
% \fontsize{9}{10}\selectfont{}
\begin{tabular} {|c|c|c|c|}
  \hline
  & Capability & Knowledge & Attack mode \\ 
  & $\in\{\texttt{MP},\texttt{DP}\}$ & $\in\{\texttt{Nb},\texttt{Wb}\}$ & $\in\{\texttt{Off},\texttt{On}\}$ \\ \hline
  T1 & Model poison & Nobox & Offline \\ \hline
  T2 & Model poison & Nobox & Online \\ \hline
  \cellcolor{green!10}T3 & \cellcolor{green!10}Model poison & \cellcolor{green!10}Whitebox & \cellcolor{green!10}Offline \\ \hline
  \cellcolor{green!40}T4 & \cellcolor{green!40}Model poison & \cellcolor{green!40}Whitebox & \cellcolor{green!40}Online \\ \hline
  \cellcolor{green!40}T5 & \cellcolor{green!40}Data poison & \cellcolor{green!40}Nobox & \cellcolor{green!40}Offline \\ \hline
  \cellcolor{green!10}T6 & \cellcolor{green!10}Data poison & \cellcolor{green!10}Nobox & \cellcolor{green!10}Online \\ \hline
  \cellcolor{green!10}T7 & \cellcolor{green!10}Data poison & \cellcolor{green!10}Whitebox & \cellcolor{green!10}Offline \\ \hline
  \cellcolor{green!10}T8 & \cellcolor{green!10}Data poison & \cellcolor{green!10}Whitebox & \cellcolor{green!10}Online \\ \hline
\end{tabular}
\vspace*{-1.85em}
\end{table}

\begin{comment}
\red{near practical ranges table discuss: 1) why targeted works use practical ranges but untargeted dont 2) (somewhat related to 1) why techniques of targeted attacks cannot be extended to untargeted? point-to-make: targeted and untargeted are different, xxx}
\red{community knows efficient alg for targeted but not for untargeted. reasons specially for backdoor is the use of trigger.}
\end{comment}

% \vspace*{-1em}
\subsubsection{Practical Ranges of FL Parameters}\label{threat:practical_ranges}
% We argue that, the main body of FL robustness literature~\cite{baruch2019a,fang2020local,shejwalkar2021manipulating,blanchard2017machine,mhamdi2018the} completely ignores evaluating their attacks and/or defenses using FL parameters of production FL systems. 
% \red{The literature on measuring FL robustness using poisoning attacks can be broadly divided in targeted/backdoor and untargeted attacks.
We argue that the literature on untargeted poisoning~\cite{baruch2019a,fang2020local,shejwalkar2021manipulating,blanchard2017machine,mhamdi2018the} rarely evaluates their proposed attacks/defenses for the production FL settings, primarily due to their motivation to perform worse-case analyses. But, we show that such analyses lead to conclusions that do not apply to production FL.

Table~\ref{tab:practical_ranges} demonstrates the stark differences between the parameter ranges used in the untargeted poisoning literature and their practical ranges, which we have  obtained from recent surveys~\cite{kairouz2019advances,bonawitz2019towards} and discussion among FL experts~\cite{fed_learning_workshop}. 
% On the other hand, the targeted/backdoor literature (Table~\ref{tab:practical_ranges}) generally uses practical ranges. 
This is due to the more challenging nature of untargeted poisoning in FL.
We attribute this to the difficulty of establishing successful untargeted attacks for practical settings, as we will also show in our evaluations.

% The most notable and widespread mistakes include using very high percentages of compromised clients (up to 25\% and 50\% in attacks and defenses works, respectively) and mounting model poisoning attacks on cross-silo FL.

% To address this, we extensively review the real-world FL systems from the literature~\cite{kairouz2019advances,bonawitz2019towards} and experts' discussions~\cite{fed_learning_workshop}; 
% \red{To demonstrate this fallacy, we extensively review the real-world FL systems from the literature~\cite{kairouz2019advances,bonawitz2019towards} and experts' discussions~\cite{fed_learning_workshop}; in Table~\ref{tab:practical_ranges} and~\ref{tab:practical_ranges_targeted}, we list practical ranges for some of the most important parameters that we encounter in FL evaluations. Table~\ref{tab:practical_ranges} gives the ranges of these parameters that previous untargeted works use, while Table~\ref{tab:practical_ranges_targeted} gives the ranges that previous targeted/backdoor works use.}
% % 

% The most  widespread mistakes include using very high percentages of compromised clients and mounting model poisoning attacks only on cross-silo FL (Table~\ref{tab:practical_ranges}).
Contrary to what production FL settings encounter, previous works commonly  evaluate robustness using very high percentages of compromised clients and/or using model poisoning attacks on cross-silo FL (Table~\ref{tab:practical_ranges}).
However, we use small percentages of compromised clients $M\leq$1, for cross-device FL, use large numbers of clients $N\in$[$1,000$, $34,000$] and use $n\in[25, 50]\ll N$ in each round; we use $N$=$n$=50.

In particular, consider the percentages of compromised clients; state-of-the-art attacks~\cite{fang2020local,baruch2019a,shejwalkar2021manipulating} (defenses~\cite{blanchard2017machine,yin2018byzantine,xie2018generalized,chen2018draco}) assume adversaries who can compromise up to 25\% (50\%) of FL clients.
% While the state-of-the-art defenses~\cite{blanchard2017machine,yin2018byzantine,xie2018generalized,chen2018draco} aim to defend against the adversaries who can compromise up to 50\% of FL clients.
% 
The cost of creating and operating a compromised client botnet at scale (which includes breaking into devices)  is non-trivial.
To create the botnet, the adversary would need to either buy many physical devices ($\sim$\$25 each) and root them (for state-of-the-art model poisoning attacks~\cite{fang2020local,baruch2019a,shejwalkar2021manipulating}), pay for access to large but undetected botnets with remote administrative access, or develop an entirely new botnet via compromising a popular app/sdk to exploit unpatched security holes and gain persistence.
To operate the botnet, the adversary must avoid detection by antimalware services~\cite{google_play_protect} as well as dynamic anti-abuse services (such as Android's SafetyNet~\cite{android_safetyney}).
With a botnet in place, the adversary may further need to pay for a skilled engineering team to keep malicious FL code in sync with the target FL-enabled app and to reverse-engineer frequently-shifting ML workloads.
Such an engineering team could instead change apps' behaviors to mimic the effect of a compromised FL-trained model, they might use their privileged access to steal login credentials for account hijacking, or they might participate in ad/click fraud or bank fraud or ransomware for financial gain. More plausible scenarios for an adversary reaching double-digit client percentages\textemdash such as an app insider\textemdash likely enable attacker-controlled FL servers, thereby removing them from the literature's standard threat model.

For data poisoning attacks, we assume that compromised clients can have a limited amount of poisoned data $D_p$. Because, in cross-device FL, the devices with low processing powers (e.g., smart phones and watches) can process limited $D_p$ in the short duration of FL rounds. 
However, in cross-silo FL, silos can inspect $D_p$ and remove $D_p$ with sizes much larger than the average size of clients' data $|D|_\mathsf{avg}$. 
Hence, we argue that $|D_p|$ should be up to $100\times|D|_\mathsf{avg}$. 
We discuss rest of the parameters from Table~\ref{tab:practical_ranges} in the corresponding sections.

\subsection{Threat Models in Practice}\label{threat:practice}
Here we discuss the two threat models of practical interest.

\subsubsection{Nobox Offline Data Poisoning (T4)}\label{threat:practice_dp}

In this setting, the adversary does not know the architecture, parameters, or outputs of the global model. The adversary knows the server's AGR, but may or may not know the global model architecture; we evaluate both cases. We assume that the adversary knows the benign data of the compromised clients and mounts offline data poisoning attacks (DPAs).
% , i.e., poisons the local datasets of compromised devices only once before the FL training starts.

This adversary does not require any access to the internals (e.g., FL binaries, memory) of compromised devices, and therefore, can compromise large percentages of production FL clients, e.g., on order of up to 0.1\%~\cite{kairouz2019advances,fed_learning_workshop}. However, the poisoning impact of the corresponding poisoned updates is very limited. This is partly because arbitrarily poisoned updates (e.g., of model poisoning attacks (MPAs)~\cite{fang2020local,baruch2019a,shejwalkar2021manipulating}) need not map to the valid data domain.
For instance, consider the standard \emph{max} function:  $f(x,y)$=$\text{max}(x,y)$. Gradient of this function with respect to either $x$ or $y$ is always 0 or 1~\cite{cs231n_backprop}.
% For instance, if a certain global model parameter is zero (e.g., $cw=0$), the gradient of the loss of valid data with respect to the parameter will be the constant $c$ (=$\frac{\partial cw}{\partial w}$). 
Hence, a DPA cannot have a poisoned update with an arbitrary value for gradients of the parameters. But an MPA can, because it can directly assign any arbitrary value to the parameters' gradients.

\subsubsection{Whitebox Online Model Poisoning (T5)}\label{threat:practice_mp}

The adversary knows the parameters and predictions of the global model whenever the server selects at least one compromised client.
We assume that the adversary knows the server's aggregation rule and the benign data on the compromised devices. The adversary mounts  online MPAs.
% , i.e., directly poisons the model updates of compromised clients each time the server selects them.

Unlike data poisoning adversary, this adversary breaks into the compromised devices, which is extremely costly as discussed in Section~\ref{threat:practical_ranges}. 
% e.g., by circumventing the security firewall of the client's operating system such as Android. 
% 
% Circumventing the security protocols of modern operating systems is a difficult task and requires advanced skill sets. Furthermore, compromising a client for whitebox access has high costs, because the adversary should either own the compromised device or pay its owner.
% 
Hence, in practice, a model poisoning adversary can compromise very small percentages of FL clients, e.g., on order of up to 0.01\%~\cite{kairouz2019advances,fed_learning_workshop}.  However, due to their ability to directly manipulate the model updates, in theory, a model poisoning adversary can craft highly poisonous updates. We can justify this claim from the example of a zero-value parameter discussed in Section~\ref{threat:practice_dp}.

\section{Exploring the Space of FL Poisoning Attacks} \label{method}

% \red{In this section, we discuss existing data (DPAs) and model (MPAs) poisoning attacks, and then present our improved DPAs and MPAs.}\virat{can be removed}

% \vspace*{-.75em}
\subsection{Existing FL Poisoning Attacks }\label{existing}

\subsubsection{Data Poisoning Attacks (DPAs)}\label{existing_dp}
DPAs have been studied mainly for centralized ML~\cite{xiao2015support,xiao2012adversarial,yang2017generative,munoz2019poisoning,chen2017targeted}, and no prior work has studied untargeted DPAs that are tailored to FL settings. Fang et al.~\cite{fang2020local} show the possibility of applying simple label flipping attacks to FL, where each compromised client flips the labels of their data from true label $y\in[0,C-1]$ to false label $(C-1-y)$ if $C$ is even and to false label $(C-y)$ if $C$ is odd, where $C$ is the number of classes.
% \vspace*{-.2em}

\subsubsection{Model Poisoning Attacks (MPAs)}\label{existing_mp}
% Multiple works have proposed MPAs on FL~\cite{fang2020local,baruch2019a,shejwalkar2021manipulating}. 
These consider our whitebox online model poisoning threat model (T4) from Section~\ref{threat:practice_mp}).
% but use unrealistic FL parameter values, e.g., very high percentages of compromised clients.

\noindent\textbf{\em Little Is Enough (LIE)}  attack~\cite{baruch2019a} adds small amounts of noise to each dimension of the average of the benign updates. Specifically, the adversary computes the average ($\nabla^b$)  and  the standard  deviation ($\sigma$) of the available benign updates; then computes a coefficient $z$ based on the number of benign and  compromised  clients;  and  finally  computes  the  poisoned update as $\nabla'=\nabla^b + z\sigma$.
\cite{baruch2019a} shows that such noises easily evade the detection by robust AGRs as well as effectively poison the global model.

\noindent\textbf{\em Static Optimization (STAT-OPT)}  attack~\cite{fang2020local} proposes a general FL  poisoning framework and then tailors it to specific AGRs. STAT-OPT computes the average ($\nabla^b$) of the available benign updates and computes a \emph{static malicious direction}, $\omega = -sign(\nabla^b)$; the final poisoned update, $\nabla'$, is $-\gamma\omega$ and the attack finds a suboptimal $\gamma$ that circumvents the target AGR; for details please refer to~\cite{fang2020local}.
Unlike LIE, STAT-OPT attacks carefully tailor themselves to the target AGR, and hence, perform better.

\noindent\textbf{\em Dynamic Optimization (DYN-OPT)} attack~\cite{shejwalkar2021manipulating} proposes a general FL poisoning framework and then tailors it to specific FL settings. DYN-OPT computes an average of the available benign updates, $\nabla^b$,  and perturbs it in a \emph{dynamic, data-dependent malicious direction} $\omega$ to compute the final poisoned update $\nabla'= \nabla^b + \gamma \omega$. DYN-OPT finds the largest $\gamma$ that successfully circumvents the target AGR. 
DYN-OPT is much stronger, because unlike STAT-OPT, it finds the largest $\gamma$ and uses a dataset tailored $\omega$.

\subsection{Our Improved FL Poisoning Attacks}\label{improved}

We first present a general optimization problem to model FL poisoning attacks. Then we use it to design improved poisoning attacks on state-of-the-art AGRs from Section~\ref{background:agr}. 
% Due to space restrictions, we defer design details of our attacks on Mkrum and Trmean to Appendix~\ref{appdx:missing_attacks}.

\subsubsection{Formulating FL Poisoning as an Optimization Problem}\label{improved:general_opt}
Our optimization problem for poisoning attacks is based on that of~\cite{shejwalkar2021manipulating}.
Specifically, we aim to craft poisoned updates (via data or model poisoning) which will increase the overall distance between the poisoned aggregate (computed using poisoned and benign updates) and the benign aggregate (computed using only benign updates).
This can be formalized as follows:
\begin{align}\label{eq:gen_opt}
&\underset{\nabla'\in R^d}{\text{argmax}}\quad \Vert\nabla^b - \nabla^p\Vert \\
\text{...} \nabla^b = &f_\mathsf{avg}(\nabla_{i\in\{[n']\}}),
\ \ \nabla^p = f_\mathsf{agr}(\nabla'_{\{i\in[m]\}}, \nabla_{\{i\in[n']\}})\nonumber
\end{align}
where, $m$ is the number of compromised clients selected in the given round, $f_\mathsf{agr}$ is the target AGR, $f_\mathsf{avg}$ is the Average AGR, $\nabla_{\{i\in[n']\}}$ are the benign updates available to the adversary (e.g., updates computed using the benign data of compromised clients), $\nabla^b$ is a reference benign aggregate, and $\nabla'_{\{i\in[m]\}}$ are $m$ replicas of the poisoned update, $\nabla'$, of our attack. $\nabla^p$ is the final poisoned aggregate.

Although our optimization problem in~\eqref{eq:gen_opt} is same as~\cite{shejwalkar2021manipulating}, \emph{two key differences from~\cite{shejwalkar2021manipulating} are:} \textbf{(1)} We are the first to use~\eqref{eq:gen_opt} to construct systematic data poisoning attacks on FL. \textbf{(2)} Our model poisoning attacks not only tailor the optimization in~\eqref{eq:gen_opt} to the given AGR (as in~\cite{shejwalkar2021manipulating}), but also to the given dataset and global model, by using stochastic gradient ascent algorithm (Section~\ref{improved:mp}); this boosts the efficacy of our attack.

\subsubsection{Our Data Poisoning Attacks (DPAs)}\label{improved:dp}
We formulate a general DPA optimization problem using~\eqref{eq:gen_opt} as follows:
\begin{align}\label{eq:gen_opt_dp}
&\underset{D_p\subset \mathcal{D}}{\text{argmax}}\quad \Vert\nabla^b - \nabla^p\Vert\\
\text{...} \nabla^b &\text{ and } \ \nabla^p \text{ as in~\eqref{eq:gen_opt} }
\text{and   }\nabla' = A(D_p, \theta^g)-\theta^g \nonumber
\end{align}
where $\mathcal{D}$ is the entire input space and $D_p$ is the poisoning data used to compute the poisoned update $\nabla'$ using a training algorithm $A$, e.g., mini-batch SGD, and global model $\theta^g$.
The rest of the notations are the same as in~\eqref{eq:gen_opt}.
To solve~\eqref{eq:gen_opt_dp}, we find $D_p$ such that when $\theta^g$ is fine-tuned using $D_p$, the resulting model $\theta'$ will have high cross-entropy loss on some benign data $D_b$ (e.g., that of compromised clients), i.e.,  high $L(D_b; \theta')$, and the corresponding update $\nabla'=\theta'-\theta^g$ will circumvent the target AGR. 
Our intuition is that, when the global model is updated using such $\nabla'$, it will have high loss on benign data~\cite{biggio2012poisoning,jagielski2018manipulating,munoz2017towards}.

% One way to compute such $D_p$ is via backgradient optimization based DPA~\cite{munoz2017towards}. However, Fang et al.~\cite{fang2020local} show that such DPAs are neither practical nor effective, even for extremely simple FL settings: It takes \emph{10 days} to generate $D_p$ of size 240 against logistic regression model trained using 100 clients, each with 60 MNIST samples; the attack impact is same as that of simple label flip attack.
Sun et al.~\cite{sun2021data} propose DPAs on federated multi-task learning where each client learns a different task. Hence, their attacks are orthogonal to our work.
On the other hand, as \cite{fang2020local} demonstrates, backgradient optimization based DPAs~\cite{munoz2017towards} are computationally very expensive ($\sim$\emph{10 days} to compute poison for a subset of MNIST task) yet ineffective.

\begin{figure}
\hspace*{-1em}
\centering
\includegraphics[scale=.55]{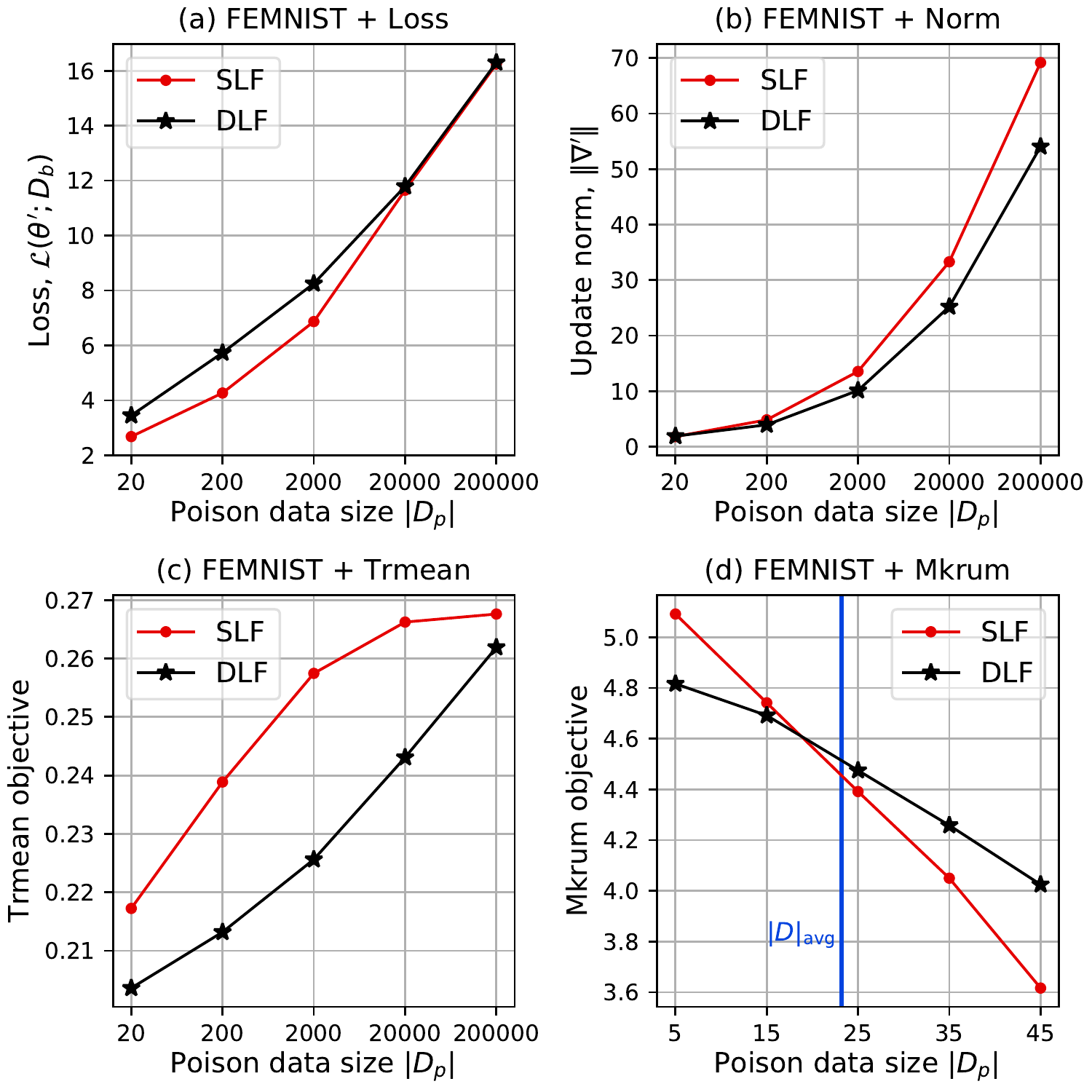}
\vspace*{-.5em}
\caption{Effect of varying the sizes of poisoned data, $D_p$, on the objectives of DPAs (Section~\ref{improved:dp}) on various AGRs.
We compute $D_p$ by flipping the labels of benign data.}
\label{fig:femnist_obj}
\vspace*{-1.75em}
\end{figure}

Instead, because the central server has no visibility into the clients' data or their sizes, we propose to use an appropriate amount of label flipped data as $D_p$ for each of the compromised clients.
Our intuition behind this approach is the same as before: the larger the amount of label flipped data used to compute $\theta'$, the larger the $L(D_p; \theta')$ and $\Vert\nabla'\Vert$, and therefore, the higher the deviation in~\eqref{eq:gen_opt_dp}.
We validate this intuition using FEMNIST dataset in Figure~\ref{fig:femnist_obj} for various AGRs.
For instance, Figures~\ref{fig:femnist_obj} (a) and (b) show that increasing $|D_p|$ monotonically increases update's loss and norm, respectively, and hence, can effectively poison the Average AGR~\cite{blanchard2017machine,mhamdi2018the}.
% Hence, using very high $|D_p|$ causes poisoned updates' losses and norms to explode and suffices to compromise the Average AGR~\cite{blanchard2017machine,mhamdi2018the}.

In our work, we propose two label flipping (LF) strategies: \textbf{static LF (SLF)} and \textbf{dynamic LF (DLF)}.
In SLF, for a sample $(\mathbf{x},y)$, the adversary flips labels in a static fashion as in Section~\ref{existing_dp}.
% the true label in a static fashion from $y$ to $C-1-y$, where $C$ is the total number of classes.
On the other hand, in DLF, the adversary computes a surrogate model $\hat{\theta}$, an estimate of $\theta^g$, e.g., using the available benign data, and flips $y$ to the least probable label with respect to $\hat{\theta}$, i.e., to ${argmin}\ \hat{\theta}(\mathbf{x})$.
We observe that the impacts of the two LF strategies are dataset dependent. Therefore, for each dataset, we experiment with both of the strategies and, when appropriate, present the best results.
We now specify our DPA for the AGRs in Section~\ref{background:agr}.

\paragraphb{\em Average:}
% Average~\cite{mcmahan2017communication} does not impose any robustness constraints on its input updates. 
To satisfy the attack objective in~\eqref{eq:gen_opt_dp} for Average AGR, we produce updates with large loss and norm~\cite{blanchard2017machine,mhamdi2018the} using very large amounts of label flipped data (Figures~\ref{fig:femnist_obj}-(a,b)).
% To achieve this, we leverage our observations from Figures~\ref{fig:femnist_obj}-(a,b) and use large $|D_p|$ for each compromised client. 

To obtain large $|D_p|$, we combine the benign data of all compromised clients and flip their labels using either SLF or DLF strategy (simply SLF/DLF).
To increase $|D_p|$ further, we add Gaussian noise to existing feature vectors of $|D_p|$ to obtain new feature vectors and flip their labels using SLF or DLF.

\paragraphb{\em Norm-bounding:}
% Norm-bounding AGR scales an input update if its norm is higher than a threshold~\cite{sun2019can}.
To attack Norm-bounding AGR (Section~\ref{agr:normb}), we use large $|D_p|$ to generate poisoned updates that incur high losses on benign data (as we show in Figure~\ref{fig:femnist_obj}-(b)).
As our evaluations will show, even if their norms are bounded, such poisoned updates remain far from benign updates and have high poisoning impacts. This leads to effective attacks, but only at high percentages of compromised clients (e.g., $M$=10\%).
Due to space restrictions, we provide the details of our DPAs on Mkrum and Trmean in Appendix~\ref{appdx:missing_attacks:dp}.

\begin{figure}
\centering
\includegraphics[scale=.65]{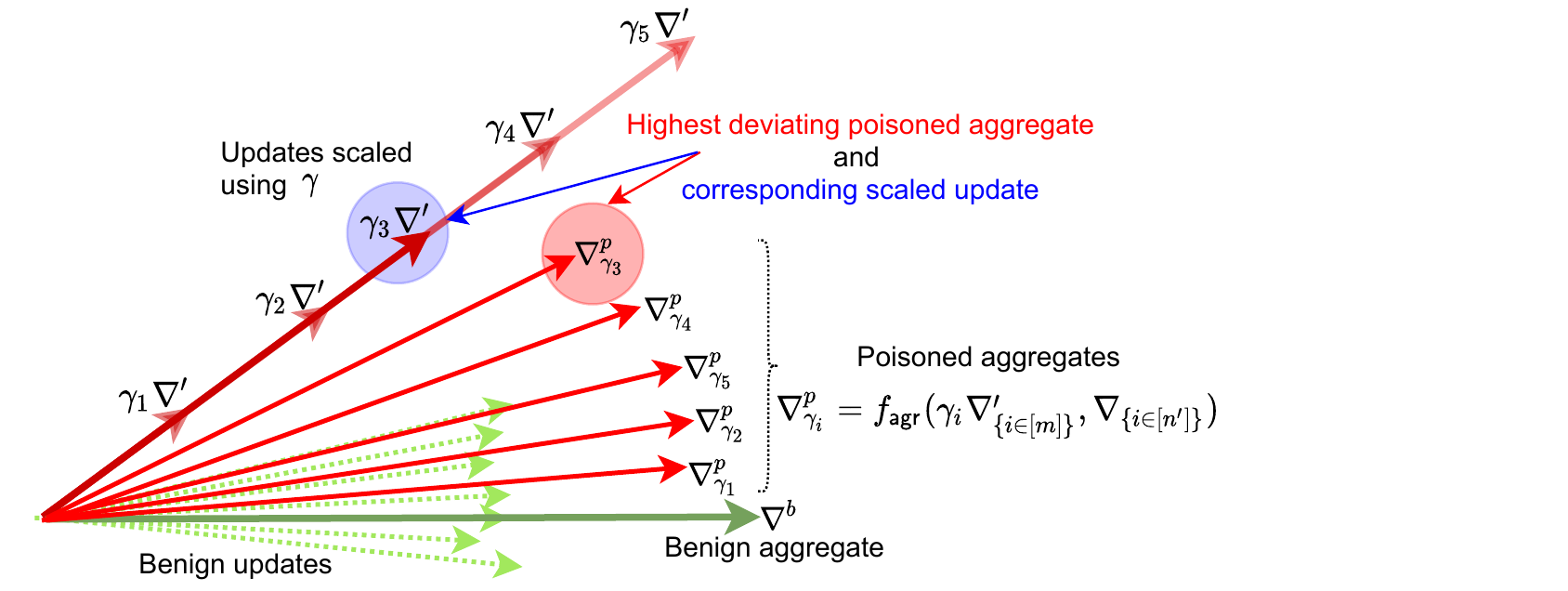}
\caption{Schematic of our PGA attack: PGA first computes a \emph{poisoned update} $\nabla'$ using stochastic gradient ascent (SGA). Then, $f_\mathsf{project}$ finds the scaling factor  $\gamma$ that maximizes the deviation between benign aggregate $\nabla^b$ and poisoned aggregate $\nabla^p_{\gamma}$. Robust aggregations easily discard the scaled poisoned updates, $\gamma\nabla'$, with very high $\gamma$ (e.g., $\gamma_{\{4,5\}}$), while those with very small $\gamma$ (e.g., $\gamma_{\{1,2\}}$) have no impact.}
\label{fig:fproject}
\vspace*{-1.75em}
\end{figure}

\subsubsection{Our Model Poisoning Attacks (MPAs)}\label{improved:mp}

We use~\eqref{eq:gen_opt} as the general optimization problem for our MPAs.
To solve this optimization, we craft a poisoned model $\theta'$ with high $L(D_b;\theta')$ while ensuring that the corresponding poisoned update, $\nabla'$, circumvents the target AGR.

% Recall that, the model poisoning adversary (Section~\ref{threat:practice_mp}) can access the global model parameters and directly manipulate the updates of compromised clients.
Model poisoning adversary can directly manipulate the compromised clients' updates (Section~\ref{threat:practice_mp}).
Hence, first, our attack uses the stochastic gradient ascent (SGA) algorithm (instead of SGD) and fine-tunes $\theta^g$ to increase (instead of decreasing) the loss on some benign data, $D_b$,  to obtain a malicious $\theta'$.
But, in order to ensure that the corresponding poisoned update, i.e., $\nabla'=\theta'-\theta^g$, circumvents the target AGR, we \emph{project} the update on a ball of radius $\tau$ around origin, i.e., scale the update to have a norm $\Vert\nabla'\Vert\leq \tau$, where $\tau$ is the average of norms of the available benign updates.
Hence, we call our attack \textbf{projected gradient ascent (PGA)}.
To perform stochastic gradient ascent, we increase the loss on batch $b$ of data by \emph{using the opposite of a benign gradient direction}, i.e., $-\nabla_{\theta}\mathcal{L}(\theta;b)$.

% We propose two ways to perform stochastic gradient ascent: We increase the loss on batch $b$ of data by either (1) \emph{using the opposite of a benign gradient direction}, i.e., $-\nabla_{\theta}\mathcal{L}(\theta;b)$, or  (2) \emph{using a malicious gradient direction}, e.g., $\nabla_{\theta}\mathcal{L}(\theta;b')$, similar to the label flipping attacks~\cite{fang2020local},  where $b'$ is the data in $b$, but with incorrect labels. 
% Training using $\nabla_{\theta}\mathcal{L}(\theta;b')$ overfits $\theta$ to the incorrectly labeled data in $b'$ and increases the loss of $\theta$ on benign data. The impacts of these two SGA strategies are dataset dependent, hence we experiment with the both and report the best of results.

% \red{why dont you call condition function as projection function? overall, you need to better explain your algorithm. things are scrambled. also maybe a figure can help}

Algorithm~\ref{alg:mp} (Appendix~\ref{appdx:missing_attacks}) gives the overview of our MPA. The adversary first computes $\tau$ (line 2), an average of the norms of some benign updates available to her  ($\nabla_{\{i\in[n']\}}$).
Then, the adversary fine-tunes $\theta^g$ using $D_p$ and SGA to compute a poisoned update $\nabla'$; our attack computes $\nabla'$ for any AGR in the same manner.
Finally, the adversary uses $f_\mathsf{project}$ function to appropriately project $\nabla'$ in order to circumvent the robustness criteria of the target AGR, $f_\mathsf{agr}$.

% \begin{figure}
% \centering
% % \hspace*{-1em}
% \includegraphics[scale=.75]{figures/pga3.pdf}
% % \vspace*{-2em}
% \caption{Schematic of our PGA attack: PGA first computes a \emph{poisoned update} $\nabla'$ using stochastic gradient ascent (SGA). Then, $f_\mathsf{project}$ finds the scaling factor  $\gamma$ that maximizes the deviation between benign aggregate $\nabla^b$ and poisoned aggregate $\nabla^p_{\gamma}$. Robust aggregations easily discard the scaled poisoned updates, $\gamma\nabla'$, with very high $\gamma$ (e.g., $\gamma_{\{4,5\}}$), while those with very small $\gamma$ (e.g., $\gamma_{\{1,2\}}$) have no impact.}
% \label{fig:fproject}
% % \vspace*{-1em}
% \end{figure}

Algorithm~\ref{alg:scale} (Appendix~\ref{appdx:missing_attacks}) describes $f_\mathsf{project}$: It computes $\nabla^b=f_\mathsf{avg}(\nabla_{\{i\in[n']\}})$. Then, it finds a scaling factor $\gamma$ for $\nabla'$ that maximizes the distance between  the benign aggregate $\nabla^b$ and the poisoned aggregate $\nabla^p=f_\mathsf{agr}(\gamma\nabla'_{\{i\in[m]\}},\nabla_{\{i\in[n']\}})$. 
Note that, there can be many ways to optimize $\gamma$~\cite{shejwalkar2021manipulating}, but we empirically observe that simply searching for $\gamma$ in a pre-specified range (e.g., $[1,\Gamma]$ with $\Gamma\in\mathbb{R}^+$) yields strong attacks (line 6).
Figure~\ref{fig:fproject} depicts the idea of $f_\mathsf{project}$ algorithm.

Due to the modular nature of our attacks, one can attack any given AGR by plugging its algorithm in Algorithm~\ref{alg:scale}.
This is unlike Sun et al.~\cite{sun2019can}, who propose a similar \emph{targeted} attack which only works against norm-bounding AGR.

Furthermore, to reduce computation, below we tailor $f_\mathsf{project}$ to the state-of-the-art AGRs from Section~\ref{background:agr}; note that, the adversary obtains a poisoned update, $\nabla'$, using Algorithm~\ref{alg:mp} before tailoring $f_\mathsf{project}$ to the target AGR.

\paragraphb{\em Average:}
Average does not impose any robustness constraints, therefore, we simplify $f_\mathsf{project}$ by scaling $\nabla'$ by an arbitrarily large constant, e.g., $10^{20}$.
If the server selects a compromised client, such poisoned update suffices to completely poison $\theta^g$.

\paragraphb{\em Norm-bounding:}
Following the Kirchoff's law, we assume that the attacker knows the norm-bounding threshold, $\tau$, and therefore, $f_\mathsf{project}$ scales $\nabla'$ by $\frac{\tau}{\Vert\nabla'\Vert}$, so that the norm of the final $\nabla'$ will be $\tau$.
We provide the details of our MPAs on Mkrum and Trmean in Appendix~\ref{appdx:missing_attacks:mp}.

% \input{experimental_setup}
% !TEX root = main.tex

% \begin{figure*}
% \centering
% \includegraphics[scale=.45]{figures/vggbn_compare_attacks_average.pdf}
% % \vspace*{-1em}
% \caption{{Attack impacts ($I_\theta$) of state-of-the-art data (DPA-DLF/SLF) and model (MPA) poisoning attacks on cross-device FL with average AGR. \emph{$I_\theta$'s are significantly lower for practical percentages of compromised clients ($\leq0.1\%$)}.} }
% \label{fig:nonrobust_fl}
% % \vspace*{-1em}
% \end{figure*}

% \vspace{-.5em}
\section{Analysis of FL Robustness in Practice} \label{results}
\vspace{-.25em}

In this section, we evaluate state-of-the-art data (DPAs) and model poisoning attacks (MPAs) against  non-robust and robust FL algorithms (Section~\ref{background:agr}), under practical threat models from Section~\ref{threat:practice}.
We start by  analyzing cross-device FL (Sections~\ref{exp:nonrobust_fl} to \ref{exp:params}), as it is barely studied in previous works and is more susceptible to poisoning.
Then,   we will analyze  cross-silo FL in Section~\ref{exp:cross_silo}.

\paragraphb{Experimental setup:} Please refer to Appendix~\ref{exp_setup}.

\paragraphb{Attack impact metric:}
$A_\theta$ denotes the maximum accuracy that the global model achieves over all  FL training rounds, without any attack. $A^*_\theta$ for an attack denotes the maximum accuracy of the model under the given attack. We define \emph{attack impact}, $I_\theta$, as \emph{the reduction in the accuracy of the global model due to the attack}, hence for a given attack,  $I_\theta = A_\theta - A^*_\theta$.

% \vspace*{-.75em}
\subsection{Evaluating Non-robust FL (Cross-device)}\label{exp:nonrobust_fl}
% As discussed in Section~\ref{agr:average}, 
We study Average AGR due to its practical significance and widespread use.
Previous works~\cite{blanchard2017machine,fang2020local,shejwalkar2021manipulating,baruch2019a,mhamdi2018the,yin2018byzantine} have argued that even a single compromised client can prevent the convergence of FL with Average AGR.
However, our results contradict those of previous works: we show that this established belief about Average AGR is \emph{incorrect} for production cross-device FL.
% \red{i dont understand this sentence. you mean " we contradict the restuls shown in prior work?}\virat{yes}.

Figure~\ref{fig:nonrobust_fl} shows the attack impacts ($I_\theta$) of various DPAs and MPAs.
% $I_\theta$ of an attack is the reduction in the accuracy of the global model, $\theta^g$, due to the attack.
Note that, for the Average AGR, all MPAs~\cite{shejwalkar2021manipulating,fang2020local,baruch2019a}, including ours, are the same and craft arbitrarily large updates in a malicious direction. Hence, we show a single line for MPAs in Figure~\ref{fig:nonrobust_fl}.

We see that \emph{for cross-device FL, when percentages of compromised clients ($M$) are in practical ranges (Table~\ref{tab:practical_ranges}), $I_\theta$'s of all the attacks are very low, i.e., the final $\theta^g$ converges with high accuracy}.
For FEMNIST, $I_\theta$ of MPAs at $M$=0.01\% is $\sim$2\% and $I_\theta$ of DPAs at 0.1\% is $\sim$5\%. In other words, compared to the no attack accuracy (82.3\%), the attacks reduce the accuracy by just 2\% and 5\%. Similarly, we observe very low $I_\theta$'s for the Purchase and CIFAR10 datasets. 
% Our findings \emph{directly contradict the  previous works~\cite{blanchard2017machine,yin2018byzantine,mhamdi2018the} which claim that with even a single compromised client, FL with Average AGR cannot converge}.
% \red{well, we dont need to remove this if we contradic prev work}\virat{actually, it repeats so i removed; we already say this in the beginning of this section}

Note that, here we use very large local poisoned data ($D_p$) for our DPAs, as DPAs on Average AGR become stronger with higher $|D_p|$ (Section~\ref{improved:dp}); $|D_p|$'s are 20,000, 50,000, and 20,000 for FEMNIST, CIFAR10, and Purchase, respectively. 
% But, the devices in cross-device FL have low processing powers (e.g., smart phones) and cannot process such large $D_p$ in short duration of FL rounds. 
However, as we will show in Section~\ref{exp:local_poison_size}, \emph{under practical $|D_p|$, $I_\theta$'s of DPAs are negligible even with $M$=10\%}.

{\em The inherent robustness of cross-device FL is due to its client sampling procedure}. In an FL round, the server selects a very small fraction of all FL clients. Hence, in many FL rounds no  compromised clients are chosen when $M$ ($<1\%$) is in  practical ranges.
\vspace*{-.5em}
\begin{mybox}
\paragraphb{(Takeaway~\ref{exp:nonrobust_fl})} Contrary to the common belief, production cross-device FL 
% ($M<0.1\%$ for DPAs, $M<0.01\%$ for MPAs, and $n\ll N$) 
with (the naive) Average AGR converges with high accuracy
% , i.e., it is highly robust, 
even in the presence of untargeted poisoning attacks.
\end{mybox}

\begin{figure*}
    \begin{subfigure}[b]{0.2\textwidth}
      \includegraphics[scale=.48]{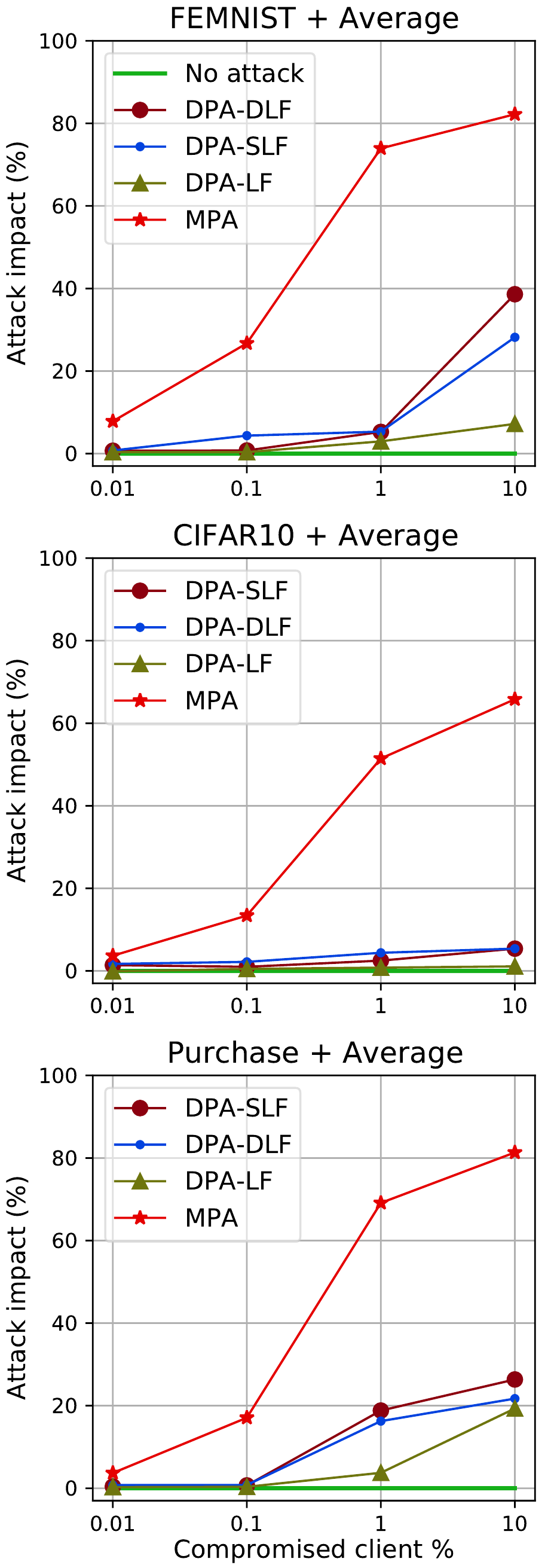}
      \caption{\small Non-robust FL}
      \label{fig:nonrobust_fl}
    \end{subfigure}
    {\hspace*{2.3em}\hfill\color{black}\vrule\hfill}%
    \begin{subfigure}[b]{0.725\textwidth}
      \includegraphics[scale=.48]{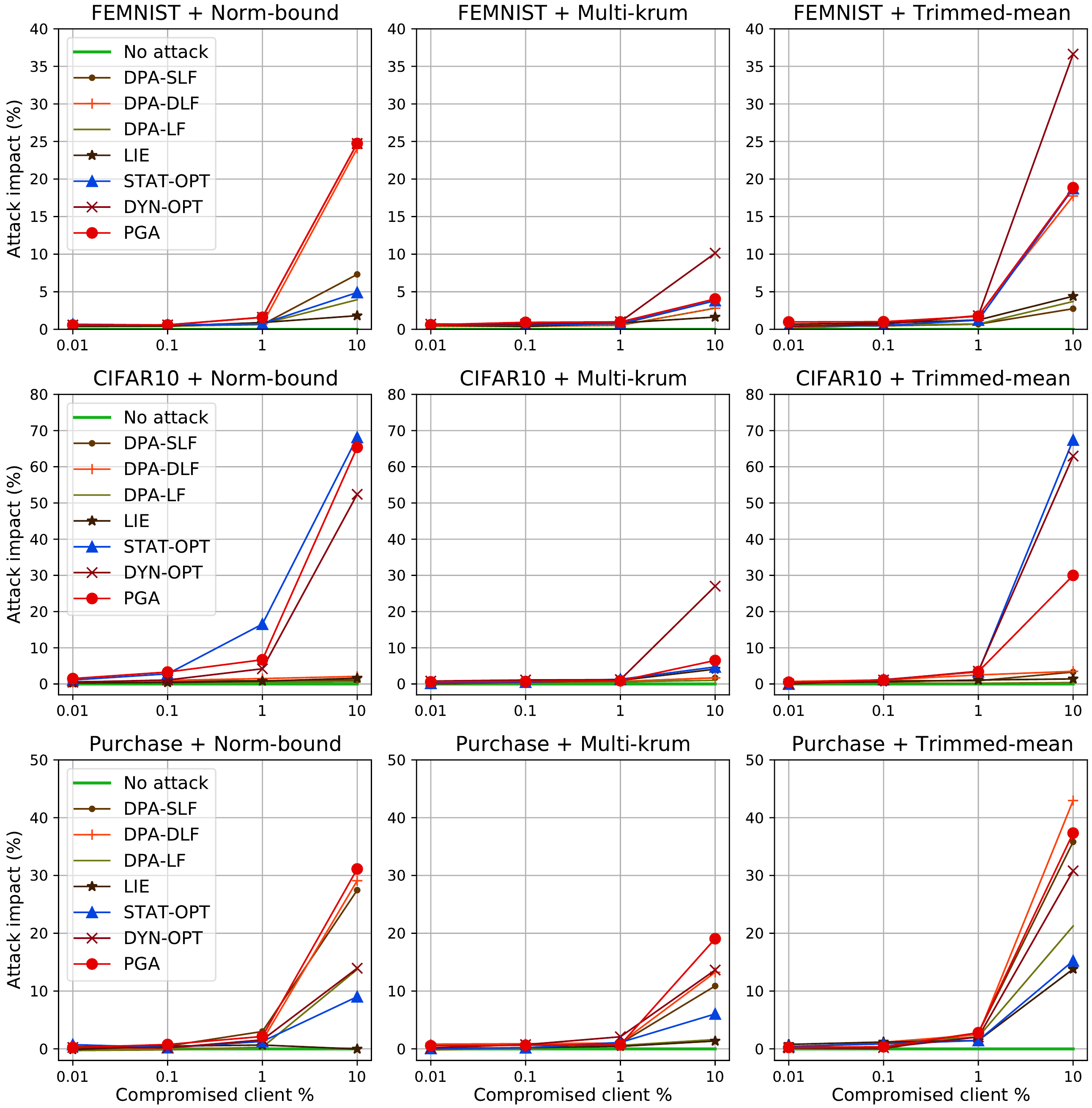}
      \caption{\small Robust FL}
      \label{fig:robust_fl}
    \end{subfigure}
    \vspace*{-.25em}
    \caption{(\ref{fig:nonrobust_fl}) Attack impacts ($I_\theta$) of state-of-the-art data (DPA-DLF/SLF) and model (MPA) poisoning attacks on cross-device FL with average AGR. \emph{$I_\theta$'s are significantly low for practical percentages of compromised clients ($M$$\leq$0.1\%)}. (\ref{fig:robust_fl}) $I_\theta$ of various poisoning attacks (Section~\ref{method}) on robust AGRs (Section~\ref{background:agr}). \emph{These AGRs are highly robust for practical $M$ values}.}
    \label{fig:all_fl}
    \vspace*{-1.7em}
  \end{figure*}

% \vspace*{-.5em}
\subsection{Evaluating Robust FL (Cross-device)}\label{exp:robust_fl}
% \vspace*{-.5em}
% As discussed before, previous works  make misleading conclusions about the robustness of robust AGRs to poisoning attacks, due to their impractical assumptions.
In this section, contrary to previous works, we study the robustness of robust AGRs for cross-device FL when percentages of compromised clients ($M$) are in practical ranges.
Figure~\ref{fig:robust_fl} shows the poisoning impact ($I_\theta$) of DPAs and MPAs for Norm-bounding (Normb), Multi-krum (Mkrum), and Trimmed-mean (Trmean) AGRs. Below, we discuss three \textbf{key takeaways}: 

\subsubsection{\bf Cross-device FL with robust AGRs is highly robust in practice}\label{exp:robust_fl_takeaway1}
% to state-of-the-art poisoning attacks in practice
$I_\theta$ of attacks on robust AGRs are negligible in practice, i.e., when $M\leq0.1\%$ for DPAs and $M\leq0.01\%$ for MPAs.
For instance, $I_\theta\leq 1\%$ for all of state-of-the-art attacks on all the three datasets, i.e., the attacks reduce the accuracy of $\theta^g$ by less that 1 percent.
% , while $I_\theta<5\%$ for CIFAR10 and $I_\theta<1\%$ for Purchase.

We also run FL with a robust AGR for a very large number (5,000) of rounds to investigate if the strongest of MPAs against the AGR with $M=0.1\%$ can break the AGR after long rounds of continuous and slow poisoning. Figure~\ref{fig:5k_rounds} shows the results: Mkrum and Trmean remain completely unaffected (in fact accuracy of the global model increases), while accuracy due to Normb reduces by $<$5\%.

% Our findings directly contradict some of the conclusions of state-of-the-art poisoning attacks works that study the empirical robustness of robust AGRs~\cite{shejwalkar2021manipulating,baruch2019a,fang2020local}. 
In summary, state-of-the-art poisoning attacks~\cite{shejwalkar2021manipulating,baruch2019a,fang2020local} demonstrate that the robust AGRs are significantly less robust than their theoretical guarantees. On the other hand, our findings show that these AGRs are more than sufficient to protect, more practical, production cross-device FL against untargeted poisoning.
This is due to the peculiar client sampling of cross-device FL, as discussed in Section~\ref{exp:nonrobust_fl}.
\vspace*{-.5em}
\begin{mybox}
\paragraphb{(Takeaway~\ref{exp:robust_fl_takeaway1})} 
% Contrary to the conclusions of prior works~\cite{shejwalkar2021manipulating,fang2020local,baruch2019a}, 
Cross-device FL with robust AGRs is highly robust to state-of-the-art poisoning attacks under production FL environments ($M<$0.1\%, $n\ll N$).
% (but  significantly less robust than their theoretical guarantees). 
\end{mybox}

% \amir{the takeaways and headings are repetitive }\virat{do u mean different takeaways are the same or takeaway and heading are the same?}

% \subsubsection{\bf Simple AGRs are sufficient to protect cross-device FL in practice}\label{exp:robust_fl_takeaway2}
\subsubsection{\bf Investigating simple and efficient robustness checks is necessary}\label{exp:robust_fl_takeaway2}

Most of the state-of-the-art robust AGRs with strong theoretical guarantees~\cite{blanchard2017machine,mhamdi2018the,yin2018byzantine,xie2018generalized} have complex robustness checks on their inputs, which incur high  computation and storage overheads.  
For instance, to process $n$ updates of length $d$, the computational complexity of Mkrum is $\mathcal{O}(dn^2)$ and that of Trmean is $\mathcal{O}(dn\text{log}n)$. Therefore, in production FL systems where $n$ can be up to $5,000$~\cite{kairouz2019advances,bonawitz2019towards}, the computation cost prohibits the use of such robust AGRs.

On the other hand, Norm-bounding only checks for the norm of its inputs and has computation complexity of $\mathcal{O}(d)$, same as Average.
% Here, we study the robustness of Normb, which is studied only in the context of targeted attacks~\cite{sun2019can,wang2020attack}.
% 
Figure~\ref{fig:robust_fl} shows that a simple and efficient AGR, Norm-bounding, protects cross-device FL against state-of-the-art poisoning attacks similarly to the theoretically robust  (and expensive) AGRs, under practical $M$.
For instance, for all the datasets with $M\leq1\%$, $I_\theta<1\%$ for all of the AGRs (Figure~\ref{fig:robust_fl}).
% (Figures~\ref{fig:robust_fl_femnist},~\ref{fig:robust_fl_cifar10},~\ref{fig:robust_fl_purchase}). 
% While for CIFAR10 with $M\leq0.1\%$, $I_\theta<5\%$ in Figure~\ref{fig:robust_fl_cifar10}.
Our evaluation highlights that  simple robust AGRs, e.g., Norm-bounding, can effectively  protect cross-device FL in practice, and calls for further investigation and invention of such low-cost robust AGRs.
\vspace*{-.25em}
\begin{mybox}
\paragraphb{(Takeaway~\ref{exp:robust_fl_takeaway2})} Even the simple, low-cost Norm-bounding AGR is enough to protect production FL 
%  ($M$~$<$0.1\%, $n\ll N$) 
against untargeted poisoning, questioning the need for the more sophisticated (and costlier) AGRs.  
\end{mybox}
\vspace*{-.5em}

% \subsubsection{\bf Empirical assessment is a must to understand robustness of AGRs}\label{exp:robust_fl_takeaway3}

\subsubsection{\bf Thorough empirical assessment of robustness is inevitable}\label{exp:robust_fl_takeaway3}
% \amir{i dont understand this takeaway. u mean compared o simple agrs? or are you contradicting their theoretical claims?}\virat{Both: simple agrs without guarantees perform as well as the fancy agrs even at high $m$'s and at high $m$'s fancy agrs do not provide the guarantees they claim.}
Theoretically robust AGRs claim robustness to poisoning attacks at high $M$'s, e.g., in theory, Mkrum~\cite{blanchard2017machine} and Trmean~\cite{yin2018byzantine} are robust for $M\leq25\%$. But, we observe that, even at the theoretically claimed values of $M$, these robust AGRs do not exhibit high robustness; in fact, simple AGRs, e.g., Norm-bounding, are equally robust.
Note in Figure~\ref{fig:robust_fl} that, for FEMNIST at $M$=10\%, $I_\theta$'s on Trmean are \emph{higher} than on Norm-bounding. For CIFAR10 at $M$=10\%, $I_\theta$'s for Norm-bounding and Trmean are almost similar. 
% Note that prior works~\cite{baruch2019a,fang2020local,shejwalkar2021manipulating} make similar empirical observations for cross-silo FL.

Sections~\ref{exp:robust_fl_takeaway2} and~\ref{exp:robust_fl_takeaway3} show that, some of the sophisticated, theoretically robust AGRs do not outperform simpler robust AGRs at \emph{any} ranges of $M$.
More importantly they demonstrate the shortcomings of the methodology used to assess the robustness of AGRs in previous works~\cite{blanchard2017machine,mhamdi2018the,yin2018byzantine,xie2018generalized} (because these works use very preliminary attacks) and highlight that a thorough empirical assessment is necessary to understand the robustness of AGRs in production FL systems.
% \vspace*{-.5em}
\begin{mybox}
\paragraphb{(Takeaway~\ref{exp:robust_fl_takeaway3})} 
Understanding the robustness of AGRs in production FL requires a thorough empirical assessment of AGRs, on top of their theoretical robustness analysis.
\end{mybox}
\vspace*{-.5em}

\begin{figure}
\centering
\hspace*{-1em}
\includegraphics[scale=.685]{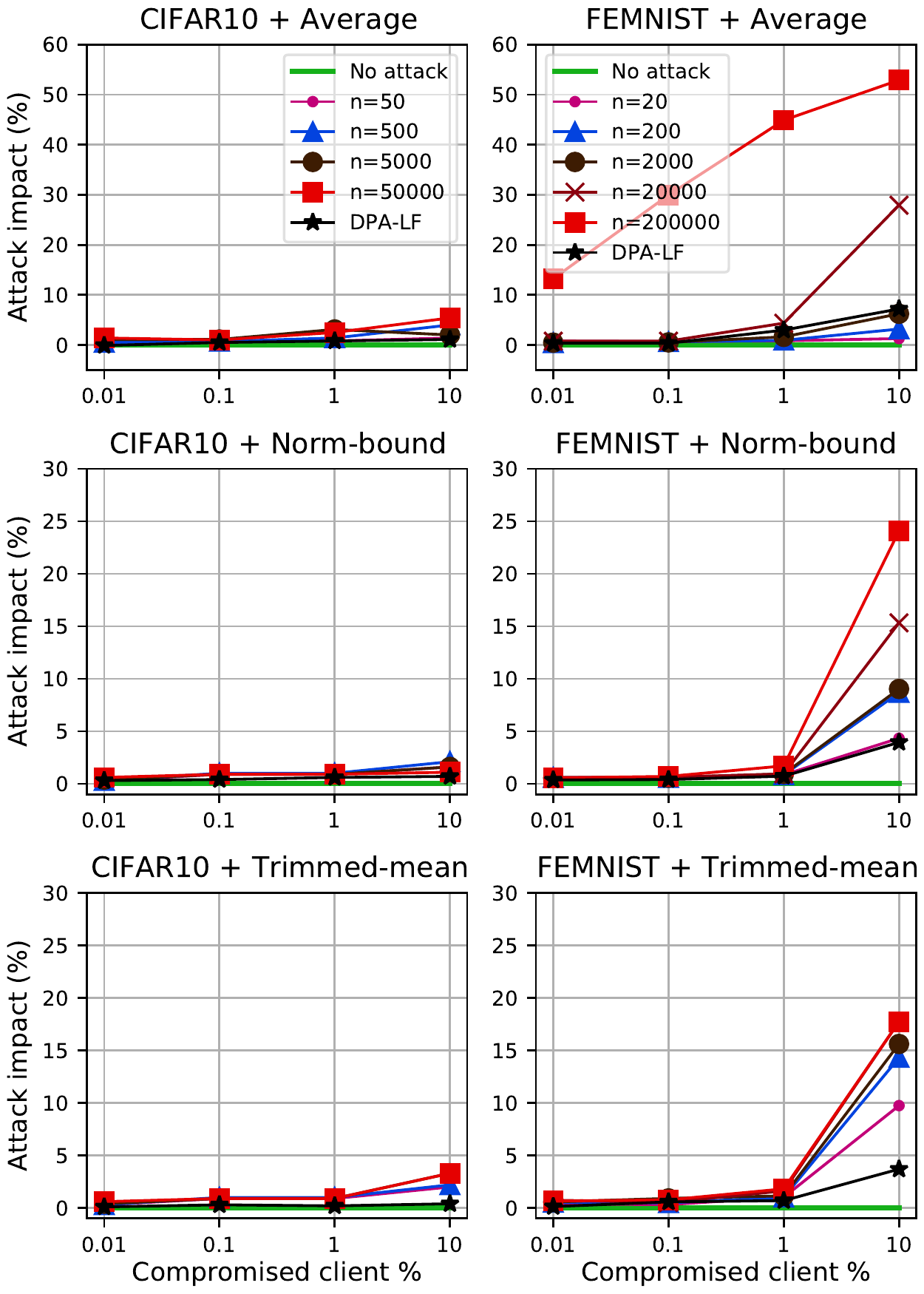}
\vspace*{-.25em}
\caption{Effect of varying sizes of local poisoned dataset $D_p$ on impacts $I_\theta$ of the best of DPAs. When $|D_p|$ and $M$ are in practical ranges, $I_\theta$'s are negligible for robust AGRs and are dataset dependent for non-robust Average AGR.}
\label{fig:nsamples_impact}
\vspace*{-1.75em}
\end{figure}

\vspace*{-.5em}
\subsection{Effect of FL Parameters on Poisoning (Cross-device)}\label{exp:params}
\vspace*{-.25em}

\subsubsection{Effect of the Size of Local Poisoning Datasets ($|D_p|$) on DPAs.}\label{exp:local_poison_size}
The success of our state-of-the-art data poisoning attacks depends on $|D_p|$ of compromised clients (Section~\ref{improved:dp}). In Sections~\ref{exp:nonrobust_fl} and~\ref{exp:robust_fl}, we use large $|D_p|$ (e.g., 50,000 for CIFAR10) to find the highest impacts of DPAs. {But, as argued in Section~\ref{threat:practical_ranges},  in practice $|D_p|\leq 100\times|D|_\mathsf{avg}$; $|D|_\mathsf{avg}$ is the average size of local datasets of benign clients and it is around 20 (50) for FEMNIST (CIFAR10).
In Figure~\ref{fig:nsamples_impact}, we report $I_\theta$ of the best of DPA-SLF or DPA-DLF for $|D_p|\in\{1, 10, 10^2, 10^3, 10^4\}\cdot|D|_\mathsf{avg}$; we use impractically high $|D_p|$'s of up to $10^4\cdot|D|_\mathsf{avg}$ only for experimental analyses.}

Figure~\ref{fig:nsamples_impact} shows that $I_\theta$'s of DPAs slightly increase with $|D_p|$.
For FEMNIST and CIFAR10 with any AGR, including Average, $I_\theta$'s are negligible even for unrealistically high $|D_p|$ of $1000\times|D|_\mathsf{avg}$ for $M\leq 1\%$.
% But, for CIFAR10 and Average, $I_\theta$ is negligible only when $D_p\sim D_\mathsf{avg}$, otherwise $I_\theta$ remains high.
% For CIFAR10 and Normb, $I_\theta$ is negligible for $|D_p|$=$100\times|D|_\mathsf{avg}$ even for high $M$=1\%.
We omit Mkrum here, as  $|D_p|$ of the effective DPAs on Mkrum is always in practical ranges and close to $|D|_\mathsf{avg}$ (Section~\ref{improved:dp}).

To summarize, \emph{for all robust AGRs, DPAs have negligible impacts on FL when $|D_p|$ and $M$ are in practical ranges, while for non-robust AGRs, the reductions in $I_\theta$ are non-trivial and dataset dependent}. 
This also means that using a reasonable upper bound on the dataset sizes of FL clients can make FL highly robust to DPAs.

\begin{mybox}
\paragraphb{(Takeaway~\ref{exp:local_poison_size})} 
Enforcing a limit on the size of the local dataset of each client can act as a highly effective (yet simple) defense against untargeted DPAs in production FL.  
\end{mybox}

% \begin{figure}
% \centering
% \hspace*{-1em}
% \includegraphics[scale=.655]{figures/new_impact_total_nclients_cifar10_vggbn_femnist_ai_partial.pdf}
% % \vspace*{-.5em}
% \caption{With 1\% compromised clients, \emph{increasing $|D|_\mathsf{avg}$ has no clear pattern of effects of  on attack impacts, but it increases the global model accuracy} (Figure~\ref{fig:total_nclients_impact_acc} in Appendix~\ref{appdx:missing_figures}). Due to space restrictions, we defer the plots of attack impacts and the global model accuracy for Multi-krum and Trimmed-mean AGRs to Figure~\ref{fig:total_nclients_impact_acc_mkrum_trmean} in Appendix~\ref{appdx:missing_figures}. 
% % For Mkrum and Trmean, for both CIFAR10 and FEMNIST, attack impacts for all $|D|_\mathsf{avg}$ are negligible. We omit them here due to space restrictions.
% }
% \label{fig:total_nclients_impact_ai}
% \vspace*{-1.25em}
% \end{figure}

\subsubsection{Effect of the Average Dataset Size of Benign FL Clients ($|D|_\mathsf{avg}$)}\label{exp:fl_data_sizes}

% Now, we study the effect of the average of local dataset sizes of benign clients ($|D|_\mathsf{avg}$).
Figure~\ref{fig:total_nclients_impact_ai} in Appendix~\ref{appdx:missing_figures} shows $I_\mathsf{\theta}$ when we vary $|D|_\mathsf{avg}$.
% For presentation clarity, we present the error rates of the global model, i.e., (100 - model accuracy).
To emulate varying $|D|_\mathsf{avg}$, we vary the total number of FL clients, $N$, for given dataset, e.g., for CIFAR10, $|D|_\mathsf{avg}$ is 50 (10) for  $N$=1,000 ($N$=5,000).
As discussed in Section~\ref{threat:practical_ranges}, we use $|D_p|$=$100\times|D|_\mathsf{avg}$ for DPAs.

We observe \emph{no clear effect of varying $|D|_\mathsf{avg}$ on $I_\mathsf{\theta}$'s}. For instance, at $M$=1\%, $I_\mathsf{\theta}$'s of our PGA and DPA-SLF on CIFAR10 + Normb reduce with increase in $|D|_\mathsf{avg}$, while $I_\mathsf{\theta}$ of any attacks on FEMNIST with robust AGRs do not change with varying $|D|_\mathsf{avg}$. Due to space restrictions, we defer the explanations of each of these observations to Appendix~\ref{appdx:fl_data_sizes}.

More importantly, we observe that \emph{even with moderately high $|D|_\mathsf{avg}$, cross-device FL completely mitigates state-of-the-art DPAs and MPAs despite $M$ being impractically high}, with an exception of MPAs on Average AGR. For instance, for CIFAR10 with $|D|_\mathsf{avg}$=50 and FEMNIST with $|D|_\mathsf{avg}$=200, all robust AGRs almost completely mitigate all of DPAs and MPAs, while Average AGR mitigates all DPAs. However, as MPAs are very effective against Average, their $I_\mathsf{\theta}$ remains high.
As clients in FL continuously generate data locally~\cite{mcmahan2018learning,mcmahan2017communication}, it is common to have large $|D|_\mathsf{avg}$ in practice. Interestingly, our evaluation also implies that simply lower bounding the dataset sizes of FL clients improves FL robustness.
% \vspace*{-.5em}
\begin{mybox}
\paragraphb{(Takeaway~\ref{exp:fl_data_sizes})} When local dataset sizes of benign clients are in practical regimes (Table~\ref{tab:practical_ranges}), cross-device FL with robust AGRs is highly robust to untargeted poisoning.
\end{mybox}

\subsubsection{Number of Clients Selected Per Round.}\label{exp:round_nclients}

{Figure~\ref{fig:round_nclients_impact}  (Appendix~\ref{additional_exp_details}) shows the effect of varying the number of clients ($n$) selected by the server in each round (for $M$=1\%). Similar to~\cite{fang2020local}, we do not observe any noticeable effect of $n$ on the impact of attacks, since the expected percentage of compromised clients ($M$) does not change with $n$.
But, we observe the opposite behavior for MPAs on Average AGR. This is because, as soon as the server selects even a single compromised client, MPA prevents any further learning of the global model.
An increase in $n$ increases the chances of selecting compromised clients, hence amplifying the attack.}
\vspace*{-.5em}
\begin{mybox}
\paragraphb{(Takeaway~\ref{exp:round_nclients})} {The number of clients selected in each round of production cross-device FL  has no noticeable effect on the impacts of untargeted poisoning attacks, with the exception of MPAs on Average AGR.}
\end{mybox}

% \begin{figure}
% \centering
% % \hspace*{-1em}
% \includegraphics[scale=.5]{figures/new_diff_arch_femnist.pdf}
% % \vspace*{-1em}
% \caption{Impacts of the DPA-DLF (i.e., dynamic label flipping based data poisoning attack from Section~\ref{improved:dp}) attack, which uses the knowledge of model architecture, reduce if the architecture is unknown.}
% \label{fig:diff_arch}
% % \vspace*{-.75em}
% \end{figure}

\subsubsection{Effect of Unknown Global Model Architecture on DPAs}\label{exp:unknown_arch}

% \begin{figure}
% % \vspace*{-2em}
% \centering
% \includegraphics[scale=.5]{figures/new_diff_arch_femnist.pdf}
% % \vspace*{-.5em}
% \caption{Impacts of the DPA-DLF (Section~\ref{improved:dp}) attack, which uses the knowledge of model architecture, reduce if the architecture is unknown.}
% \label{fig:diff_arch}
% \end{figure}

DPA-DLF attack (Section~\ref{improved:dp}) uses the knowledge of global model's architecture to train a surrogate model. However, in practice, the nobox offline data poisoning adversary (Section~\ref{threat:practice_dp}) may not know the architecture. Hence, we evaluate impact of DPA-DLF under the unknown architecture setting. 

We emulate the unknown architecture setting for FEMNIST dataset. We assume that the adversary uses  a substitute convolutional neural network given in Table~\ref{tab:unknown_arch} (Appendix~\ref{appdx:missing_figures}) as they do not know the true architecture, which is LeNet in our experiments.
Figure~\ref{fig:diff_arch} (Appendix~\ref{appdx:missing_figures}) compares the impacts of DPA-DLF when the adversary uses the true and the substitute architectures.
Note that, \emph{impacts of DPA-DLF reduce  when the adversary uses the substitute architecture}.
% \vspace*{-.5em}

% 
\begin{mybox}
\paragraphb{(Takeaway~\ref{exp:unknown_arch})} 
The DPAs that rely on a surrogate model (e.g., our DLF) are less effective if the architectures of the surrogate and  global models do not match.
% Impacts of the DPAs, which use the knowledge of the global model architecture, reduce in the absence of such knowledge; MPAs always  know the model architecture.
\end{mybox}

% \vspace*{-.5em}
\subsection{Evaluating Robustness of Cross-silo FL}\label{exp:cross_silo}

In cross-silo FL, each of $N$ clients, i.e., silos (e.g., corporations like banks, hospitals, insurance providers, government organizations, etc.), collects data from many \emph{users} (e.g., bank customers or hospital patients) and collaboratively train the FL model; we denote the total number of users by $N'$.

Recall from Section~\ref{threat:practice_mp} that the model poisoning adversary completely breaks into the devices of compromised clients and, to be effective, persists in their systems for long duration because  model poisoning attacks are online attacks (Section~\ref{threat:practice_mp}). For cross-silo FL, this means that the  adversary should break into large corporations, e.g., a bank, who are bound by contract and have professionally maintained  software stacks.
Plausible cross-silo poisoning scenarios involve strong incentives (e.g., financial) and require multiple parties to be willing to risk the breach of contract by colluding or for one party to hack thereby risking criminal liability. 
This makes breaking into these silos practically unlikely, hence  we argue that \emph{model poisoning threats in cross-silo FL are impractical}. 
% We also argue that model poisoning attacks are unlikely to play a major risk when the clients involved are bound by contract and their software stacks professionally maintained (e.g., in banks, hospitals, etc.). 

Note that this is unlike the large scale data-breaches~\cite{passwords_windows_pc,passwords_1,passwords_2} which are short-lived  and are only capable of  stealing information, but not changing the infrastructure.

Hence, we only study the data poisoning threat for cross-silo FL. 
For worse-case analyses, we assume that the silos train their models on all the data contributed by their users. If the silos inspect the users' data and remove the mislabeled data, one should consider clean-label data poisoning attacks~\cite{shafahi2018poison,goldblum2020dataset}; we leave this study to future work. Note that, data inspection is not possible in cross-device FL as data of clients (who are also the users) is completely local, hence clean-label poisoning is not relevant in cross-device FL.

{We assume that each silo collects data from equal number (i.e., $N'/N$) of users.
For DPAs, we assume $M\%$ of the  $N'$ users are compromised and each of them shares poisoned data $D_p$ (computed as described in Section~\ref{improved:dp}) with their parent silo; as discussed in Section~\ref{threat:practical_ranges}, we assume $|D_p|=100\times|D|_\mathsf{avg}$ for \emph{each user}.
% \amir{why do you describe these here again? remove} 
We distribute the compromised users either \emph{uniformly} across the silos or \emph{concentrate} them in a few silos. For instance, consider 50 silos and 50 compromised users and that, each silo can have a maximum of 50 users. Then in the uniform case, a single compromised user shares her $D_p$ with each silo, while in the concentrated case, all the 50 compromised users share their $D_p$ with a single silo.}
% \amir{this i cant buy it. for a silo, is dp the poisoned part of that silo's dataset, or the whole size of its dataset? I assume for a silo only a fraction of local dataset is poisoned. make this clear earlier when you talk about dp} 
% for compromised clients. Because, unlike the cross-device FL, silos can inspect the compromised clients' data if it is suspiciously large.}

{Figure~\ref{fig:cross_silo} (Appendix~\ref{appdx:missing_figures}) shows the impacts of best of DPAs for the concentrated case. We see that \emph{cross-silo FL is highly robust to state-of-the-art DPAs}. Because, in the concentrated case, very large numbers of \emph{benign silos mitigate the poisoning impact of the very few ($M\%$) compromised silos}. 
We observe the same results for the uniform distribution case, because very large numbers of \emph{benign users in each silo mitigate the poisoning impacts of the very few ($M\%$) compromised users}}.  
% 
% \vspace*{-.5em}

\begin{mybox}
\textbf{(Takeaway~\ref{exp:cross_silo})} In production cross-silo FL, model poisoning attacks are not practical, and state-of-the-art data poisoning attacks have no impact even with  Average AGR.
% Cross-silo FL is highly robust to state-of-the-art DPAs and MPAs, even with non-robust Average AGR.\red{correct?---> Nope}\virat{we argue that MPAs are impractical for cross-silo}
\end{mybox}

% !TEX root = main.tex
% \vspace*{-.5em}
\section{Conclusions}
% \vspace*{-.5em}
In this work, we systematized the threat models of poisoning attacks on federated learning (FL), provided the practical ranges of various parameters relevant to  FL robustness, and designed a suite of untargeted model and data poisoning attacks on FL (including existing and our improved attacks). Using these attacks, we thoroughly evaluated the state-of-the-art defenses under production FL settings. 
% We showed that, due to their unrealistic assumptions, previous works on FL robustness make multiple misleading conclusions. 
We showed that the conclusions of previous FL robustness literature cannot be directly extended to production FL.
We presented concrete takeaways from our evaluations to correct some of the established beliefs and highlighted the need to consider production FL environments in research on FL robustness.

% \virat{removed the following as it is repeated and space issues}
% Supported by our thorough analysis and experimentation, we offered several key recommendations, including:
% \textbf{(1)} One should use practical ranges for  the parameters of threat models that directly influence the conclusions related to FL robustness. \textbf{(2)}
% Understanding the robustness of AGRs in production FL requires a thorough empirical assessment of AGRs (e.g., using our comprehensive suite of untargeted attacks), on the top of theoretical analysis. 

We hope that our systematization of practical poisoning threat models can steer the  community towards practically significant research problems in FL robustness. For instance, one such open problem is to obtain concrete theoretical robustness guarantees of existing defenses in production FL settings where only a very small fraction of all clients is randomly selected in each FL round.

\section*{Acknowledgements}
% \vspace*{-.5em}
% \paragraphb{Acknowledgements.}
The work was supported
by DARPA and NIWC under contract HR00112190125, and by the
%  1553301 and  
 NSF grants 1953786, 1739462, and 1553301.
The U.S.\
Government is authorized to reproduce and distribute reprints for Governmental purposes notwithstanding any copyright notation thereon. The views, opinions, and/or findings expressed are those
of the author(s) and should not be interpreted as representing the
official views or policies of the Department of Defense or the U.S.\ Government.

\bibliographystyle{IEEEtranS_7.bst}
\bibliography{privacy}

% Generated by IEEEtranS.bst, version: 1.12 (2007/01/11)
\begin{thebibliography}{10}
\providecommand{\url}[1]{#1}
\csname url@samestyle\endcsname
\providecommand{\newblock}{\relax}
\providecommand{\bibinfo}[2]{#2}
\providecommand{\BIBentrySTDinterwordspacing}{\spaceskip=0pt\relax}
\providecommand{\BIBentryALTinterwordstretchfactor}{4}
\providecommand{\BIBentryALTinterwordspacing}{\spaceskip=\fontdimen2\font plus
\BIBentryALTinterwordstretchfactor\fontdimen3\font minus
  \fontdimen4\font\relax}
\providecommand{\BIBforeignlanguage}[2]{{%
\expandafter\ifx\csname l@#1\endcsname\relax
\typeout{** WARNING: IEEEtranS.bst: No hyphenation pattern has been}%
\typeout{** loaded for the language `#1'. Using the pattern for}%
\typeout{** the default language instead.}%
\else
\language=\csname l@#1\endcsname
\fi
#2}}
\providecommand{\BIBdecl}{\relax}
\BIBdecl

\bibitem{gboard}
``Federated learning: Collaborative machine learning without centralized
  training data,''
  \url{https://ai.googleblog.com/2017/04/federated-learning-collaborative.html},
  2017.

\bibitem{alistarh2018byzantine}
D.~Alistarh, Z.~Allen-Zhu, and J.~Li, ``{Byzantine stochastic gradient
  descent},'' in \emph{NeurIPS}, 2018.

\bibitem{bagdasaryan2018how}
E.~Bagdasaryan, A.~Veit, Y.~Hua, D.~Estrin, and V.~Shmatikov, ``{How to
  backdoor federated learning},'' in \emph{AISTATS}, 2020.

\bibitem{barreno2010the}
M.~Barreno, B.~Nelson, and A.~D. Joseph, ``{The security of machine
  learning},'' \emph{Machine Learning}, 2010.

\bibitem{baruch2019a}
M.~Baruch, B.~Gilad, and Y.~Goldberg, ``{A Little Is Enough: Circumventing
  Defenses For Distributed Learning},'' in \emph{NeurIPS}, 2019.

\bibitem{bernstein2018signsgd}
J.~Bernstein, J.~Zhao, K.~Azizzadenesheli, and A.~Anandkumar, ``{signSGD with
  Majority Vote is Communication Efficient and Fault Tolerant},'' in
  \emph{ICLR}, 2018.

\bibitem{bhagoji2019analyzing}
A.~N. Bhagoji, S.~Chakraborty, P.~Mittal, and S.~Calo, ``{Analyzing federated
  learning through an adversarial lens},'' in \emph{ICML}, 2019.

\bibitem{biggio2012poisoning}
B.~Biggio, B.~Nelson, and P.~Laskov, ``{Poisoning attacks against support
  vector machines},'' in \emph{ICML}, 2012.

\bibitem{biggio2018wild}
B.~Biggio and F.~Roli, ``{Wild patterns: Ten years after the rise of
  adversarial machine learning},'' \emph{Pattern Recognition}, 2018.

\bibitem{blanchard2017machine}
P.~Blanchard, R.~Guerraoui, J.~Stainer \emph{et~al.}, ``{Machine learning with
  adversaries: Byzantine tolerant gradient descent},'' in \emph{NeurIPS}, 2017.

\bibitem{bonawitz2019towards}
K.~Bonawitz, H.~Eichner, W.~Grieskamp, D.~Huba, A.~Ingerman, V.~Ivanov,
  C.~Kiddon, J.~Konecn{\'{y}}, S.~Mazzocchi, B.~McMahan, T.~V. Overveldt,
  D.~Petrou, D.~Ramage, and J.~Roselander, ``Towards federated learning at
  scale: System design,'' in \emph{MLSys}, 2019.

\bibitem{mcmahan2018learning}
M.~H. Brendan, D.~Ramage, K.~Talwar, and L.~Zhang, ``{Learning differentially
  private recurrent language models},'' in \emph{ICLR}, 2018.

\bibitem{caldas2018leaf}
S.~Caldas, P.~Wu, T.~Li, J.~Kone{\v{c}}n{\`y}, H.~B. McMahan, V.~Smith, and
  A.~Talwalkar, ``{LEAF: A benchmark for federated settings},''
  \emph{arXiv:1812.01097}, 2018.

\bibitem{cao2021provably}
X.~Cao, J.~Jia, and N.~Z. Gong, ``{Provably Secure Federated Learning against
  Malicious Clients},'' in \emph{AAAI}, 2021.

\bibitem{chang2019cronus}
H.~Chang, V.~Shejwalkar, R.~Shokri, and A.~Houmansadr, ``{Cronus: Robust and
  Heterogeneous Collaborative Learning with Black-Box Knowledge Transfer},''
  \emph{arXiv:1912.11279}, 2019.

\bibitem{chen2018draco}
L.~Chen, H.~Wang, Z.~Charles, and D.~Papailiopoulos, ``{Draco:
  Byzantine-resilient distributed training via redundant gradients},'' in
  \emph{ICML}, 2018.

\bibitem{chen2017targeted}
X.~Chen, C.~Liu, B.~Li, K.~Lu, and D.~Song, ``{Targeted backdoor attacks on
  deep learning systems using data poisoning},'' \emph{arXiv:1712.05526}, 2017.

\bibitem{cohen2017emnist}
G.~Cohen, S.~Afshar, J.~Tapson, and A.~Van~Schaik, ``{EMNIST: Extending MNIST
  to handwritten letters},'' in \emph{IJCNN}, 2017.

\bibitem{cs231n_backprop}
``{CS231n: Convolutional Neural Networks for Visual Recognition},''
  \url{https://cs231n.github.io/optimization-2/#grad}, 2021.

\bibitem{data2020byzantine}
D.~Data and S.~Diggavi, ``{Byzantine-resilient SGD in high dimensions on
  heterogeneous data},'' \emph{arXiv:2005.07866}, 2020.

\bibitem{el2019sgd}
E.-M. El-Mhamdi, R.~Guerraoui, A.~Guirguis, and S.~Rouault, ``{Sgd:
  Decentralized byzantine resilience},'' \emph{arXiv:1905.03853}, 2019.

\bibitem{fake_acc_1}
``{Facebook has shut down 5.4 billion fake accounts this year},''
  \url{https://www.cnn.com/2019/11/13/tech/facebook-fake-accounts/index.html},
  2019.

\bibitem{fang2020local}
M.~Fang, X.~Cao, J.~Jia, and N.~Z. Gong, ``{Local Model Poisoning Attacks to
  Byzantine-Robust Federated Learning},'' in \emph{USENIX}, 2020.

\bibitem{fed_learning_workshop}
``{Google Workshop on Federated Learning and Analytics},''
  \url{https://docs.google.com/document/d/1dWzVeFLrPinonQMauxIo0oI-Vbvqup5cZzgdPXvu97Y/edit#heading=h.7dsxad3c3nf7},
  2020.

\bibitem{fu2019attack}
S.~Fu, C.~Xie, B.~Li, and Q.~Chen, ``{Attack-resistant federated learning with
  residual-based reweighting},'' \emph{arXiv:1912.11464}, 2019.

\bibitem{fung2020limitations}
C.~Fung, C.~J. Yoon, and I.~Beschastnikh, ``{The limitations of federated
  learning in sybil settings},'' in \emph{RAID}, 2020.

\bibitem{goldblum2020dataset}
M.~Goldblum, D.~Tsipras, C.~Xie \emph{et~al.}, ``{Dataset Security for Machine
  Learning: Data Poisoning, Backdoor Attacks, and Defenses},''
  \emph{arXiv:2012.10544}, 2020.

\bibitem{google_play_protect}
``{Google Play Protect},''
  \url{https://developers.google.com/android/play-protect}, 2021.

\bibitem{huang2011adversarial}
L.~Huang, A.~D. Joseph, B.~Nelson, B.~I. Rubinstein, and J.~D. Tygar,
  ``{Adversarial machine learning},'' in \emph{AISec}, 2011.

\bibitem{jagielski2018manipulating}
M.~Jagielski, A.~Oprea, B.~Biggio, C.~Liu, C.~Nita-Rotaru, and B.~Li,
  ``{Manipulating machine learning: Poisoning attacks and countermeasures
  against regression learning},'' \emph{39th IEEE Symposium on S\&P}, 2018.

\bibitem{jere2020taxonomy}
M.~S. Jere, T.~Farnan, and F.~Koushanfar, ``{A taxonomy of attacks on federated
  learning},'' \emph{IEEE Security \& Privacy}, 2020.

\bibitem{kairouz2019advances}
P.~Kairouz, H.~B. McMahan, B.~Avent \emph{et~al.}, ``{Advances and open
  problems in federated learning},'' \emph{arXiv:1912.04977}, 2019.

\bibitem{konevcny2016federated}
J.~Kone{\v{c}}n{\`y}, H.~B. McMahan, F.~X. Yu, P.~Richt{\'a}rik, A.~T. Suresh,
  and D.~Bacon, ``{Federated learning: Strategies for improving communication
  efficiency},'' \emph{NIPS Workshop on Private Multi-Party ML}, 2016.

\bibitem{krizhevsky2009learning}
A.~Krizhevsky, ``Learning multiple layers of features from tiny images,''
  University of Toronto, Tech. Rep., 2009.

\bibitem{lecun1998gradient}
Y.~LeCun, L.~Bottou, Y.~Bengio \emph{et~al.}, ``{Gradient-based learning
  applied to document recognition},'' \emph{Proceedings of the IEEE}, 1998.

\bibitem{li2019rsa}
L.~Li, W.~Xu, T.~Chen, G.~B. Giannakis, and Q.~Ling, ``{{RSA: Byzantine-robust
  stochastic aggregation methods for distributed learning from heterogeneous
  datasets}},'' in \emph{AAAI}, 2019.

\bibitem{li2021ditto}
T.~Li, S.~Hu, A.~Beirami, and V.~Smith, ``{Ditto: Fair and robust federated
  learning through personalization},'' in \emph{ICML}, 2021.

\bibitem{lin2020ensemble}
T.~Lin, L.~Kong, S.~U. Stich, and M.~Jaggi, ``{Ensemble distillation for robust
  model fusion in federated learning},'' in \emph{NeurIPS}, 2020.

\bibitem{ludwig2020ibm}
H.~Ludwig, N.~Baracaldo, G.~Thomas \emph{et~al.}, ``{IBM Federated Learning: An
  Enterprise Framework White Paper v0.1},'' \emph{arXiv:2007.10987}, 2020.

\bibitem{mcmahan2017communication}
H.~B. McMahan, E.~Moore, D.~Ramage, S.~Hampson, and B.~A.~y. Arcas,
  ``{Communication-efficient learning of deep networks from decentralized
  data},'' in \emph{AISTATS}, 2017.

\bibitem{mhamdi2018the}
E.~M.~E. Mhamdi, R.~Guerraoui, and S.~Rouault, ``{The Hidden Vulnerability of
  Distributed Learning in Byzantium},'' in \emph{ICML}, 2018.

\bibitem{minka2000estimating}
T.~Minka, ``{Estimating a Dirichlet distribution},'' 2000.

\bibitem{munoz2017towards}
L.~Mu{\~n}oz-Gonz{\'a}lez, B.~Biggio, A.~Demontis, A.~Paudice, V.~Wongrassamee,
  E.~C. Lupu, and F.~Roli, ``{Towards poisoning of deep learning algorithms
  with back-gradient optimization},'' in \emph{AISec}, 2017.

\bibitem{munoz2019poisoning}
L.~Mu{\~n}oz-Gonz{\'a}lez, B.~Pfitzner, M.~Russo, J.~Carnerero-Cano, and E.~C.
  Lupu, ``{Poisoning attacks with generative adversarial nets},''
  \emph{arXiv:1906.07773}, 2019.

\bibitem{newell2014practicality}
A.~Newell, R.~Potharaju, L.~Xiang, and C.~Nita-Rotaru, ``{On the practicality
  of integrity attacks on document-level sentiment analysis},'' in
  \emph{AISec}, 2014.

\bibitem{passwords_1}
``{Billion Passwords Stolen: Change All of Yours, Now!}''
  \url{https://www.nbcnews.com/tech/security/billion-passwords-stolen-change-all-yours-now-n174321},
  2014.

\bibitem{passwords_2}
``{Hackers Expose 8.4 Billion Passwords Post them Online in Possibly Largest
  Dump of Passwords Ever},''
  \url{https://www.thegatewaypundit.com/2021/06/hackers-expose-8-4-billion-passwords-post-online-possibly-largest-dump-passwords-ever/},
  2014.

\bibitem{passwords_windows_pc}
``{26 million stolen passwords found online — see if you're affected},''
  \url{https://www.tomsguide.com/news/mystery-malware-info-stealer}, 2021.

\bibitem{paulik2021federated}
M.~Paulik, M.~Seigel, H.~Mason \emph{et~al.}, ``{Federated Evaluation and
  Tuning for On-Device Personalization: System Design \& Applications},''
  \emph{arXiv:2102.08503}, 2021.

\bibitem{pillutla2019robust}
K.~Pillutla, S.~M. Kakade, and Z.~Harchaoui, ``{Robust aggregation for
  federated learning},'' \emph{arXiv:1912.13445}, 2019.

\bibitem{purchase}
``{Acquire Valued Shoppers Challenge at Kaggle},''
  \url{https://www.kaggle.com/c/acquire-valued-shoppers-challenge/data}, 2019.

\bibitem{reddi2020adaptive}
S.~J. Reddi, Z.~Charles, M.~Zaheer, Z.~Garrett, K.~Rush, J.~Kone{\v{c}}n{\`y},
  S.~Kumar, and H.~B. McMahan, ``{Adaptive Federated Optimization},'' in
  \emph{ICLR}, 2020.

\bibitem{android_safetyney}
``{SafetyNet Attestation API},''
  \url{https://developer.android.com/training/safetynet/attestation}, 2021.

\bibitem{shafahi2018poison}
A.~Shafahi, W.~R. Huang, M.~Najibi, O.~Suciu, C.~Studer, T.~Dumitras, and
  T.~Goldstein, ``{Poison frogs! targeted clean-label poisoning attacks on
  neural networks},'' in \emph{NeurIPS}, 2018.

\bibitem{shejwalkar2021manipulating}
V.~Shejwalkar and A.~Houmansadr, ``{Manipulating the Byzantine: Optimizing
  Model Poisoning Attacks and Defenses for Federated Learning},'' in
  \emph{NDSS}, 2021.

\bibitem{simonyan2014very}
K.~Simonyan and A.~Zisserman, ``{Very deep convolutional networks for
  large-scale image recognition},'' in \emph{ICLR}, 2015.

\bibitem{sun2021data}
G.~Sun, Y.~Cong, J.~Dong, Q.~Wang, L.~Lyu, and J.~Liu, ``{Data poisoning
  attacks on federated machine learning},'' \emph{IEEE IoT Journal}, 2021.

\bibitem{sun2019can}
Z.~Sun, P.~Kairouz, A.~T. Suresh, and H.~B. McMahan, ``{Can you really backdoor
  federated learning?}'' \emph{NeurIPS FL Workshop}, 2019.

\bibitem{tolpegin2020data}
V.~Tolpegin, S.~Truex, M.~E. Gursoy, and L.~Liu, ``{Data poisoning attacks
  against federated learning systems},'' in \emph{ESORICS}, 2020.

\bibitem{wang2019neural}
B.~Wang, Y.~Yao, S.~Shan, H.~Li, B.~Viswanath, H.~Zheng, and B.~Y. Zhao,
  ``{Neural cleanse: Identifying and mitigating backdoor attacks in neural
  networks},'' in \emph{40th IEEE Symposium on S\&P}, 2019.

\bibitem{wang2020attack}
H.~Wang, K.~Sreenivasan, S.~Rajput, H.~Vishwakarma, S.~Agarwal, J.-y. Sohn,
  K.~Lee, and D.~Papailiopoulos, ``{Attack of the tails: Yes, you really can
  backdoor federated learning},'' in \emph{NeurIPS}, 2020.

\bibitem{webank_credit}
``{Utilization of FATE in Risk Management of Credit in Small and Micro
  Enterprises},''
  \url{https://www.fedai.org/cases/utilization-of-fate-in-risk-management-of-credit-in-small-and-micro-enterprises/},
  2019.

\bibitem{wu2020mitigating}
C.~Wu, X.~Yang, S.~Zhu, and P.~Mitra, ``{Mitigating backdoor attacks in
  federated learning},'' \emph{arXiv:2011.01767}, 2020.

\bibitem{xiao2012adversarial}
H.~Xiao, H.~Xiao, and C.~Eckert, ``{Adversarial label flips attack on support
  vector machines},'' in \emph{ECAI}, 2012.

\bibitem{xiao2015support}
H.~Xiao, B.~Biggio, B.~Nelson, H.~Xiao, C.~Eckert, and F.~Roli, ``{Support
  vector machines under adversarial label contamination},''
  \emph{Neurocomputing}, 2015.

\bibitem{xie2021crfl}
C.~Xie, M.~Chen, P.-Y. Chen, and B.~Li, ``{{CRFL: Certifiably Robust Federated
  Learning against Backdoor Attacks}},'' in \emph{ICML}, 2021.

\bibitem{xie2019dba}
C.~Xie, K.~Huang, P.-Y. Chen, and B.~Li, ``{{DBA: Distributed backdoor attacks
  against federated learning}},'' in \emph{ICLR}, 2019.

\bibitem{xie2018generalized}
C.~Xie, O.~Koyejo, and I.~Gupta, ``{Generalized byzantine-tolerant sgd},''
  \emph{arXiv:1802.10116}, 2018.

\bibitem{yang2017generative}
C.~Yang, Q.~Wu, H.~Li, and Y.~Chen, ``{Generative poisoning attack method
  against neural networks},'' \emph{arXiv:1703.01340}, 2017.

\bibitem{yin2018byzantine}
D.~Yin, Y.~Chen, K.~Ramchandran, and P.~Bartlett, ``{Byzantine-robust
  distributed learning: Towards optimal statistical rates},'' in \emph{ICML},
  2018.

\bibitem{yu2020salvaging}
T.~Yu, E.~Bagdasaryan, and V.~Shmatikov, ``{Salvaging federated learning by
  local adaptation},'' \emph{arXiv:2002.04758}, 2020.

\end{thebibliography}

\appendix
% !TEX root = main.tex
\subsection{Related Work}\label{related}
% \red{moved here and mentioned in section~\ref{background:agr} that we talk about targeted/backdoor related works in appendix}

% \vspace*{-.5em}
\subsubsection{Targeted and Backdoor Attacks}\label{related:attacks} 

Section~\ref{existing} discusses all state-of-the-art untargeted attacks in detail. Below, we discuss existing works on targeted and backdoor attacks.

\paragraphb{\em Targeted attacks}~\cite{bhagoji2019analyzing,sun2019can,tolpegin2020data} aim to make the global model misclassify a specific set of samples at test time. Bhagoji et al.~\cite{bhagoji2019analyzing} aimed to misclassify a \emph{single sample} and proposed a model poisoning attack based on alternate minimization to make poisoned update look similar to benign updates. \cite{bhagoji2019analyzing} shows that their attack, with a single attacker, can misclassify a single sample with 100\% success against the non-robust Average AGR. try Sun et al.~\cite{sun2019can} investigated constrain-and-scale attack~\cite{bagdasaryan2018how} with the aim to misclassify all samples of a few victim FL clients. Tolpegin et al.~\cite{tolpegin2020data,fang2020local} investigated targeted data poisoning attacks when compromised clients compute their updates by mislabeling the target samples.

\paragraphb{\em Backdoor attacks}~\cite{bagdasaryan2018how,wang2020attack,xie2019dba} aim to make the global model misclassify the samples with adversary-chosen backdoor trigger. Backdoor attacks are \emph{semantic}, if the trigger is naturally present in samples~\cite{bagdasaryan2018how,wang2020attack} and \emph{artificial} if the trigger needs to manually added at test time~\cite{xie2019dba}.
Bagdasaryan at al.~\cite{bagdasaryan2018how} demonstrate a constrain-and-scale attack against simple Average AGR to inject semantic backdoor in the global model. They show that their attacks achieve accuracy of $>$90\% on backdoor task in a next word prediction model. Wang et al.~\cite{wang2020attack} propose data and model poisoning attacks to inject backdoor to misclassify out-of-distribution samples. Xie et al.~\cite{xie2019dba} show how multiple colluding clients can distribute backdoor trigger to improve the stealth of poisoned updates. 
Backdoor (as well as targeted) attacks can be further divided in \emph{specific-label} and \emph{arbitrary-label} attacks. For a backdoored test sample, specific-label attack aims to misclassify it to a specific target class, while arbitrary-label attack aims to  misclassify it to any class.

Note that, trivial extensions of the targeted and backdoor attack algorithms to mount untargeted attacks cannot succeed, because untargeted attacks aim at  affecting almost \emph{all} FL clients and test inputs. For instance, a simple label flipping based data poisoning~\cite{wang2020attack} can insert a backdoor in FL with state-of-the-art defenses. However, such label flipping based untargeted poisoning attacks have no effect even on  unprotected FL (Section~\ref{exp:nonrobust_fl}). 

% Finally, note that, targeted/backdoor attacks are generally more successful as they aim to impact only a few samples. For instance targeted model poisoning attack in~\cite{bhagoji2019analyzing} achieves 

\subsubsection{Existing Defenses Against Targeted and Backdoor Attacks}\label{related:defenses}
In Section~\ref{background:agr}, we discuss the defenses against untargeted poisoning in detail. Here, we review existing defenses against targeted and  backdoor attacks.
FoolsGold~\cite{fung2020limitations} identifies clients with similar updates as attackers, but incur very high losses in performances as noted in~\cite{fu2019attack}. Sun et al.~\cite{sun2019can} investigate efficacy of norm-bounding to counter targeted poisoning and, as we will show, is also effective against untargeted poisoning.
CRFL~\cite{xie2021crfl} counters backdoor attacks by providing certified accuracy for a given test input, but incurs large losses in FL performance (Table~\ref{tab:other_agrs}). Defenses based on pruning techniques~\cite{wu2020mitigating,wang2019neural} remove parts of model that are affected by targeted/backdoor attacks, and hence cannot be used against untargeted attacks which affect the entire model.

% !TEX root = main.tex
% \newpage

% !TEX root = main.tex

\begin{algorithm}
\caption{Our PGA model poisoning attack algorithm}\label{alg:mp}
\small{
\begin{algorithmic}[1]
\State \textbf{Input}: $\nabla_{\{i\in[n']\}}$, $\theta^g$, $f_\mathsf{agr}$, $D_p$
\State $\tau=\frac{1}{n'}\sum_{i\in[n']}\Vert\nabla_i\Vert$
\Comment{Compute norm threshold}

\Comment{$\tau$ is given for norm-bounding AGR}

\State $\theta' \leftarrow A_\mathsf{SGA}(\theta^g, D_p)$
\Comment{Update using stochastic gradient ascent}

\State $\nabla' = \theta'-\theta^g$
\Comment{Compute poisoned update}

\State $\nabla' = f_\mathsf{project}(f_\mathsf{agr}, \nabla', \tau, \nabla_{\{i\in[n']\}})$
\Comment{Scale $\nabla'$ appropriately}

\State\textbf{Output} $\nabla'$

\end{algorithmic}
}

\end{algorithm}

\begin{algorithm}
\caption{The projection function ($f_\mathsf{project}$) of our PGA from Section~\ref{improved:mp}.} \label{alg:scale}
\small{
\begin{algorithmic}[1]
\State \textbf{Input}: $f_\mathsf{agr}$, $\nabla'$, $\tau$, $\nabla_{\{i\in[n']\}}$

\State $d^*=0$
\Comment{Initialize maximum deviation}

\State $\gamma^*=1$
\Comment{Optimal scaling factor that maximizes deviation in~\eqref{eq:gen_opt}}

\State $\nabla' =\frac{\nabla'\times\tau}{\Vert\nabla'\Vert} $
\Comment{Scale $\nabla'$ to have norm $\tau$}

\State $\nabla^b = f_\mathsf{avg}(\nabla_{\{i\in[n']\}})$
\Comment{Compute reference benign update}

\For{$\gamma \in [1,\Gamma]$}
    \State $\nabla'' = \gamma\cdot\nabla'$
    \State $d = \Vert f_\mathsf{agr}(\nabla''_{\{i\in[m]\}}, \nabla_{\{i\in[n']\}}) -\nabla^b\Vert$
    \State $\gamma^*=\gamma$ \textbf{if} $d>d^*$
    \Comment{Update optimal $\gamma$}
    \State $\gamma = \gamma + \delta$
    \Comment{Update $\gamma$}
\EndFor
\State\textbf{Output} $\gamma^*\cdot\nabla'$
\end{algorithmic}
}
\vspace*{-.25em}
\end{algorithm}
\vspace*{-1.em}

\subsection{Missing details of our data and model poisoning attacks from Sections~\ref{improved:dp} and~\ref{improved:mp}}\label{appdx:missing_attacks}
% \vspace*{-.5em}

\subsubsection{Missing data poisoning attack methods}\label{appdx:missing_attacks:dp}

\paragraphb{\em Multi-krum.}
Following~\cite{shejwalkar2021manipulating}, our attack aims to maximize the number of poisoned updates in the selection set ($S$) of Multi-krum AGR (Section~\ref{agr:mkrum}).
As the size of $S$ is fixed, maximizing the number of poisoned updates in $S$ implicitly means minimizing the number of benign updates.
This objective is formalized as:
\begin{align}\label{eq:obj_dp_mkrum}
\underset{D_p\subset D'_p}{\text{argmax}} \quad m' = |\{\nabla\in\nabla'_{\{i\in[m]\}} | \nabla\in S \}|
\end{align}
where $D'_p$ is all the available labels flipped data and $m'$ is the final number of poisoned updates in $S$ of Multi-krum.

We solve~\eqref{eq:obj_dp_mkrum} based on an observation: In Figure~\ref{fig:femnist_obj}-(d) we vary $|D_p|$ and plot the fraction of corresponding poisoned updates that Multi-krum selects.
Let $|D|_\mathsf{avg}$ be the average dataset size of benign clients, e.g., $|D|_\mathsf{avg}$ is 23.7 for FEMNIST.
Note from Figure~\ref{fig:femnist_obj}-(d) that, even for $|D_p|$ slightly higher than $|D|_\mathsf{avg}$, Multi-krum easily discards most of the poisoned updates. Only when $|D_p|$ is small ($\sim$10), Multi-krum selects most of the poisoned updates. 
Hence, we sample $D_p\subset D'_p$, where we vary $|D_p|\in [0.5\cdot|D|_\mathsf{avg}, 3\cdot|D|_\mathsf{avg}]$, and check the poisoning impact of $D_p$ on Multi-krum; to reduce variance, we repeat this 10 times for each $|D_p|$.
We report the results for $D_p$ with the maximum poisoning impact.

\paragraphb{\em Trimmed-mean.}
For Trimmed-mean AGR (Section~\ref{agr:trimmed-mean}), we use the objective in~\eqref{eq:gen_opt_dp}, but it is cumbersome to solve it directly. Hence, similar to our attacks on Average and Norm-bounding AGRs, we use large $|D_p|$ for poisoned data on each of the compromised clients.
Our approach is based on the observation in Figure~\ref{fig:femnist_obj}-(c): The higher the $|D_p|$ (obtained using DLF/SLF strategies), the higher the Trimmed-mean objective value, i.e.,  $\Vert\nabla^p-\nabla^b\Vert$.

\subsubsection{Missing model poisoning attack methods}\label{appdx:missing_attacks:mp}

\paragraphb{\em Multi-krum.}
Similar to our DPA (Section~\ref{improved:dp}), the  objective of our MPA on Multi-krum is to maximize the number of poisoned updates in the selection set $S$.
We aim to find a scaling factor $\gamma$ for $\nabla'$ such that maximum number of $\nabla''=\gamma\nabla'$ are selected in $S$. This is formalized below:
\begin{align}\label{eq:obj_mp_mkrum}
\underset{\gamma^*\in\mathbb{R}}{\text{argmax}} \quad m = |\{\nabla\in\nabla''_{\{i\in[m]\}} | \nabla\in S \}|
\end{align}

To solve the optimization in \eqref{eq:obj_mp_mkrum}, our $f_\mathsf{project}$ searches for the maximum $\gamma$ in a pre-specified range $[1,\Gamma]$ such that Multi-krum selects all the scaled poisoned updates.
Specifically, in Algorithm~\ref{alg:scale}, instead of computing the deviation (line-8), we compute the number of $\nabla''$ selected in $S$ and update $\gamma^*$ if $S$ has all of $\nabla''$s.

\paragraphb{\em Trimmed-mean.}
Here, we directly plug Trimmed-mean algorithm in Algorithm~\ref{alg:scale} (line-8).
% with $\Gamma=100$ and $\delta=5$\red{why these numbers?}. 
Our attack is similar to that of~\cite{shejwalkar2021manipulating}, but instead of using one of several perturbation vectors, $\omega$'s, we use stochastic gradient ascent to tailor $\omega$ to the entire FL setting ( e.g., $\theta^g$, data, optimizer, etc.) to improve the attack impact.

\section{Additional experimental details}\label{additional_exp_details}

\subsection{Experimental setup} \label{exp_setup}
Real-world FL datasets~\cite{gboard,paulik2021federated} are proprietary and cannot be publicly accessed. Hence, we follow the literature on untargeted poisoning in FL~\cite{fang2020local,shejwalkar2021manipulating,baruch2019a,tolpegin2020data} and focus on image and categorical datasets. But, we ensure that our setup embodies the production FL~\cite{kairouz2019advances}, e.g., by using large number of clients with extremely non-iid datasets.

\subsubsection{Datasets and Model Architectures}

\paragraphb{FEMNIST~\cite{caldas2018leaf,cohen2017emnist}} is a character recognition classification task with 3,400 clients, 62 classes (52 for upper and lower case letters and 10 for digits), and 671,585  grayscale images. 
Each client has data of her own handwritten digits or letters. Considering the huge number of clients in real-world cross-device FL (up to $10^{10}$), we further divide each of the clients' data in $p\in\{2,5,10\}$ non-iid parts using Dirichlet distribution~\cite{minka2000estimating} with $\alpha=1$. Increasing the Dirichlet distribution parameter, $\alpha$, generates more iid datasets. Unless specified otherwise, we set $p=10$, i.e., the total number of clients is 34,000. We use LeNet~\cite{lecun1998gradient} architecture.
% FEMNIST is a non-iid, class-imbalanced dataset commonly encountered in cross-device FL settings~\cite{kairouz2019advances}, while the previous datasets are more common in cross-silo FL settings.

\paragraphb{CIFAR10~\cite{krizhevsky2009learning}} is a 10-class classification task with 60,000 RGB images (50,000 for training and 10,000 for testing), each of size 32 $\times$ 32.
% `Class-balanced' datasets have the same number of samples per class, e.g., each class of CIFAR10 has 6,000 images.
% We use 5000 clients each with 1,000 samples and use validation and test data of sizes 5,000 each. 
Unless specified otherwise, we consider 1,000 total FL clients and divide the 50,000 training data using Dirichlet distribution~\cite{minka2000estimating} with $\alpha=1$. We use VGG9 architecture with batch normalization~\cite{simonyan2014very}.

\paragraphb{Purchase~\cite{purchase}} is a   classification task with 100 classes and 197,324 binary feature vectors each of length 600. We use 187,324 of total data for training and divide it among 5,000 clients using Dirichlet distribution with $\alpha=1$. We use validation and test data of sizes 5,000 each. We use a fully connected network with layer sizes \{600, 1024, 100\}. 
% Due to space restriction, we defer the details of the parameters of FL training and various attacks to Section~\ref{appdx:setup}.

% % \vspace*{-.75em}
% % \subsubsection{Measurement Metric}\label{metric}
% % %
% % $A_\theta$ denotes the maximum accuracy that the global model achieves over all  FL training rounds, without any attack. $A^*_\theta$ for an attack denotes the maximum accuracy of the model under the given attack. We define \emph{attack impact}, $I_\theta$, as \emph{the reduction in the accuracy of the global model due to the attack}, hence for a given attack,  $I_\theta = A_\theta - A^*_\theta$.

\subsubsection{Details of Federated learning and attack parameters}\label{appdx:setup}

For FEMNIST, we use 500 rounds, batch size, $\beta=10$, $E=5$ local training epochs, and in the $e^{th}$ round use SGD optimizer with a learning rate $\eta=0.1\times 0.995^{e}$ for local training; we select $n=50$ clients per round and achieve baseline accuracy $A_\theta$=82.4\% with $N$=34,000 clients.
For CIFAR10, we use 1,000 rounds, $\beta=8$, $E=2$, and in the $e^{th}$ round use SGD with momentum of 0.9 and $\eta=0.01\times 0.9995^{e}$; we use $n=25$ and achieve $A_\theta$=86.6\% with  $N$=1,000.
For Purchase, we use 500 rounds, $\beta=10$, $E=5$, and in the $e^{th}$ round use SGD with $\eta=0.1\times 0.999^{e}$; we use $n=25$ and achieve $A_\theta$=81.2\% with $N$=5,000.

We generate large poisoned data $D_p$ required for our DPAs (Section~\ref{improved:dp}) by combining the dataset of compromised clients and adding Gaussian noise to their features. We round the resulting feature for categorical Purchase dataset.

% Our DPAs (Section~\ref{improved:dp}) need large amounts of data (feature vectors) which are not  available for the smaller datasets. Hence, we expand the datasets by generating synthetic data by adding Gaussian noise with mean 0 and standard deviation of 0.01 to existing data samples. Purchase has binary features, hence we use round the obfuscated data to obtain new binary data. We keep the labels of the obfuscated data the same as the original data. For our MPAs (Section~\ref{improved:mp}), we use the same learning parameters for SGA on compromised clients and SGD on benign clients.
% \vspace*{-.5em}

\begin{figure}
\centering
\hspace*{-1em}
\includegraphics[scale=.45]{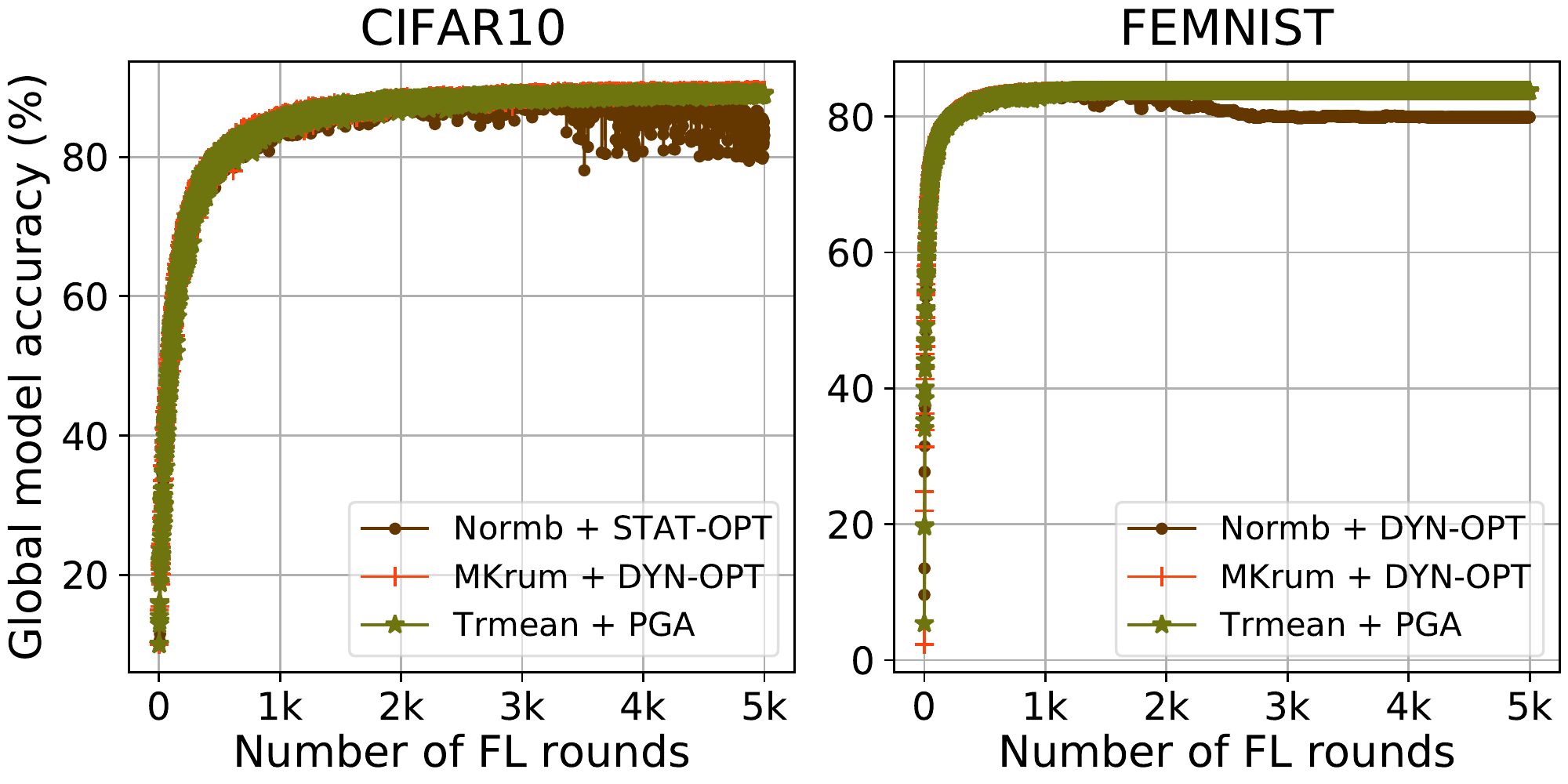}
% \vspace*{-1em}
\caption{Even with a very large number of FL rounds (5,000), the state-of-the-art model poisoning attacks with $M$=0.1\% cannot break the robust AGRs (Section~\ref{exp:robust_fl}).}
\label{fig:5k_rounds}
% \vspace*{-2em}
\end{figure}

\subsection{Explanations of effects of $|D|_\mathsf{avg}$ from Section~\ref{exp:fl_data_sizes}}\label{appdx:fl_data_sizes}

% In Section~\ref{exp:fl_data_sizes}, we study the effect of the average of local dataset sizes of benign clients ($D_\mathsf{avg}$) and present the attack impacts ($I_\mathsf{\theta}$) in Figure~\ref{fig:total_nclients_impact_ai}. \emph{We observe no clear effect of $|D|_\mathsf{avg}$ on the impacts of attacks}. But, in this section, we give the intuitive explanations behind our observations.

At $M$=1\%, $I_\mathsf{\theta}$'s of STAT-OPT on CIFAR10 + Normb reduce with increase in $|D|_\mathsf{avg}$. This is because, increasing $|D|_\mathsf{avg}$ improves the quality of updates of benign clients, but does not improve the attacks. Hence, when the benign impact of benign updates overpowers the poisoning impact of poisoned updates, $I_\mathsf{\theta}$'s reduce. 

On the other hand, $I_{\theta}$'s of any attacks on FEMNIST with robust AGRs do not change with varying $|D|_\mathsf{avg}$. This is because, FEMNIST is an easy task, and therefore, the presence of compromised clients does not affect the global models.

Interestingly, $I_{\theta}$ of MPAs on CIFAR10 with Average AGR \emph{increases} with $|D|_{\mathsf{avg}}$. This is because, due to the difficulty of CIFAR10 task, MPAs on CIFAR10 with Average AGR are very effective and when the server selects even a single compromised client, it completely corrupts the global model. 
% Hence, as we show in Figure~\ref{fig:total_nclients_impact_acc}, with increasing $|D|_{\mathsf{avg}}$, the accuracy of global model with the MPA (i.e., $A^*_{\thata}$) remains close to 10\%, but the accuracy of global model without any attacks (i.e., $A_{\thata}$) increases. Hence we see increase in $I_{\theta}$ (i.e., $A_{\theta} - A^*_{\theta}$).

\subsection{Miscellaneous figures}\label{appdx:missing_figures}
Below, we provide all the missing figures and the corresponding sections in main paper.

\begin{compactitem}
    \item Figure~\ref{fig:5k_rounds} for Section~\ref{exp:robust_fl} shows the impacts of strongest of model poisoning attacks on robust AGRs over a very large number of FL rounds.
    
    \item  Figure~\ref{fig:diff_arch} for Section~\ref{exp:unknown_arch} shows impact of unknown architecture on our state-of-the-art data poisoning attacks from Section~\ref{improved:dp}. Table~\ref{tab:unknown_arch} shows the convolutional neural network architecture that the adversary uses as a substitute to the true LeNet architecture.
    
    \item Figure~\ref{fig:cross_silo} for Section~\ref{exp:cross_silo} shows impacts of data poisoning attacks on cross-silo FL.
    
    \item Figure~\ref{fig:round_nclients_impact} for Section~\ref{exp:round_nclients} shows impacts of poisoning attacks for increasing the number of clients selected in each FL round.
    
    \item Figures~\ref{fig:total_nclients_impact_ai} and~\ref{fig:total_nclients_impact_acc} for Section~\ref{exp:fl_data_sizes} show the attack impacts and accuracy of the global model, respectively, when the average size of benign clients' local data increases.
    
    \item Figure~\ref{fig:total_nclients_impact_acc_mkrum_trmean} for Section~\ref{exp:fl_data_sizes} shows attack impacts (on the left y-axes) and global model accuracy (on the right y-axes) for Multi-krum and Trimmed-mean robust AGRs for CIFAR10 and FEMNIST datasets.

\end{compactitem}

\begin{figure}
% \vspace*{-2em}
\centering
\includegraphics[scale=.5]{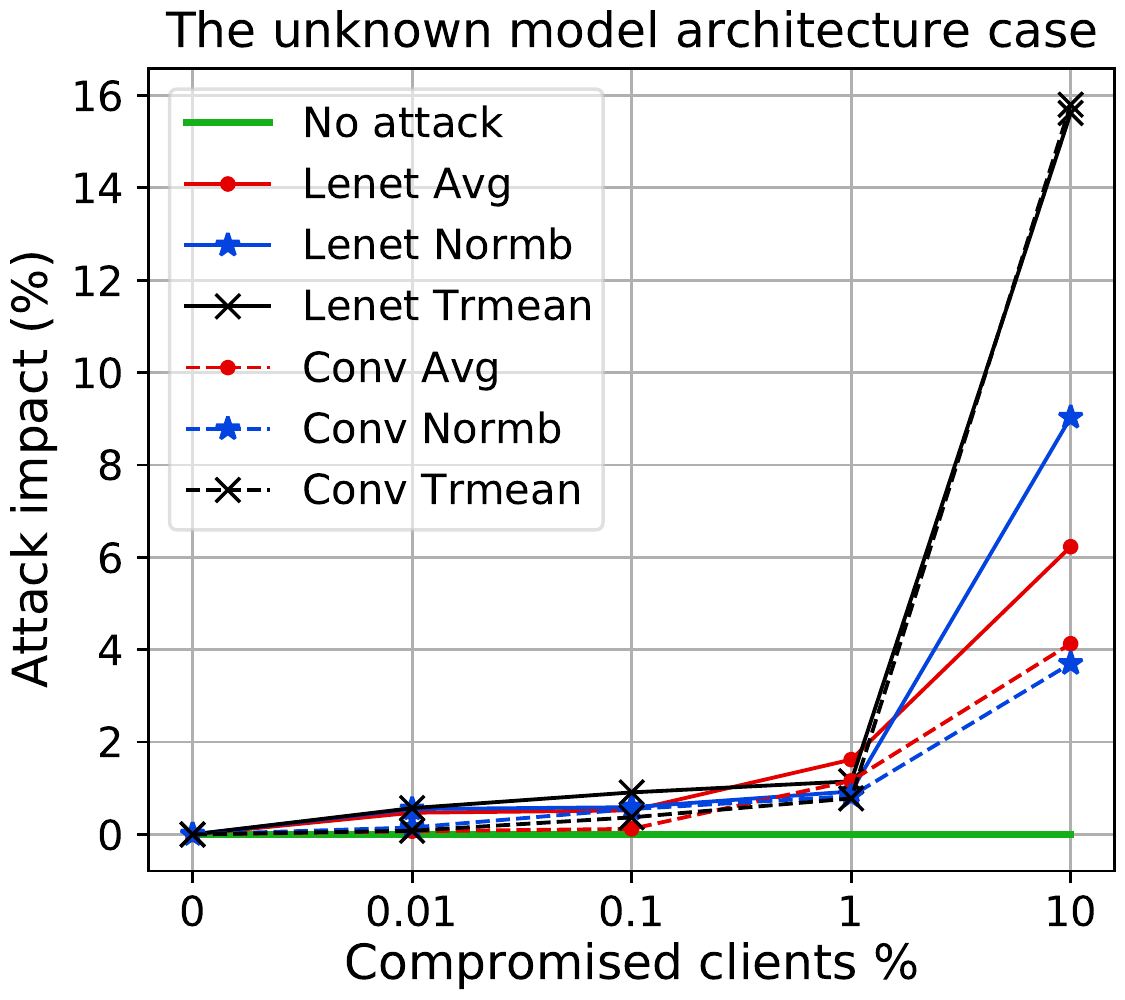}
\vspace*{-.5em}
\caption{As discussed in Section~\ref{exp:unknown_arch}, impacts of the DPA-DLF attack from Section~\ref{improved:dp} reduce if the architectures of the surrogate and the global model are different.}
\label{fig:diff_arch}
\vspace*{-1em}
\end{figure}

\begin{table}
\caption{The architecture of the surrogate model that we use to emulate the unknown architecture setting (Section~\ref{exp:unknown_arch}).} \label{tab:unknown_arch}
% \vspace*{-.5em}
\centering
% \fontsize{9}{10}\selectfont{}
\begin{tabular} {|c|c|}
  \hline
  Layer name & Layer size \\ \hline
  Convolution + Relu & $5\times 5\times 32$ \\ \hline
  Max pool & $2\times 2$ \\ \hline
  Convolution + Relu & $5\times 5\times 64$ \\ \hline
  Max pool & $2\times 2$ \\ \hline
  Fully connected + Relu & 1024 \\ \hline
  Softmax & 62 \\ \hline
\end{tabular}
% \vspace*{-2em}
\end{table}

\begin{figure}
\centering
\hspace*{-1em}
\includegraphics[scale=.45]{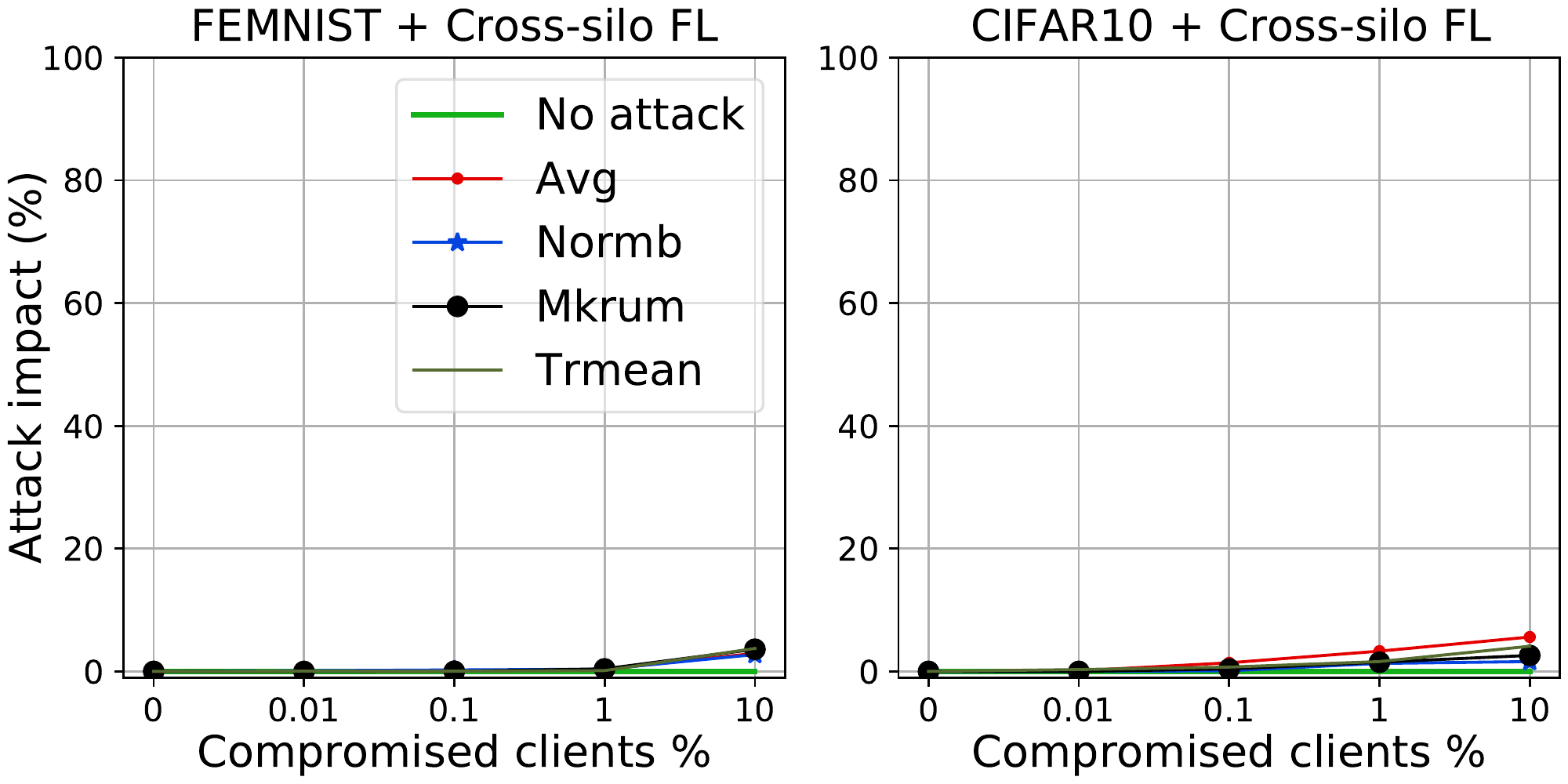}
% \vspace*{-.75em}
\caption{All data poisoning attacks have negligible impacts on cross-silo FL, when compromised clients are concentrated in a few silos or distributed uniformly across silos (Section~\ref{exp:cross_silo}).
}
\label{fig:cross_silo}
% \vspace*{-1.25em}
\end{figure}

\begin{figure}
\centering
\hspace*{-1em}
\includegraphics[scale=.655]{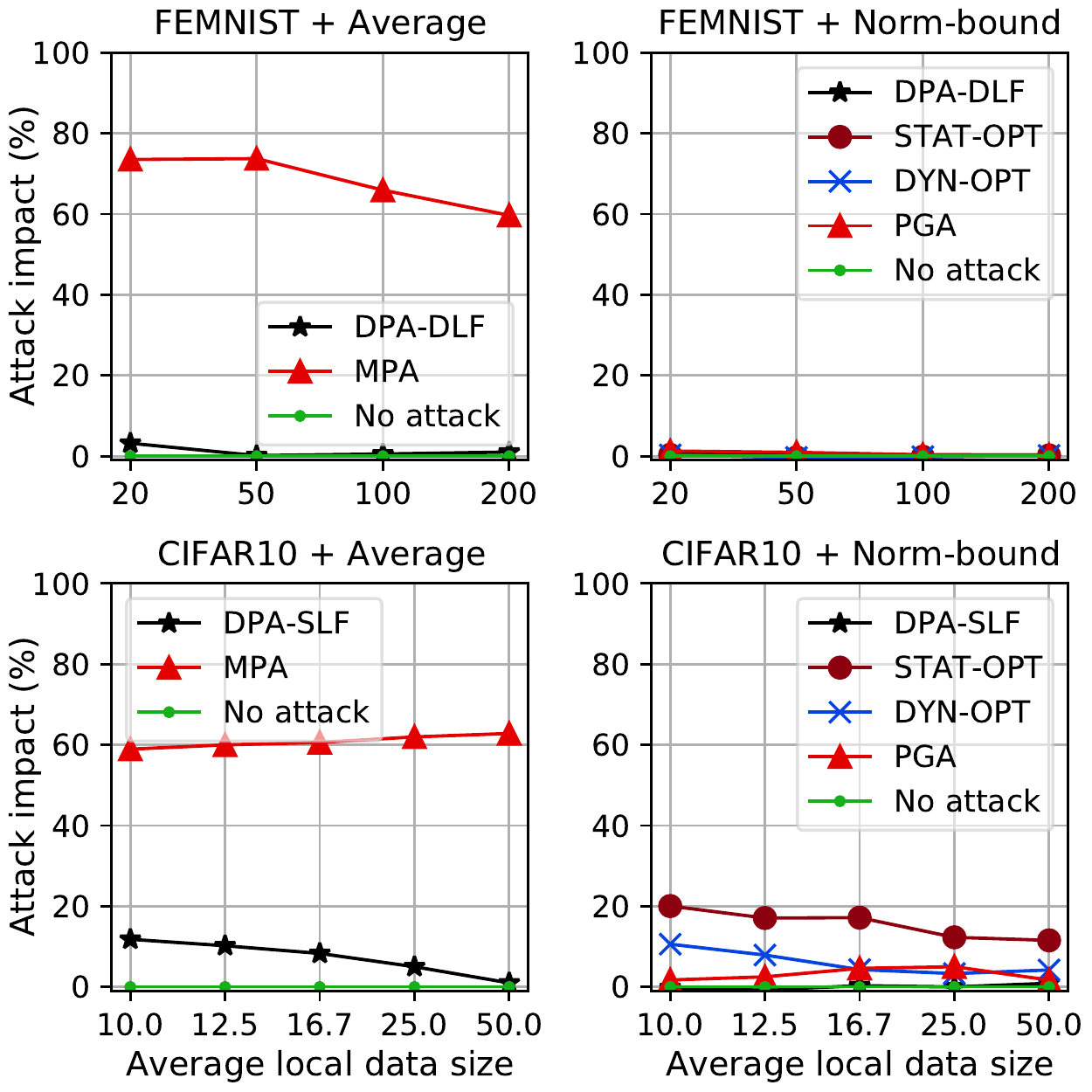}
\vspace*{-.5em}
\caption{With 1\% compromised clients, \emph{increasing $|D|_\mathsf{avg}$ has no clear pattern of effects of  on attack impacts, but it increases the global model accuracy} as shown in Figure~\ref{fig:total_nclients_impact_acc}. Figure~\ref{fig:total_nclients_impact_acc_mkrum_trmean} shows the plots of attack impacts and the global model accuracy for Multi-krum and Trimmed-mean AGRs. 
}
\label{fig:total_nclients_impact_ai}
\vspace*{-1.5em}
\end{figure}

\begin{figure*}
\centering
\vspace*{-.25em}
\hspace*{-1em}
\includegraphics[scale=.7]{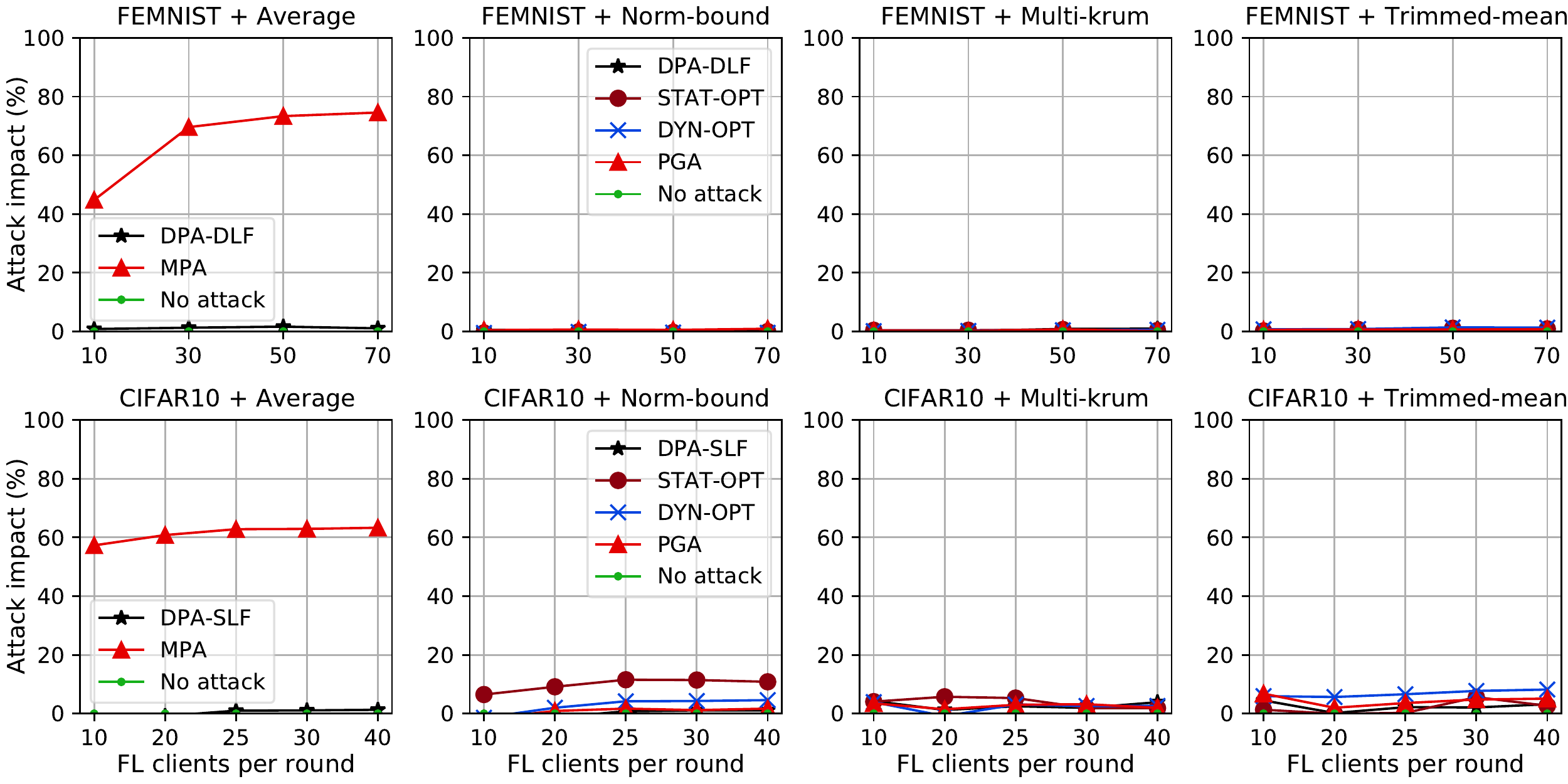}
% \vspace*{-.25em}
\caption{As discussed in Section~\ref{exp:round_nclients}, the number of clients, $n$, chosen in each FL round has no noticeable effect on the attack impacts, with the exception of model poisoning on Average AGR. We use $M=1\%$ of compromised clients. 
% We observe the same for Mkrum and Trmean, but omit due to space restrictions.
}
\label{fig:round_nclients_impact}
% \vspace*{-1em}
\end{figure*}

\begin{figure}
\centering
\vspace*{-.25em}
\hspace*{-1em}
\includegraphics[scale=.7]{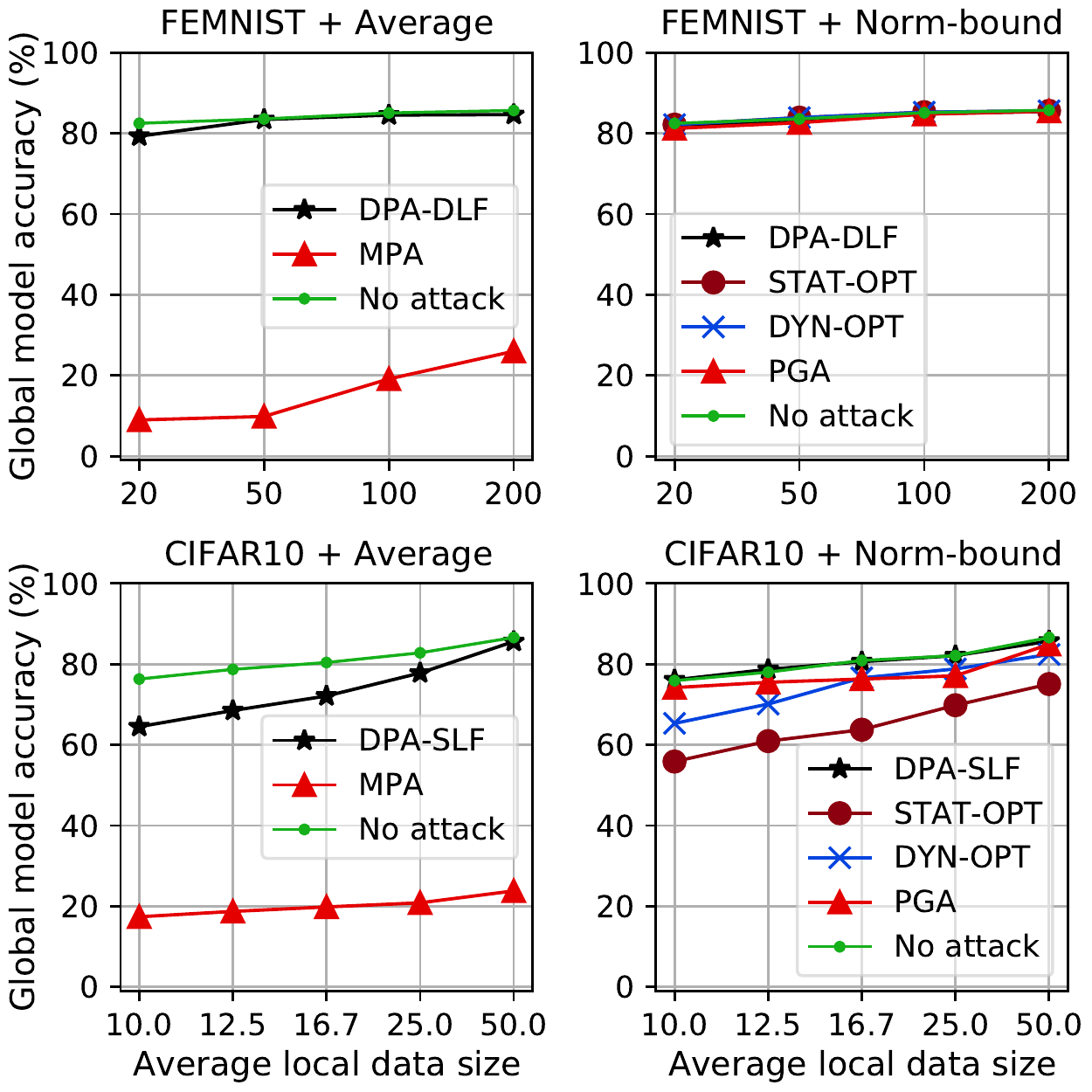}
% \vspace*{-.25em}
\caption{Effect on \textbf{the accuracy of global models} of the average of local dataset sizes, $|D|_\mathsf{avg}$, of the benign clients, with 1\% compromised clients. As discussed in Section~\ref{exp:fl_data_sizes}, \emph{increasing $|D|_\mathsf{avg}$ increases the accuracy of the global models}.
}
\label{fig:total_nclients_impact_acc}
% \vspace*{-1em}
\end{figure}

\begin{figure}
\centering
\vspace*{-.25em}
\hspace*{-1em}
\includegraphics[scale=.75]{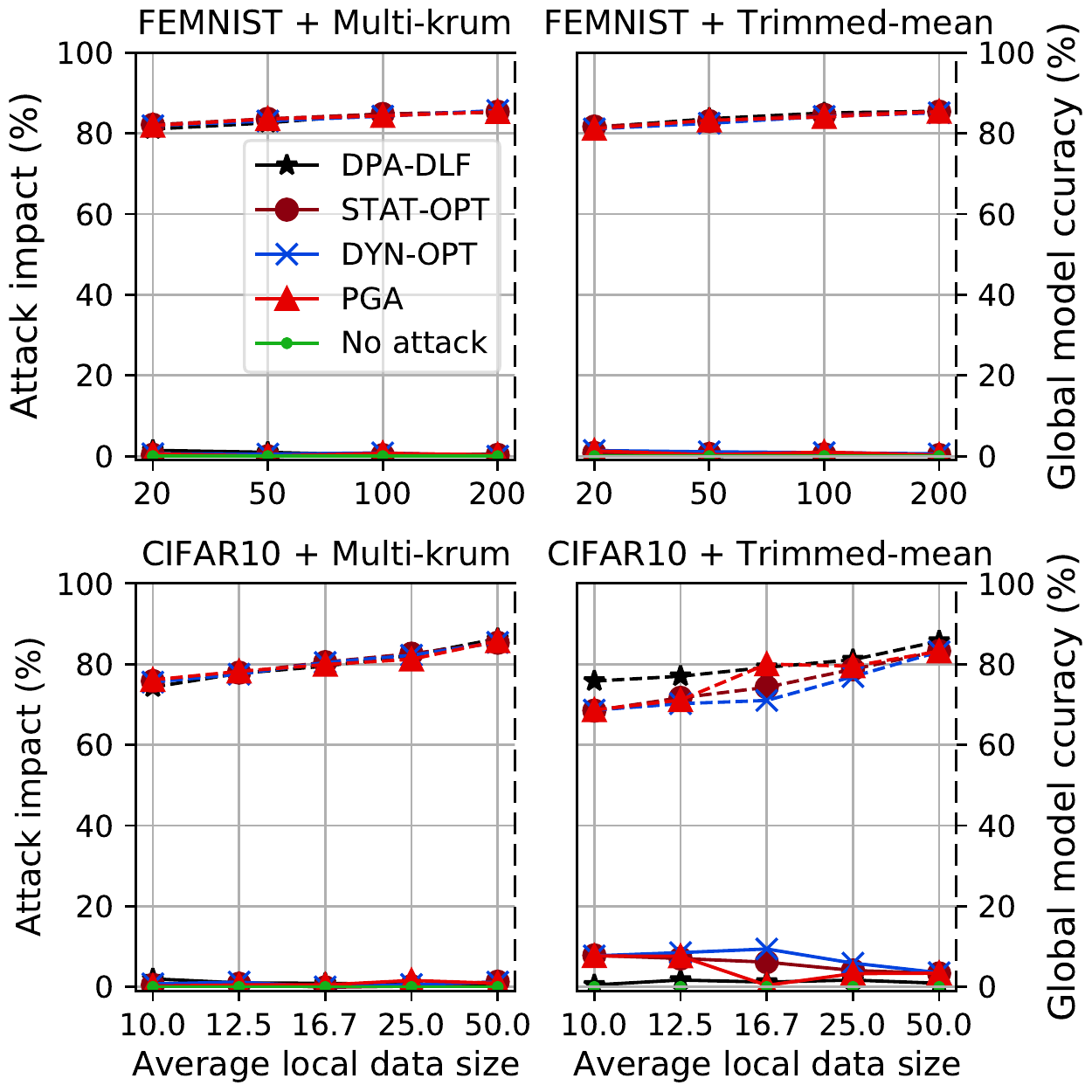}
% \vspace*{-.5em}
\caption{ We make observations similar to Average and Norm-bound AGRs (Figures~\ref{fig:total_nclients_impact_ai},~\ref{fig:total_nclients_impact_acc} in Section~\ref{exp:fl_data_sizes}) for Multi-krum and Trimmed-mean about the effect of $|D|_\mathsf{avg}$ on the attack impacts (left y-axes, solid lines) and on the global model accuracy (right y-axes, dotted lines), with $M$=1\%. All y-axes are from 0 to 100.
}
\label{fig:total_nclients_impact_acc_mkrum_trmean}
% \vspace*{-1em}
\end{figure}

% \begin{figure*}
% \centering
% % \hspace*{-1em}
% \includegraphics[scale=.65]{figures/new_impact_round_nclients_cifar10_vggbn_femnist_ai.pdf}
% % \vspace*{-1em}
% \caption{The number of clients, $n$, chosen in each FL round has no noticeable effect on the attack impacts, with the exception of model poisoning on Average AGR. We use $M=1\%$ of compromised clients}
% \label{fig:round_nclients_impact}
% % \vspace*{-1em}
% \end{figure*}

% \begin{figure*}
% \centering
% % \hspace*{-1em}
% \includegraphics[scale=.65]{figures/new_impact_total_nclients_cifar10_vggbn_femnist_er.pdf}
% % \vspace*{-1em}
% \caption{Effect on \textbf{the accuracy of global models} of the average of local dataset sizes, $|D|_\mathsf{avg}$, of the benign clients, with 1\% compromised clients. \emph{Increasing $|D|_\mathsf{avg}$ increases the accuracy of the global models}.}
% \label{fig:total_nclients_impact_acc}
% % \vspace*{-1em}
% \end{figure*}

% \input{practical_threat_models}

% %-------------------------------------------------------------------------------
% \bibliographystyle{plain}
% \bibliography{\jobname}
}
%%%%%%%%%%%%%%%%%%%%%%%%%%%%%%%%%%%%%%%%%%%%%%%%%%%%%%%%%%%%%%%%%%%%%%%%%%%%%%%%
\end{document}